\documentclass{article}

    \usepackage[final]{neurips_2020}

\usepackage[utf8]{inputenc} 
\usepackage[T1]{fontenc}    
\usepackage{hyperref}       
\usepackage{url}            
\usepackage{booktabs}       
\usepackage{amsfonts}       
\usepackage{nicefrac}       
\usepackage{microtype}      

\usepackage{wrapfig}
\usepackage{sidecap}
\usepackage{graphicx}
\usepackage{subcaption}
\usepackage{amsfonts}
\usepackage{amsmath,xparse,bm}

\usepackage{graphicx}
\usepackage[ruled]{algorithm2e}
\usepackage[title]{appendix}
\usepackage{array}
\usepackage{multirow}

\usepackage{biblatex}
\newcommand{\argmin}{\mathop{\mathrm{argmin}}\limits}

\title{Policy Manifold Search\\for Improving Diversity-based Neuroevolution}

\author{
    Nemanja Rakicevic\\
    Imperial College London\\
    \texttt{n.rakicevic@imperial.ac.uk} \\
\And
    Antoine Cully\\
    Imperial College London \\
    \texttt{a.cully@imperial.ac.uk} \\
\AND
    Petar Kormushev\\
    Imperial College London \\
    \texttt{p.kormushev@imperial.ac.uk}
}

\begin{document}

\maketitle

\begin{abstract}
    Diversity-based approaches have recently gained popularity as an alternative paradigm to performance-based policy search.
    A popular approach from this family, Quality-Diversity (QD), maintains a collection of high-performing policies separated in the diversity-metric space, defined based on policies' rollout behaviours. 
    When policies are parameterised as neural networks, i.e. Neuroevolution, QD tends to not scale well with parameter space dimensionality.
    Our hypothesis is that there exists a low-dimensional manifold embedded in the policy parameter space, containing a high density of diverse and feasible policies.
    We propose a novel approach to diversity-based policy search via Neuroevolution, that leverages learned latent representations of the policy parameters which capture the local structure of the data.
    Our approach iteratively collects policies according to the QD framework, in order to
    (i) build a collection of diverse policies,
    (ii) use it to learn a latent representation of the policy parameters,
    (iii) perform policy search in the learned latent space.
    We use the Jacobian of the inverse transformation (i.e. reconstruction function) to guide the search in the latent space.
    This ensures that the generated samples remain in the high-density regions of the original space, after reconstruction.
    We evaluate our contributions on three continuous control tasks in simulated environments, and compare to diversity-based baselines. 
    The findings suggest that our approach yields a more efficient and robust policy search process.
\end{abstract}

    \vspace{-0.1cm}

    \begin{figure}[!h]
    \centering
        \includegraphics[width=.8\textwidth]{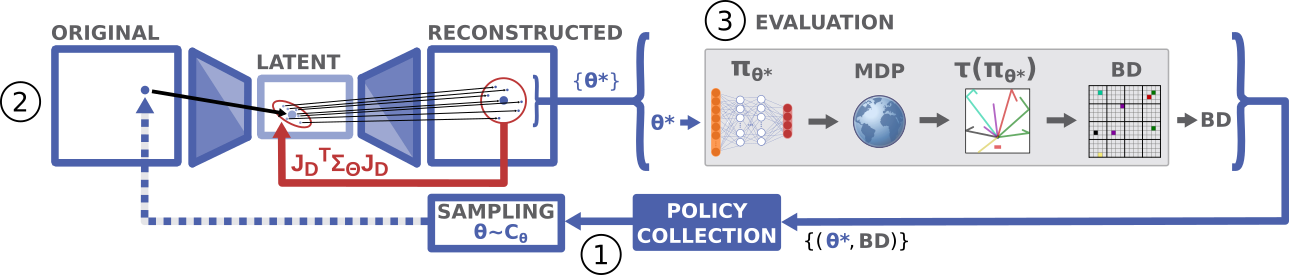}
        \caption{
           Components of the parameter search phase within the Policy Manifold Search approach: \\
            (1) sampling new policy parameters from the collection, (2) exploration in the learned latent space and covariance matrix scaling, (3) evaluation of the generated parameters and adding to the collection.
        } 
        \label{fig:schematics}
    \end{figure}

\vspace{-0.2cm}


\begin{refsection}

\section{Introduction}
    %
    %
    %
 %
  %
   %
    %
    %
    %
    %
    In recent years, we have seen significant progress in tackling continuous control tasks, owing to Deep Reinforcement Learning (RL) approaches relying on gradient-based policy optimisation \cite{schulman2017proximal}, as well as the gradient-free alternative, Neurevolution \cite{salimans2017evolution, such2017deep}.
    %
    %
    Optimising for task performance leads to a unique, locally-optimal solution, which corresponds to a unique set of policy NN parameters.
    %
    %
    In contrast to this, a diversity of solutions is sometimes necessary in cases where one environment can accommodate multiple tasks, or when different controllers are needed to solve and adapt to a dramatically changing environment or recover from damage \cite{cully2015robots}.
    Diversity can be implemented either as multiple distinct solutions \cite{pugh2016quality, cully2017quality}, or one solution modifiable during environment interactions \cite{finn2017model, rusu2018meta, eysenbach2018diversity, hausman2018transferable}. 
    Quality-Diversity (QD) \cite{pugh2016quality, cully2017quality} framework is an example of the former and introduces collections of solutions.
    Each different solution is assigned to the collection based on their \textit{behaviour descriptor}, i.e. low-dimensional representation of the corresponding policy's behaviour in the environment~\cite{cully2017quality}. 
    New solutions are then generated by modifying existing ones via \textit{mutation operators}.
    %
    %
    %
    %
    However, QD tends not to scale well with high-dimensional parameterisations \cite{vassiliades2018discovering, gaier2020discovering}.
    %
    %
    %
    %
    Inspired by the \textit{manifold hypothesis} \cite{cayton2005algorithms, bengio2013representation} and prior successes in representation learning \cite{hausman2018transferable, pere2018unsupervised}, we hypothesise that there exists a lower-dimensional, non-linear manifold, embedded in the high-dimensional policy parameter space, which contains a high density of solutions for a particular task.
    %
    %
    %
    
    In this paper, we propose a new approach, Policy Manifold Search, that learns a manifold in the policy parameter space which is used for policy search. The goal is to find and collect a diversity of policies that work in a given environment. 
    Focusing the search in the high solution density manifold improves the sample complexity and the coverage of diverse behaviours.
    %
    %
    What distinguishes our approach from standard representation learning approaches that learn latent embeddings, is that we do not use only point-wise embeddings but also the transformation function information.
    This leads to a consistent parameter search regardless of the local structure of the latent representation, by accounting for the distortions induced by the transformation function.
    Moreover, representation learning approaches usually focus on unique data e.g. states, images etc. Conversely, NN weights can have multiple configurations which lead to the same behaviour, thus adding complexity.
    
    The main question we are addressing in this paper: 
    %
    \textit{is the learned parameter space representation rich enough to improve policy search?}
    In order to answer this, we conduct ablation studies of the algorithm components, as well as comparisons to diversity-based baselines. 

\vspace{-0.1cm}
\section{Method}
\label{sec:method}
    
    %
    We present Policy Manifold Search (PoMS), an iterative algorithm, where each iteration consists of the \textit{parameter manifold learning} and \textit{parameter search} phases.
    The former consists of obtaining a latent representation and its corresponding transformations, while the latter runs MAP-Elites (presented in Appendix A) in the learned latent parameter space to generate new diverse policies, as shown in Fig.~\ref{fig:schematics}.
    Each iteration can add new policies in the collection, which are then used to refine the latent representation in the next iteration.
    The PoMS pseudocode is presented in Algorithm 1 of Appendix A.
    
    \textbf{Preliminaries}
        We consider a typical RL setting, with a deterministic environment defined as a Markov Decision Process.
        %
        We define a deterministic policy $\pi_{\theta}$, parameterised as a deep neural network, that maps the current state to the action to be taken at that state $a_t = \pi_{\theta}(s_t)$.
        The policy parameters $\theta$ are a P-dimensional set of network weights and biases, $\theta \in \mathbb{R}^{P}$, such that each point in the policy parameter space defines a unique policy.
        We use the term \textit{original parameter space} to refer to this space of policy parameters.
        %
        %
        During an episode of length $T$, an agent interacts with the environment using the policy $\pi_{\theta}$, thus generating a trajectory $\tau = \{s_i, a_i\}_1^T$.
        We want to distinguish how a certain deterministic policy $\pi_{\theta_k}$ interacts with the environment in a quantifiable way. To this end, we use the concept of a \textit{Behaviour Descriptor} (BD) from the QD literature~\cite{cully2017quality} which aims at uniquely describing the episode rollout. The BD is formalized as a mapping from a state-trajectory $\tau$ space to $b$-dimensional behaviour space $\mathcal{BD}: \mathrm{T} \rightarrow \mathcal{B}$.
        Different policies can produce the same BD, but a specific deterministic policy will map to a unique behaviour (surjective mapping).
        

    \vspace{-0.1cm}
    \subsection{Parameter Manifold Learning Phase}
        The main insight of the PoMS is learning a lower-dimensional manifold $\mathbb{R}^{M}$, embedded in $\mathbb{R}^{P}$ where $M<<P$, around which a high-density of interesting (i.e. nondegenerate) policies are located. 
        This manifold can then serve as smaller search space for a more efficient exploration.
        In order to obtain this manifold, we start by generating a uniformly sampled initial set of parameters, $\theta_i \sim \mathcal{U}(-1,1)$, which are added to the the MAP-Elites collection.
        This collection of parameters is then used to train an invertible dimensionality reduction model, like a deep AutoEncoder (AE).
        %
        %
        %
        %
        Each point in the original parameter space $\theta_i$ can be directly mapped into the corresponding point on the manifold $z_i  \in \mathbb{R}^{M}$ using the encoder ($f_{E}$), and reconstructed back using the decoder ($f_{D}$).
        %
        %
        We use a fully-connected, symmetrical AE with a reconstruction loss:
        $\mathcal{L}_{AE} = \left \| \theta_i -  \hat{\theta}_i \right \|$
        %
        %
        where $\hat{\theta}_i = f_{D} \circ f_{E}(\theta_i)$.
        The bottleneck layer of the AE defines the latent parameter representation space.
        %
        We do not apply any specific regularisation on the latent space, but simply train the AE in an unsupervised manner to achieve good reconstruction.
        As opposed to common training strategy on static datasets, in the case of PoMS it is not beneficial to normalise the training data before fitting the AE, as periodic additions to the collection lead to instability in the training.
        %
        %
        At each iteration of PoMS, the \textit{parameter manifold learning} phase uses all the policy parameters from the collection and continues the training of the AE to refine the latent representation.
        Further details on the training procedure are given in Appendix A.
        
    \vspace{-0.1cm}
    \subsection{Parameter Search Phase}
        One of the strengths of MAP-Elites is that it constantly applies small mutations to the ``elites" (the solutions contained in the grid).
        It is crucial to preserve this property in PoMS while performing search in the latent space. 
        However, a small perturbation in the latent space can lead to a very large perturbation in the parameter space due to the complexity of the learned decoder.
        Therefore, there is a significant risk that applying mutations directly in the latent space (e.g. via Gaussian noise) will lead to an uncontrolled mutation, similar to random search.  
        %
        %
        %
        %
        %
        %
        %
        %
    
        \textbf{Considering the decoder Jacobian}
        To address this issue, we propose to make the latent parameter space search heteroscedastic, as a function of the local structure. 
        We use the Jacobian of the decoder, which gives us a linear approximation (first order Taylor expansion) of the transformation around a specific point in the latent space $z$, denoted as $\mathbf{J}_{D}(z)$. 
        %
        %
        %
        %
        %
        The Jacobian allows us to impose that each mutated point in the latent space ${z}' \sim \mathcal{N}(z_k, \Sigma_{Z})$, when reconstructed, lands within a unit hyper-sphere centered aroung the reconstructed sample,
        in the original parameter space, i.e. $f_{D}(z') \sim \mathcal{N}(f_{D}(z_k), \sigma_{\Theta}\mathbb{I})$, as shown in Fig.~\ref{fig:schematics}. 
        %
        %
        %
        %
        %
        %
        %
        %
        %
        Let us define the desired covariance matrix of an isotropic unit Gaussian in the parameter space as $\Sigma_{\Theta} = \sigma_{\Theta}\mathbb{I}$, where $\sigma_{\Theta}$ is the desired radius of a hyper-sphere and $\mathbb{I} \in \mathbb{R}^{P\times P}$ is the identity matrix. 
        The objective is to estimate the appropriate Gaussian noise covariance matrix $\Sigma_{Z}$ applied in the latent space, as a function of $\Sigma_{\Theta}$ and the Jacobian, as $\Sigma_{Z} = \mathbf{J}_{D}^T \Sigma_{\Theta} \mathbf{J}_{D}$.
        %
        %
        %
        %
        %
        %
        For the full derivation using Taylor expansion, refer to Appendix C.

        \textbf{Mixing strategies for improving representation stability and diversity}
        %
        %
        Performing the parameter search in the latent space has the advantage of offering a smaller search space with a high-density of different and interesting policies. However, like in most autoregressive algorithms, the AE is unable to generalise far beyond the training set data support.
        To overcome this problem we employ a region-based exploration strategy. 
        %
        If the reconstruction error of a latent point is below a threshold value $\epsilon_{r}$, 
        we perform the mutation in the latent space as explained above. 
        Otherwise, the mutation is applied directly in the parameter space using $\mathcal{N}(\theta, \Sigma_{\Theta})$.
        The reconstruction error threshold $\epsilon_{r}$ is determined heuristically based on the average reconstruction error of all the points in the collection, after the policy manifold learning phase.
        The pseudocode for region-based policy search is given in Algorithm 2 of Appendix A.
        %
        %
        %
        Periodically adding solutions obtained via parameter space search helps reduce overfitting, thus helping the model generalise better.
        This can be regarded as a type of active learning based on model uncertainty~\cite{rakicevic2019informed}.
        %
        %


\vspace{-0.1cm}
\section{Evaluation}
\label{sec:evaluation}
    %
        

    	%
        We evaluate the algorithms on three tasks (Fig. 1 in Appendix D) in deterministic simulated environments. In each task, a robot is controlled by the policy which takes the full observation vector as input, and outputs desired actions (velocities and joint torques).
        We briefly describe the environments and the tasks, and defer further details to Appendix D.
        
        %
        
        \textbf{Striker [observation 14D, action 3D, total behaviours 15300]} 
        is a bounded air-hockey-like environment,
        with the goal of controlling the striker to hit the puck so it lands on as many diverse positions as possible.
        %
    	%
    	%
        \textbf{Bipedal-Walker [observation 26D, action 4D, total behaviours 12500]} is an OpenAI gym environment with the absolute location of the agent added to the observation, as it is used for the behaviour descriptors.
        The goal is to have the agent discover as many diverse gaits as possible. 
        %
        %
        %
        \textbf{Bipedal-Kicker [observation 26+4D, action 4D, total behaviours 10000]} extends the Bipedal-Walker task by adding a ball. 
        The goal is for the agent to kick the ball so it travels in diverse trajectories.
    	%
    	%
    	In addition to each environment, we introduce their \textbf{*-mix-scale} counterparts. The goal of this is to analyse the phenomenon when certain elements of the observation vector have different scales, which typically occurs in unbounded tasks.
    	%
    	%
        %
        %
        We use \textit{behaviour coverage}, as a metric to assess the search performance, as it quantifies distinct behaviours discovered by an algorithm.

    \vspace{-0.1cm}
    \subsection{Results}
    \label{sec:discussion}
        The results of `mix-scale' experiments are shown in Fig.~\ref{fig:results}, while the rest or the experimental results are in Fig. 2 of Appendix E, together with additional comparison details and analysis of the results.
        Our experimental evaluation aims to answer the following questions:
    \begin{figure}[tb!]
        \centering
        {\includegraphics[width=.98\textwidth]{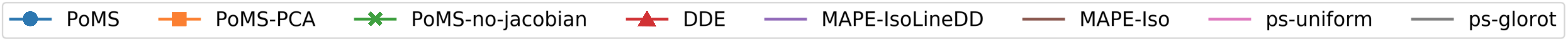}} 
        \\
        \subfloat[Bipedal-Walker mixed scale]{\includegraphics[trim=0 60 0 0, clip, width=0.31\textwidth]
        {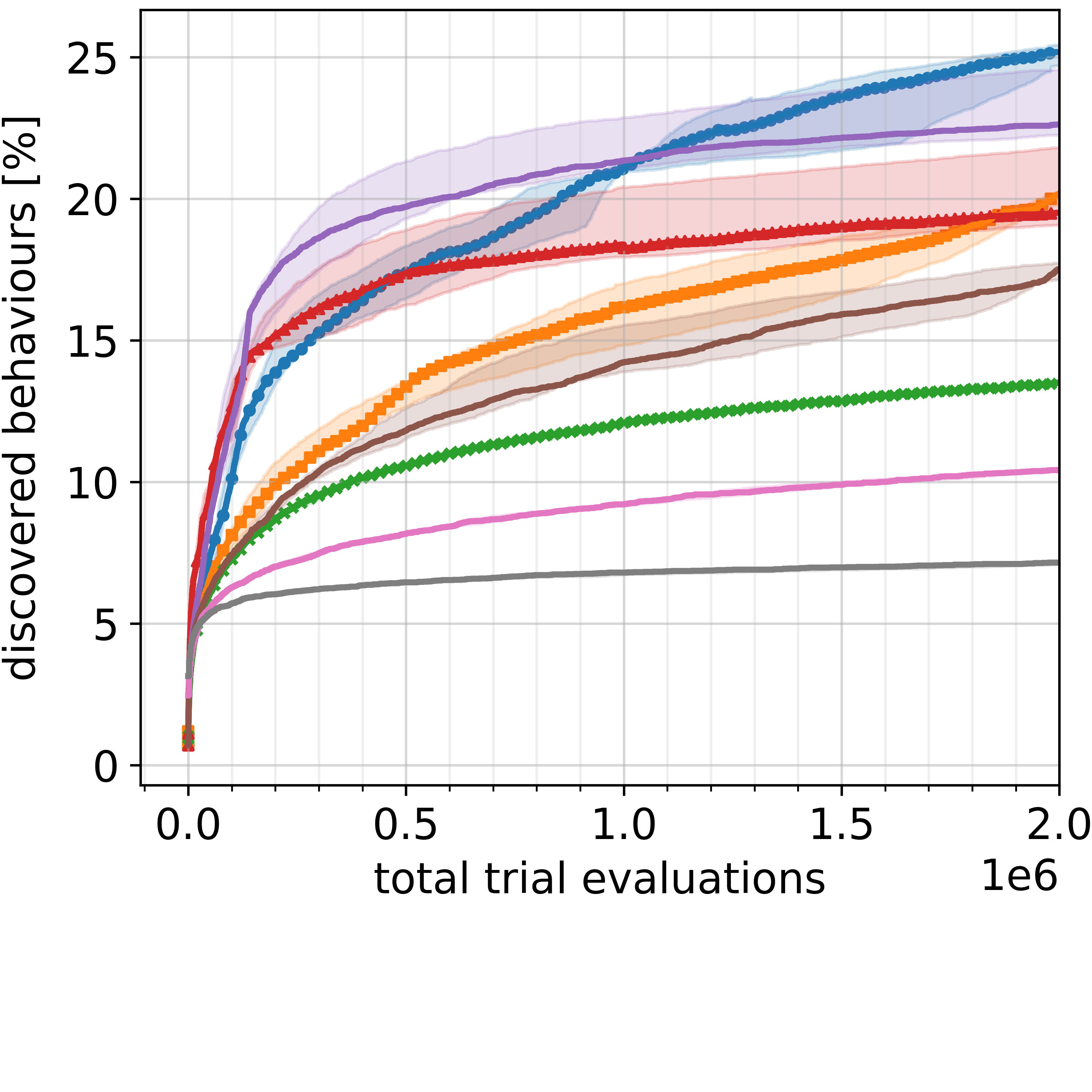}
        \label{fig:results:biped_walk_mix}} 
        \hspace{0.2cm}
        \subfloat[Bipedal-Kicker mixed scale]{\includegraphics[trim=0 60 0 0, clip, width=0.31\textwidth]
        {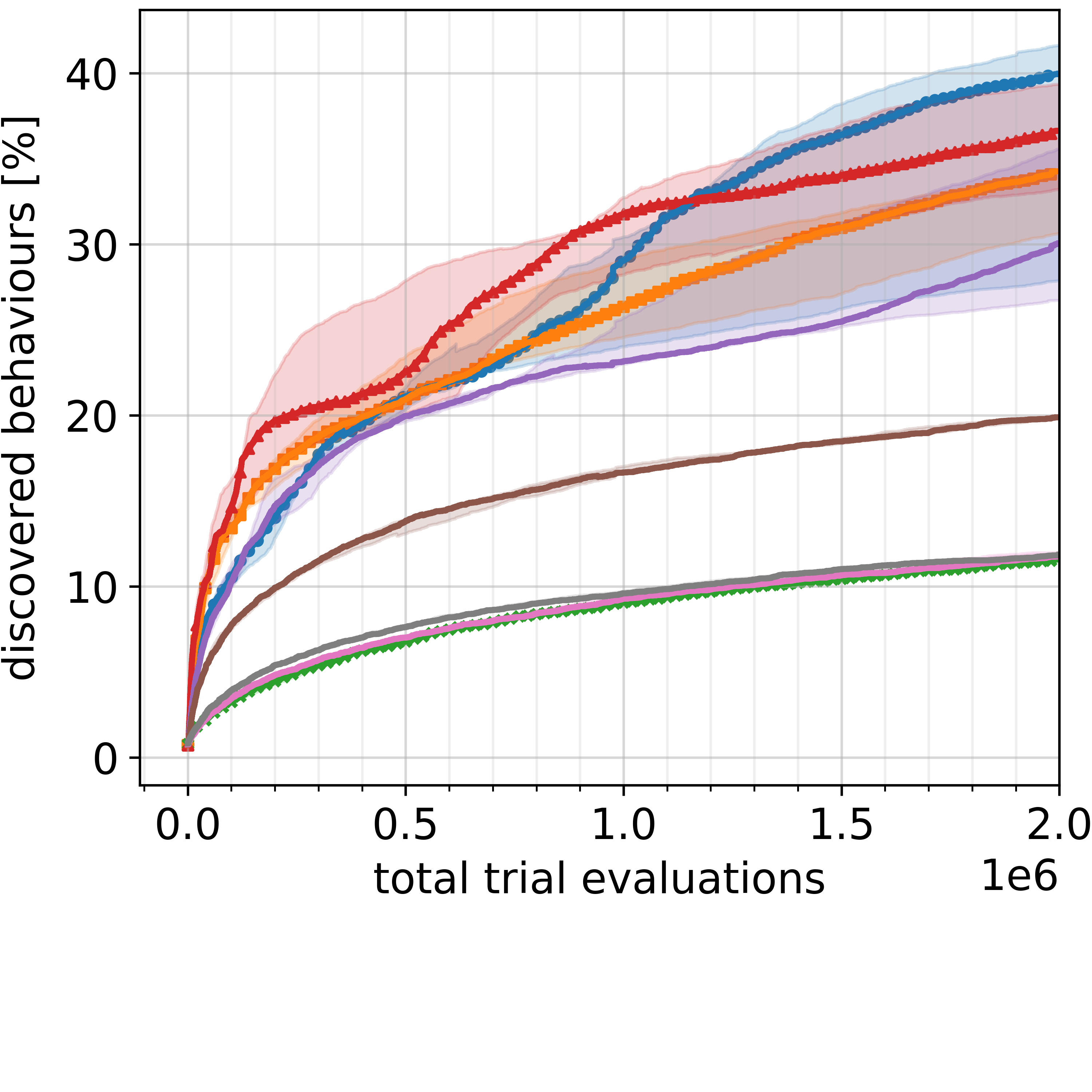}
        \label{fig:results:biped_kick_mix}} 
        \hspace{0.2cm}
        \subfloat[Striker mixed scale]{\includegraphics[trim=0 60 0 0, clip, width=0.31\textwidth]
        {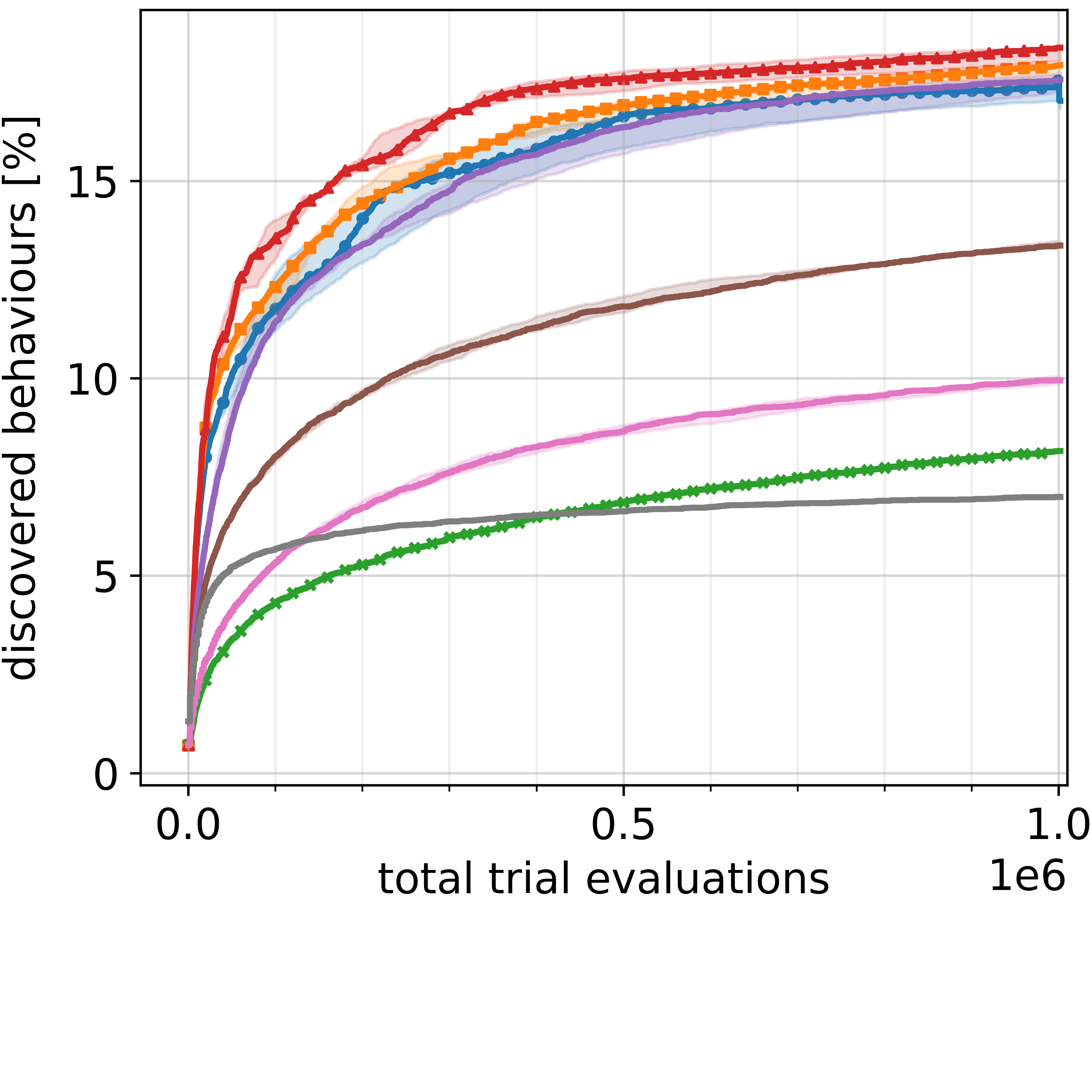}
        \label{fig:results:striker_mix}} 
    \caption{Behaviour coverage plots of the compared methods, achieved in three continuous control environments. Additional results for `normal' environments are in Fig. 2 of Appendix E.}
    \label{fig:results}
    \end{figure}
        
        \textbf{Q1. Are there benefits of using the learned latent space, over the original space, for policy search?}
            In order to answer Q1, we compare the performance of the proposed PoMS approach, to the standard MAP-Elites (MAPE-Iso) which performs search in the original parameter space.
            Although standard MAP-Elites shows competitiveness in high-dimensional parameter space problems, PoMS systematically achieves higher behaviour coverage across the tasks. The only exception is the Striker case, where methods converge to the same performance.
            As a `sanity check' we compare with two random search approaches in the parameter space, uniformly sampling policies (ps-uniform), and initialising according to Xavier-Glorot initialiser \cite{glorot2010understanding} (ps-glorot).
        \\
        \textbf{Q2. What is the effect of using the Jacobian of the decoder?}
            We perform an ablation study to demonstrate the importance of the Jacobian scaling of the latent sampling covariance matrix, by comparing PoMS to naive latent space search (PoMS-no-jacobian). The results show that not accounting for the Jacobian, achieves significantly worse behaviour, and leads to a search which is almost random.
        \\
        \textbf{Q3. Is the non-linear manifold learning necessary, or does linear projection suffice?}
            We examine the importance of non-linear manifold learning as opposed to using a linear projection of the parameter space via PCA (PoMS-PCA).
            This comparison highlights the intrinsic complexity of a given control problem, as locomotion tasks usually have an inherent non-linearity in the mapping of the policy outputs to the observation vector used for determining BDs, which is accounted for by PoMS.
        \\
        \textbf{Q4. How does PoMS compare to state-of-the-art QD approaches?}
            We consider two recently introduced MAP-Elites algorithms: 
            MAP-Elites with Line mutations~\cite{vassiliades2018discovering} (MAPE-IsoLineDD) and MAP-Elites with data-driven encoding~\cite{gaier2020discovering} (MAPE-DDE), as well as Diversity is All You Need (DIAYN) \cite{eysenbach2018diversity} from the deep RL literature.
            However, preliminary experiments showed that DIAYN is unable to scale to several thousands of skills, and takes significantly more computational time as it is not parallelizable, so we decided to exclude this algorithm from our analysis.
            The top performing methods, PoMS, PoMS-PCA, MAPE-DDE and MAPE-IsoLineDD, share the characteristic of focusing the search on a smaller subset of the parameter space, either as the hyper-volume of elites or via the learned manifold.
            %
            %
            This further solidifies our claims from Q1.
            The proposed PoMS approach outperforms the next best state-of-the-art approach by up to 5\%, except in the Striker tasks where it achieves 
            on-par performance at convergence.

\vspace{-0.1cm}
\section{Conclusion}
\label{sec:conclusion}
    In this paper, we proposed the Policy Manifold Search algorithm, that aims to discover a collection of policies with diverse behaviours. 
    %
    Our work is inspired by the manifold hypothesis, which assumes that the useful policies tend to concentrate near a low-dimensional manifold embedded in the original high-dimensional parameter space. 
    %
    Experimental evaluations of PoMS validate the benefits of using a learned manifold coupled with the Jacobian of the decoder for policy search to discover larger collections of diverse policies compared to the baselines.
    %
    

\section*{Broader Impact}

The motivation behind Policy Manifold Search is to open new avenues for future research on investigating representations of the parameter space and using its properties.
Investigating additional regularisation and manifold learning approaches, could provide a new hybrid optimisation paradigm when combined with gradient-based approaches.
In the presented form, PoMS integrated in the QD framework is useful for generating collections of diverse policies. 
Having such a collection is important for real-world continuously operating robots, which run in dynamically changing environment and are susceptible to damage.
If damaged, robots would still be able to perform their tasks by choosing more robust alternative policies from the collection, while waiting to be repaired.


\begin{ack}
Nemanja Rakicevic is funded by the President's PhD Scholarship from Imperial College
London.
\end{ack}


\printbibliography
\end{refsection}

   

\newpage


\begin{refsection}

\begin{appendices}

\section{Autoencoder training procedure and data initialisation}
    \vspace{-0.1cm}
    The Policy Manifold Search algorithm, as well as other MAP-Elites-based baselines with latent space used for comparison, run in loops. 
    Each loop consists of a parameter search phase and parameter manifold learning phase (as shown in Algorithm \ref{algo:poms}).
    The parameter search phase consists of 100 MAP-Elites iterations, and each iteration has a budget of 200 policy samples.

        \begin{minipage}{1\textwidth}
        \footnotesize
            \begin{algorithm}[H]
                \label{algo:poms}
                \caption{Policy Manifold Search}
                \SetAlgoLined
                \DontPrintSemicolon
                \For{nloop \textbf{in} PoMS\_loops}{
                    \tcp{parameter search phase}
                    \For{niter \textbf{in} MAP-Elites\_iterations}{
                        \eIf{nloop == 0 \textbf{and} niter == 0}{
                            $\bm{\theta}^{SEL} \sim \mathcal{U}(-1,1)$ \;
                        }{
                            $\bm{\theta}^{SEL} \sim \mathrm{C}_{\bm{\theta}}$ \;
                            $\bm{\theta}^{MUT} = \text{mutation\_operator}(\bm{\theta}^{SEL}, f_{E}, f_{D})$ \;
                        }
                        $\bm{\tau}^{MUT} = \text{environment\_eval}(\bm{\theta}^{MUT})$ \;
                        $\bm{b}^{MUT} = \mathcal{BD}(\bm{\tau}^{MUT})$ \;
                        $\mathrm{C}_{\bm{\theta}}[\bm{b}^{MUT}] \leftarrow \bm{\theta}^{MUT}$ \;
                    }
                    \tcp{parameter manifold learning phase}
                    \For{ $\bm{\theta}_{\text{batch}}$ \textbf{in} $\mathrm{C}_{\bm{\theta}}$}{
                        $\hat{\bm{\theta}}_{\text{batch}} = f_{D} \circ f_{E}(\bm{\theta}_{\text{batch}}; \xi)$ \;
                        $\argmin_{\xi} \mathcal{L}_{AE} = \left \| \bm{\theta}_{\text{batch}} -  \hat{\bm{\theta}}_{\text{batch}} \right \|$ \;
                    }
                }
            \end{algorithm}
        \end{minipage}
    
        \begin{minipage}{1\textwidth}
        \footnotesize
            \begin{algorithm}[H]
            \label{algo:mix_region}
            \caption{Region-based search}
            \SetAlgoLined
            \DontPrintSemicolon
            \textbf{Input:} $\mathrm{C}_{\bm{\theta}}$, $\Sigma_{\Theta}$  \;
            \textbf{Output:} $\mathrm{C}_{\bm{\theta}}$  \;
            \For{$\text{num\_loops}$}{
                $\epsilon_{recn} = \frac{1}{|\mathrm{C}_{\bm{\theta}}|}\sum_i^{|\mathrm{C}_{\bm{\theta}}|} \left \| \theta_i -  \hat{\theta}_i \right \|$ \;
                \For{$\text{map-elites\_iterations}$}{
                    $\bm{\theta}^{SEL} \sim \mathrm{C}_{\bm{\theta}}$ \;        
                    $\bm{\theta}^{MUT} = \emptyset$\;
                    \For{$\theta_i$ \textbf{in} $\mathbf{\theta}^{SEL}$}{
                        \eIf{$\left \| \theta_i - \hat{\theta}_i \right \| < \epsilon_{recn} $}{
                            $z_i = f_{E}(\theta_i)$ \;        
                            $\mathbf{J} = \text{Jacobian}(f_{D}; z_i)$ \;
                            $\Sigma_{\Phi} = \mathbf{J}^T \Sigma_{\Theta} \mathbf{J}$ \;
                            $z_i^{MUT} = z_i + \mathcal{N}(0, \Sigma_{Z})$ \;
                            $\theta_i^{MUT} = f_{D}(z_i^{MUT})$ \;
                        }{
                            $\theta_i^{MUT} = \theta_i + \mathcal{N}(0, \Sigma_{\Theta})$ \;
                        }
                        $\bm{\theta}^{MUT}  \leftarrow \theta_i^{MUT}$\;
                    }
                    $\mathrm{C}_{\bm{\theta}} \leftarrow \text{unique\_bd}( \mathrm{C}_{\bm{\theta}} \bigcup \bm{\theta}^{MUT})$ 
                }
            }
            \end{algorithm}
        \end{minipage}

    \subsection{Initialising the policy collection}
        In the first MAP-Elites iteration of the first loop, the policy collection is still empty $\mathrm{C}_{\bm{\theta}} = \emptyset$. Therefore, it is not possible to select a sample to be mutated from the collection, so we generate the initial sample by drawing 2000 policy samples from a uniform distribution $\bm{\theta}^{SEL} \sim \mathcal{U}(-1,1)$, as usually done in the MAP-Elites literature \cite{cully2017quality}.
        This is done for all MAP-Elites-based baselines.

    \subsection{Autoencoder training}
        In order to get the latent representation of the parameter space, PoMS uses a fully-connected symmetrical Autoencoder. The specific architecture varies based on the experiment, and these are presented in detail in Appendix \ref{sec:exp:hyper}.
        During training, the reconstruction loss $\mathcal{L}_{AE}$ is minimised in order to find the optimal AE parameters $\xi$, using the Adam optimiser with parameters: $\beta_1=0.9$, $\beta_2=0.999$, $\varepsilon=10^{-8}$ and learning rate of $10^{-5}$.
        The training is ran for $2 \cdot 10^{4}$ epochs with batch size of 64. 
        These parameters are fixed for all the experiments.
        
        To improve the robustness of the optimiser, we reset the momentum variables at every loop.
        Moreover, 30\% of each batch is used as a test set for early stopping of the training. If the slope of the line fitted to the last 100 test set values is larger than $10^{-5}$, the training is stopped. We found that this improves the generalisation of the AE and reduces training time.

    \subsection{MAP-Elites algorithm}
        The MAP-Elites framework maintains a collection of multiple policy parameters, in a multi-dimensional cell-grid $\mathrm{C}_{\bm{\theta}}$, which is indexed by the behaviour index $b \in \mathcal{B}$ obtained using the behaviour descriptor.
        To resolve the surjective mapping, usually each cell of the grid is populated by a single highest-performing policy based on some performance metric.
        %
        The aim of MAP-Elites is to fill all the cells of $\mathrm{C}_{\bm{\theta}}$ with the best possible policies through an iterative process. 
        Each iteration consists of (i) randomly selecting a batch of individuals from the collection, (ii) applying a mutation and evaluating these modified individuals, (iii) based on the outcome of the evaluation, add the new individuals to the grid if the corresponding cell is vacant or if they outperform the currently occupying individual.
        In this study we focus purely on policy behaviour diversity, so the performance metric is replaced by random selection.
        The most common mutation operator adds an isotropic Gaussian noise $\mathcal{N}(0, \Sigma_{\Theta})$, with a unit covariance matrix $\Sigma_{\Theta} = \sigma_{\Theta}\mathbb{I}$ where $\sigma_{\Theta}$ is a hyperparameter.


\section{Related Work}
\label{sec:related_work}
    
    In this section we position our work within the diversity-based and representation learning literature.
    
    \textbf{Quality Diversity}
        (QD) algorithms \cite{pugh2016quality, cully2017quality} have been recently introduced as a framework that generates a collection of high-performing and diverse solutions. 
        Most popular approaches are Multi-dimensional Archive of Phenotypic Elites (MAP-Elites) \cite{cully2015robots} and Novelty Search with Local Competition \cite{lehman2011evolving}, which differ in how they select and maintain a collection of behaviours.
        In this work, we focus on MAP-Elites due to its simplicity of implementation and proven performance in various applications, such as video games \cite{gravina2019procedural}, robotics \cite{cully2015robots}, routing problems \cite{urquhart2019illumination} etc.
        MAP-Elites usually uses simple low-dimensional parameterised controllers, such as periodic function generators, central pattern generators, small evolved networks or low-level controllers \cite{cully2015robots}.
        Recently, \cite{colas2020scaling} proposed to scale MAP-Elites to Deep Neuroevolution and applied it to more complex environments. The policy search process is performed directly in the high-dimensional NN parameter space.
    	
    \textbf{Diversity of policies in Deep RL}
        is usually considered as a stepping stone for improving exploration, rather than for maintaining a collection of solutions.
        Such diversity is achieved via action-space noise \cite{schulman2017proximal}, parameter-space noise \cite{plappert2017parameter, fortunato2017noisy}, or enforced via additional entropy \cite{ziebart2008maximum, haarnoja2018soft}, intrinsic motivation \cite{pere2018unsupervised, badia2020nevergiveup} or state-visitation \cite{bellemare2016unifying, tang2017exploration} terms in the reward function.
        %
        Diversity is also essential in hierarchical RL, which uses a policy network conditioned on a task sampled from a discrete \cite{hausman2018transferable, eysenbach2018diversity} or continuous \cite{sharma2019dynamics} distribution.
        Maximum Entropy (MaxEnt) RL framework \cite{gregor2017variational, achiam2018variational} focuses on maximising a state or action distribution entropy term added to the reward function, which incentivises the discovery novel state-trajectories, i.e. behaviours.
        The methods in \cite{eysenbach2018diversity, sharma2019dynamics} focus purely on discovering diverse skills, i.e. behaviours, and do not consider extrinsic environment rewards.
        Discovery of diverse state-trajectories is encouraged by maximising the mutual information between states and skills, in addition to maximizing the state entropy.
        
        While QD and MaxEnt RL frameworks both aim to promote behaviours diversity, they differ in how the diversity is maintained and in the definition of skills, i.e. behaviours.
        MaxEnt RL generates a single, task-conditioned policy, that exhibits different behaviours depending on the task, while QD learns one policy for each of the behaviours.
        In QD, behaviours are enumerated and quantifiable via behaviour descriptors derived from the state-trajectory \cite{cully2017quality} or learned \cite{meyerson2016learning, cully2019autonomous}, while MaxEnt RL methods quantify behaviour diversity by evaluating the entropy of the task context distribution \cite{gregor2017variational, eysenbach2018diversity, achiam2018variational, hausman2018transferable}. This allows QD to consider a significantly larger number of diverse behaviours.
	
    \textbf{Manifold Learning} 
        is the process of obtaining a lower-dimensional representation, i.e. manifold, embedded in the original high-dimensional input space. 
        The \textit{manifold hypothesis} assumes that a high density of datapoints is located in the vicinity of the manifold.
        This notion has been thoroughly examined as \textit{representation learning} in machine learning \cite{bengio2013representation} and RL \cite{chen2018active, eysenbach2018diversity, rusu2018meta}, with important insights on how to exploit the structure of the manifold to improve robustness \cite{rifai2011contractive, rifai2012generative}. These approaches deal with input spaces such as environment observations, images, graphs etc. which have different structural properties compared to NN parameter spaces.
        Recently, the concept of manifold learning, or \textit{intrinsic dimension} of the parameter space, has been examined for NNs \cite{li2018measuring} and specifically NN-parameterised policies \cite{richemond2019biologically, jegorova2018generative, chang2018agent}.
        %
    	To the best of our knowledge, \cite{chang2018agent} is the only work besides ours that learns representations of NN policy parameters. They generate new policies via simple interpolation within the latent space without considering its structure. 
    
        Two recently introduced MAP-Elites based approaches present ideas which are close to the notion of parameter manifolds.
        In \cite{vassiliades2018discovering} the authors examine the idea of hypervolumes in the high-dimensional parameter space containing high-performing and diverse solutions, while \cite{gaier2020discovering} exploits the reconstruction error as the perturbation needed for policy search. 
        %
        %
        %


\section{Derivation of Jacobian scaling}

    Let us assume $\theta \in \Theta \subset \mathbb{R}^P$ and $z \in Z \subset \mathbb{R}^M$ to be Gaussian in the parameter and latent space respectively, such that $\theta \sim \mathcal{N}(\mu_{\Theta}, \Sigma_{\Theta})$ and $z \sim \mathcal{N}(\mu_Z, \Sigma_Z)$.
    
    We assume a non-linear, vector-valued function $f_{D}: \mathbb{R}^M \rightarrow \mathbb{R}^P$ which maps the latent space to the original parameter space.
    
    We can get a linear approximation  $\hat{f}_{D}$ in a point $\mu_Z$ by performing a first-order Taylor expansion:
    \begin{equation}
    \begin{split} 
        \hat{f}_{D}(z)  & \approx f_{D}(\mu_z) + 
                                 \sum_i^M \frac{\partial f_{D}(z)}{\partial z_i} \biggr\rvert_{z=\mu_{Z}} 
                                 (z_i - \mu_{Zi}) \\
                        & = f_{D}(\mu_Z) + 
                                \begin{bmatrix}
                                    \nabla f_{D}(z)_1\\
                                    \nabla f_{D}(z)_2\\
                                    ...\\
                                    \nabla f_{D}(z)_P
                                \end{bmatrix}
                                (z - \mu_{Z}) \\
                        & = \mu_{\Theta} + \mathbf{J}_{D}(z - \mu_{Z})
    \end{split}
    \end{equation}
    where $\mathbf{J}_{D}$ is the Jacobian matrix of $f_{D}$ at $\mu_Z$:
    \begin{equation}
        \mathbf{J}_{D}(\mu_Z) = J_{D}(\mu_Z)_{ij} = \frac{\partial f_{D}(\mu_Z)_i}{\partial \mu_{Z_j}}
    \end{equation}
    where indices $i$ and $j$ refer to the corresponding elements of the reconstructed or latent parameter vector, respectively.
        
    Further, if $\hat{\theta} = \hat{f}_{D}(z)$ its expected value can be obtained as:
    \begin{equation}
    \begin{aligned}
        \mathbb{E}[\hat{\theta}] & = \mathbb{E} \left [ \mu_{\Theta} + \mathbf{J}_{D}(z - \mu_{Z}) \right ] &\\
                       & = \mathbb{E}[\mu_{\Theta}] + \mathbb{E} \left [\mathbf{J}_{D}(z - \mu_{Z}) \right ] & \text{(expected values of a sum is the sum of expected values)} \\
                       & = \mu_{\Theta} + \mathbb{E} \left [\mathbf{J}_{D} z \right ]  - \mathbb{E} \left [\mathbf{J}_{D} \mu_{Z} \right ] & 
                       \text{(expectation of a constant is a constant)} \\    
                       & = \mu_{\Theta} + \mathbf{J}_{D} \mathbb{E} [z] - \mathbf{J}_{D} \mathbb{E} [\mu_{Z}] & \\   
                       & = \mu_{\Theta} + \mathbf{J}_{D} \mu_{Z} - \mathbf{J}_{D} \mu_{Z} &\\ 
                       & = \mu_{\Theta}   
    \end{aligned}
    \end{equation}

    We can obtain $\Sigma_{\Theta}$ based on $\Sigma_Z$. We start with the standard equation for covariance:
    \begin{equation}
    \begin{aligned}
        \Sigma_{\Theta} & = \mathbb{E} \left [(\hat{\theta} - \mathbb{E}[\hat{\theta}])(\hat{\theta} -                               \mathbb{E}[\hat{\theta}])^T \right ] &\\
                        & = \mathbb{E} \left [(\mu_{\Theta} + \mathbf{J}_{D}(z - \mu_{Z}) - \mu_{\Theta})
                            (\mu_{\Theta} + \mathbf{J}_{D}(z - \mu_{Z}) - \mu_{\Theta})^T \right ] & 
                            \text{(using (1) and (2))} \\
                        & = \mathbb{E} \left [(\mathbf{J}_{D}(z - \mu_{Z}))
                            (\mathbf{J}_{D}(z - \mu_{Z}))^T \right ] &\\
                        & = \mathbb{E} \left [ \mathbf{J}_{D}(z - \mu_{Z})
                            (z - \mu_{Z})^T \mathbf{J}_{D}^T \right ] &\\
                        & = \mathbf{J}_{D} \mathbb{E} \left [ (z - \mu_{Z})
                            (z - \mu_{Z})^T \right ] \mathbf{J}_{D}^T & 
                            \text{(covariance definition for z)} \\
                        &= \mathbf{J}_{D} \Sigma_Z \mathbf{J}_{D}^T
    \end{aligned}
    \end{equation}
    
    By rearranging the previous equation we get:
    
    \begin{equation}
        \Sigma_Z = \mathbf{J}_{D}^T \Sigma_{\Theta} \mathbf{J}_{D}
    \end{equation}
    


\section{Experiment Details}

    \subsection{Environement details}
    
        In this section we give implementation details of the environments used in the experiments, including the observation and action vectors used within the policy.
        
\begin{figure}[tb!]
            \subfloat[Bipedal-Walker\label{fig:envs:biped-walk}]
            {\includegraphics[width=0.3\textwidth]{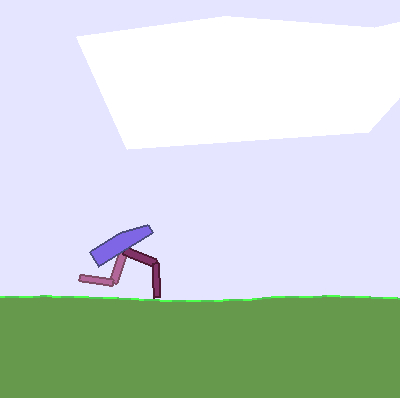}} 
            \hspace{0.31cm}
            \subfloat[Bipedal-Kicker\label{fig:envs:biped-kick}]
            {\includegraphics[width=0.3\textwidth]{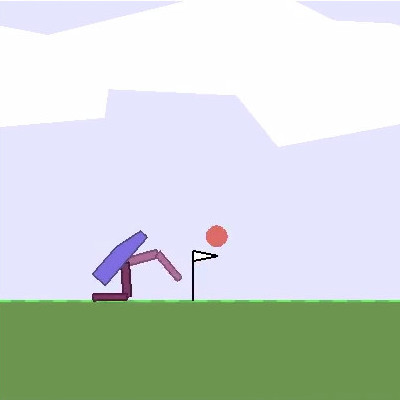}}
            \hspace{0.31cm}
            \subfloat[Striker\label{fig:envs:striker}]
            {\includegraphics[width=0.3\textwidth]{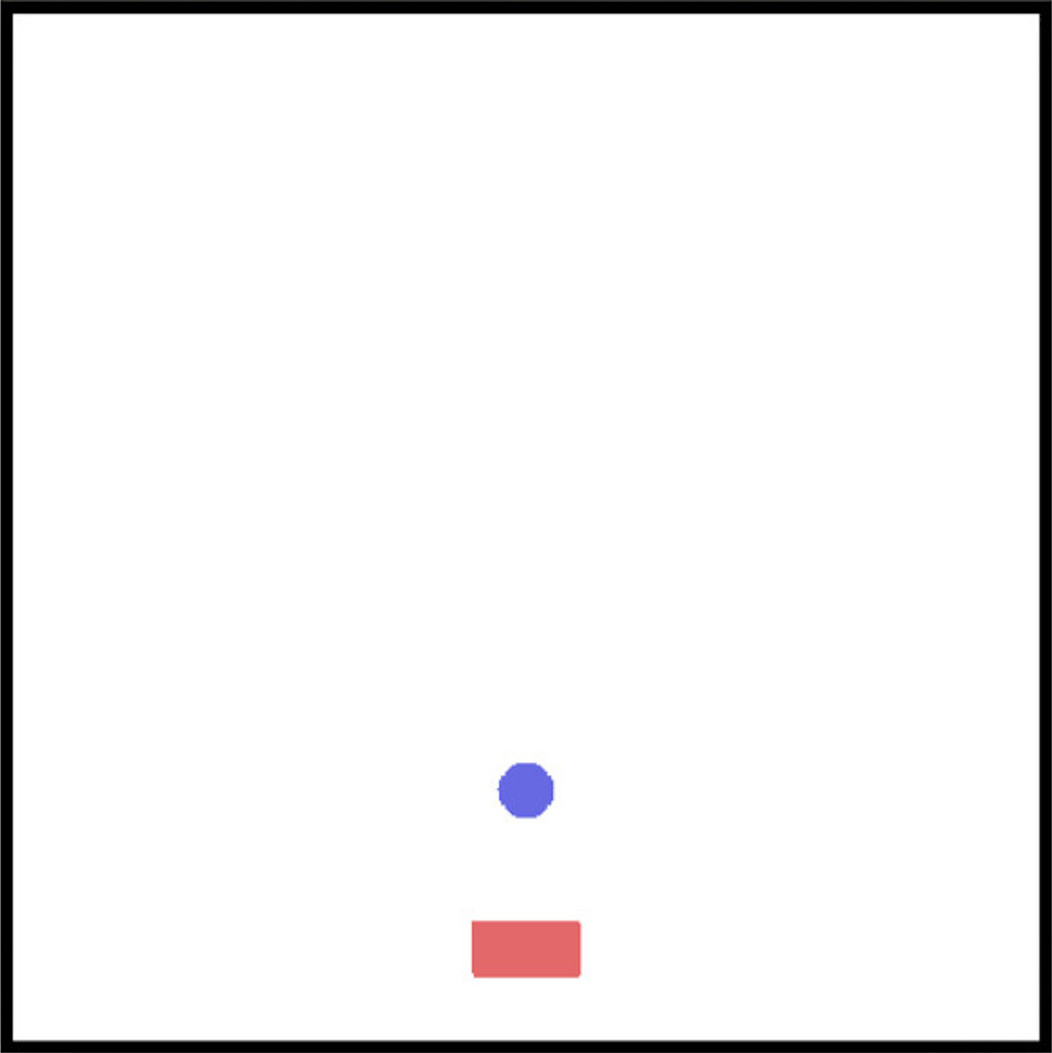}} 
        \caption{Screenshots of the three continuous control tasks used for experimental evaluation. The environments are implemented in Box2D \protect\cite{catto2011box2d}.}
        \label{fig:envs}
        \end{figure}

        \textbf{Striker} (Fig. \ref{fig:envs:striker})
        is an air-hockey-based environment implemented in Box2D \cite{catto2011box2d}. 
        The arena is bounded by four walls to the size of 100x100 units, created to be proportional to the striker size (5x2.5 units). The puck has a radius of 2.5 units.
        The input to the policy is a 14D observation vector, consisting of striker's x, y position and angle $\phi$, the puck's x, y position, as well as their corresponding velocities, and puck-wall distances for each of the walls.
        The output of the policy is a 3D action vector that controls the striker's x, y and rotational velocities.
        The striker is allowed to move for 100 episodes and then stops to avoid continuous interaction with the puck, while the puck moves until it stops due to the damping effects.
        We define the behaviour descriptor as a 3D grid: 
        D1-2: final x,y position of the puck, with 30 bins for each dimension;
        D3: wall(s) the puck bounced off during the trial, and has one of 17 possible values (no wall, south, east, north, west, and second order combinations).

        \textbf{Bipedal-Walker} (Fig. \ref{fig:envs:biped-walk}) is a standard OpenAI gym \cite{brockman2016openai} Box2D environment.
        The original observation vector has 24 elements, which include the robot hull angle, horizontal, vertical and angular velocities, joints angles and angular velocities, legs-ground contact information, and 10 lidar rangefinder measurements. 
        However, we also added the absolute coordinates of the robot hull, thus creating a 26D observation vector.
        The 4D action vector is unaltered and provides torques for each of the leg joints.
        At the start of the episode, the robot is placed in the middle of the terrain so it can walk either forward or backward in as many diverse ways as possible.
        The episode is limited to 500 steps.
        The 4D behaviour descriptor grid is based on the agent's absolute positions and leg-ground contacts during the episode: 
        D1: average hull y-coordinate, 5 bins;
        D2: final hull x-coordinate, 100 bins;
        D3-4: proportion of time left and right legs spent in contact with ground, 5 bins each. 
        
        \textbf{Bipedal-Kicker} (Fig. \ref{fig:envs:biped-kick}) extends the Bipedal-Walker task by adding a ball. Therefore, the observation vector is extended with the ball x, y position and velocities, making it 30D. The action output is the same.
        Since the goal is to have a diversity of ball ballistic trajectories, we make the terrain flat to avoid biasing the outcomes to local valleys.
        In order to facilitate kicking, as the agent does not have a foot, at the start of the episode the ball is dropped from a small height so the agent can hit it.
        Similarly to Striker, the agent is allowed to act 100 steps before the actions are set to 0, in order to have only one kick per episode.
        The behaviour descriptor is a 2D based grid, based on the ball trajectory, as a usual way of defining a 2D ballistic trajectory:
        D1: final ball x-coordinate, 200 bins;
        D2: maximum ball y-coordinate achieved during the episode, 50 bins.
        
    	In contrast to Bipedal-Walker, the Bipedal-Kicker and Striker tasks contain a ball, which is an external object manipulated by the agent. This adds complexity to the task as certain elements of the observation vector can vary independently of the agent's actions. 
    	Another distinction we note is between Striker which is bounded, and Bipedal-* environments which are unbounded. This causes certain elements of the observation vector to have different scale than the rest. Usually the difference in scale can be normalised, but in unbounded environments this is not straightforward.
    	We introduce this distinction in the environments in order to evaluate its influence on the behaviour diversity.

    	To emphasise this phenomena, we implement two versions of each of the environments:
    	We extend the Striker described above to create \textbf{Striker mix-scale} by scaling elements of the observation vector which are external to the agent (exteroceptive sensory input such as: ball position and velocity, and striker absolute position) by a factor of 100.
    	To get \textbf{Bipedal-Walker mix-scale}, we multiply only the agents absolute position elements by 100, as there is no ball. This creates an interesting example that has only two scaled elements in the 14D observation vector, while this ratio is much larger in the other environments.
    	Since the original Bipedal-Kicker has more unbounded elements than Bipedal-Walker and Striker, we consider this to be the \textbf{Bipedal-Kicker mix-scale} version. For the `normal' version we normalise the external elements of the original version to the [0, 1] range by arbitrarily defining the limits in the environment.
    	%
    	%
    	We hypothesise that having mixed scales of elements in the observation vector is propagated further to the agent policy parameters, and exploiting this structure in the parameter space affects the performance of the policy search algorithms.

    \subsection{Agent policy implementation}
        The policy of the agent is kept the same for all environments and is inspired by the Proximal Policy Optimisation \cite{schulman2017proximal} policy implementation for continuous control tasks. 
        The policy is defined as a fully-connected neural network with two hidden layers of 32 neurons each, with tanh activation functions. The output layer has a linear activation function. The input and output size vary depending on the observation and action vector sizes, which are specific to each environment.

    \subsection{Metric calculation}
        As described in Section 3, we evaluate the algorithms on three tasks in a deterministic simulated environment.
        In order to achieve determinism in the metric, e.g. surjective mapping between the agent policy and the corresponding behaviour descriptor, we enable repeatability by fixing the environment initial state as customary in the QD literature \cite{cully2015robots, mouret2015illuminating, gaier2020discovering}.
        
        Alternatives to a deterministic behaviour framework are presented in \cite{flageat2020fast, colas2020scaling, justesen2019map}. We will consider in future works how these approaches can be combined with PoMS. 
         
        For the experiments to have statistical significance, each run is executed with 5 different seeds for parameter initialisation.
        For each experiment, we show the median of the 5 seeds and 25th and 75th percentile. 
        The x-axis of the behaviour coverage plots shows the total cumulative number of episode rollouts.
    
    \subsection{Algorithm hyperparameters}
    \label{sec:exp:hyper}
    
        Here we describe the hyperparameters used in each of the algorithms, for different experiments.
        All algorithms based on MAP-Elites (PoMS versions, MAPE-Iso, MAPE-IsoLineDD and DDE) run 100 iterations of MAP-Elites with a budget of 200 samples during the parameter search phase. 
        The ps-uniform and ps-glorot versions run for the same total amount of samples as other algorithms, while the progress is saved every 2000 samples.
    
        \textbf{PoMS}\\
            The proposed PoMS algorithm has 3 main hyperparameters that need to be tuned: AE architecture, latent space dimension (LD) and $\Sigma_{\Theta}$. 
            The AE is symmetric, i.e. both encoder and decoder have one hidden layer with ELU activation, and we vary the number of nodes (HD). The activation function of the bottleneck layer forming the latent space is linear, same as for the output layer of the decoder.
            We provide the values used for each of the experiments in the table below:
            %
            
            \begin{tabular}{ |p{5cm}|p{2cm}|p{2cm}|p{2cm}|  }
                \hline
                \textbf{Environment}        & \textbf{HD}   & \textbf{LD}   & $\bm{\Sigma}_{\Theta}$ \\
                \hline
                Striker                     & 100           & 50            & 0.1    \\
                Striker mix-scale           & 100           & 20            & 0.01   \\
                \hline
                Bipedal-Walker              & 100           & 50            & 0.1    \\
                Bipedal-Walker mix-scale    & 100           & 100           & 0.1    \\
                \hline
                Bipedal-Kicker              & 100           & 100           & 0.01   \\
                Bipedal-Kicker mix-scale    & 100           & 50            & 0.01   \\
            \hline
            \end{tabular}

        \textbf{PoMS-no-jacobian}\\
            The PoMS-no-jacobian has the same hyperparameters as PoMS for each of the experiments, in order to do the proper ablation study.
            Since $\Sigma_{Z}$ is not obtained using the Jacobian scaling, we do not use a fixed matrix, rather, we dynamically update $\Sigma_{Z}$ based on the current manifold representation.
            We consider the ranges of the latent space parameters per latent dimension $r_{Z}$, in order to scale a unit covariance matrix $\Sigma_{Z} = r_{Z}^T\mathbb{I}$.
            This results in a search which scales the importance of each of the latent dimensions based on its range. 
            As we can see from the results, two issues arise with this approach: (i) range does not equal importance (solution density), (ii) applying the inverse transformation applies an additional distortion which can lead to undesirable values, because
            $\theta' \neq f_{D}(z')$, where $ z' \sim \mathcal{N}(\mu_Z, \Sigma_{Z})$ and $ \theta' \sim \mathcal{N}(f_{D}(\mu_Z), \Sigma_{\Theta})$, even if $\Sigma_{\Theta} = \Sigma_{Z} = \mathbb{I}$.

        \textbf{PoMS-PCA}\\
            The PoMS-PCA needs the latent dimension (LD) and $\Sigma_{\Theta}$ hyperparameters. These values are kept the same as in the corresponding PoMS version, in order to perform a proper ablation study.

        \textbf{DDE}\\
            We separate DDE hyperparameters into AE architecture hyperparameters and mutation operator specific hyperparameters.
            The former are kept the same as architectures of PoMS for each of the experiments, while the latter are kept the same as in the original paper \cite{gaier2020discovering}.
            Instead of running a fixed window for the multi-armed bandit upper confidence bound operator selector, we maintain a moving average.

        \textbf{MAPE-IsoLineDD}\\
            This algorithm has two main hyperparameters related to the weighing of the isometric and directional components of the mutation operator, and they are kept the same as in the original paper~\cite{vassiliades2018discovering}.

        \textbf{MAPE-Iso}\\
            The standard MAPE-Iso algorithm has only $\Sigma_{\Theta}$ that needs tuning. We set this to $\Sigma_{\Theta}$=0.1 as this achieved the best performance for MAPE-Iso across the experiments.

        \textbf{ps-uniform}\\
            The performance of ps-uniform changes based on the range from which the policy parameters are sampled. We keep this range to [-1, 1].
            It is important to mention that we examined other symmetric ranges as well, such as: [-0.1, 0.1], [-10, 10], [-100, 100], [-1000, 1000]. 
            The final performance was the best when using [-1, 1], as the values sampled in this region are diverse enough for the tanh activation in the policy network to propagate non-saturated values. Although this would be expected for [-0.1, 0.1] as well, the obtained parameters were not diverse enough.

        \textbf{ps-normal}\\
            There are no hyperparameters to tune for this baseline.


\section{Evaluation details and discussion}
    \subsection{Evaluation}
    
        Here we provide further details on the experiment organisation and the methods used for comparison, based on the main four questions:
        
        \textbf{Q1.} Are there benefits of using the learned latent space, over the original space, for policy search? \\
        \textbf{Q2.} What is the effect of using the Jacobian of the decoder? \\
        \textbf{Q3.} Is the non-linear manifold learning necessary, or does linear projection suffice? \\
        \textbf{Q4.} How does PoMS compare to state-of-the-art QD approaches?
        
        In order to answer Q1, we compare the performance of the proposed PoMS approach, to the standard MAP-Elites (MAPE-Iso) which perform search in the original parameter space. 
        
        We perform an ablation study to address Q2 and Q3. The first aim is to demonstrate the importance of the Jacobian scaling of the latent sampling covariance matrix. The alternative approach would be a naive latent space search where the latent sampling covariance matrix $\Sigma_{\Theta}$ is determined by the current ranges of the latent representations (PoMS-no-jacobian). 
        The second aim is to examine the importance of non-linear manifold learning (AE) as opposed to using a linear projection of the parameter space via PCA (PoMS-PCA).
        
        %
        %
        Regarding Q4, we consider two recently introduced MAP-Elites algorithms: 
        MAP-Elites with Line mutations~\cite{vassiliades2018discovering} (MAPE-IsoLineDD) that uses a mix of isotropic and directional Gaussian operators allowing for an adaptive search that implicitly explore the hyper-volume of the elites, which is similar to the learned manifold in PoMS, but in the original parameter space.
        Also, MAP-Elites with data-driven encoding~\cite{gaier2020discovering} (MAPE-DDE) that combines the line mutation with a reconstruction-crossover operator based on an AE trained on the policies contained in the collection, like PoMS. The AE hyperparameters of MAPE-DDE are set equal to those of PoMS in the experiments.
        Additionally, we look at the Diversity is All You Need (DIAYN) algorithm \cite{eysenbach2018diversity} from the deep RL literature, as it aims to maximise the skill diversity of a single policy conditioned on a discrete latent distribution. However, preliminary experiments showed that it is unable to scale to several thousands of skills (like MAP-Elites and PoMS). In the time required by MAP-Elites and PoMS to perform several millions of iterations, we only managed to run DIAYN over 2000 iterations, without observing any promising results. Therefore, we decided to exclude this algorithm from our analysis.
        
    \begin{figure}[tb!]
        \centering
        {\includegraphics[width=.98\textwidth]{Images/results/legend_poms.png}} 
        \\
        \subfloat[Bipedal-Walker]{\includegraphics[width=0.32\textwidth]
        {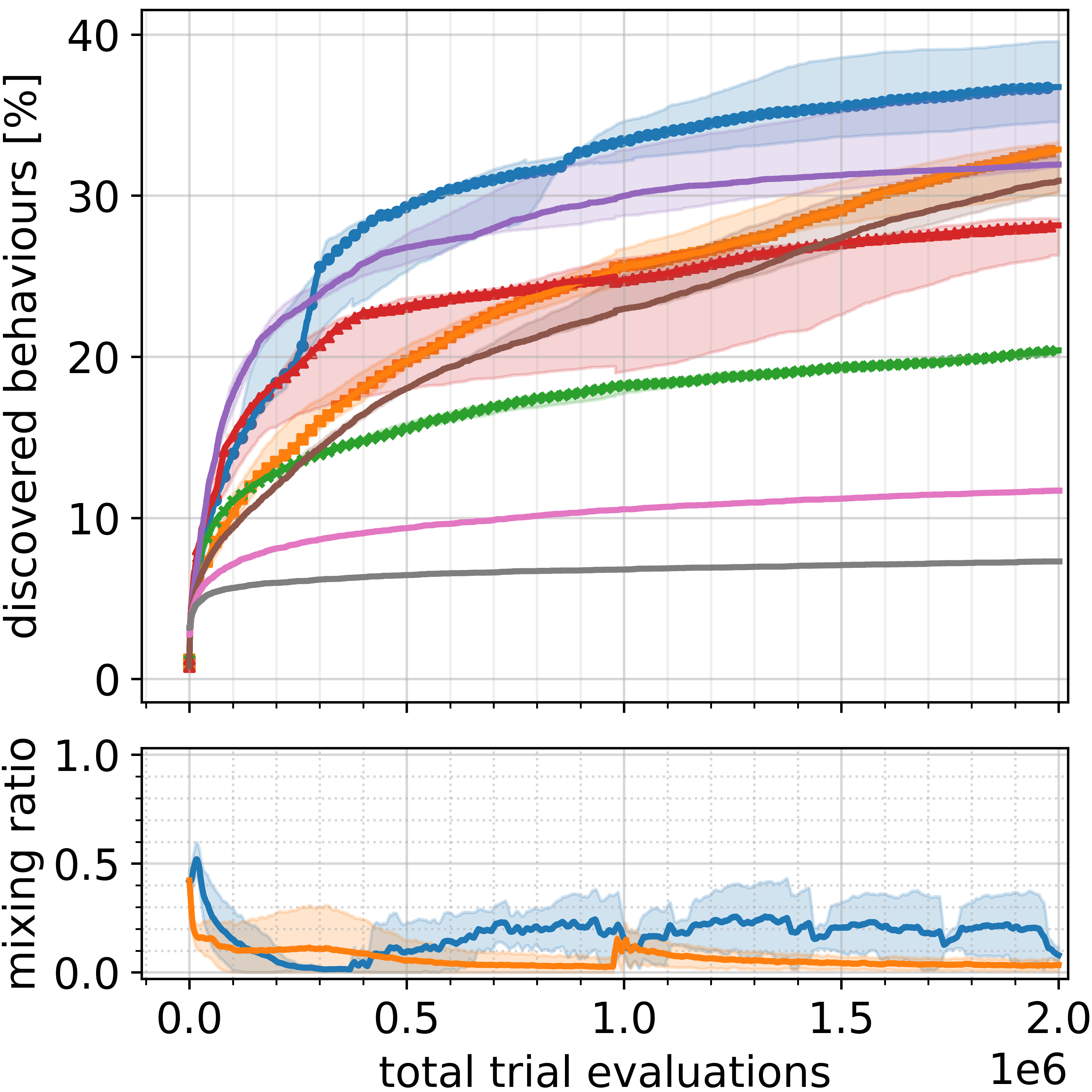}
        \label{fig:res_plots:biped_walk}} 
        \subfloat[Bipedal-Kicker]{\includegraphics[width=0.32\textwidth]
        {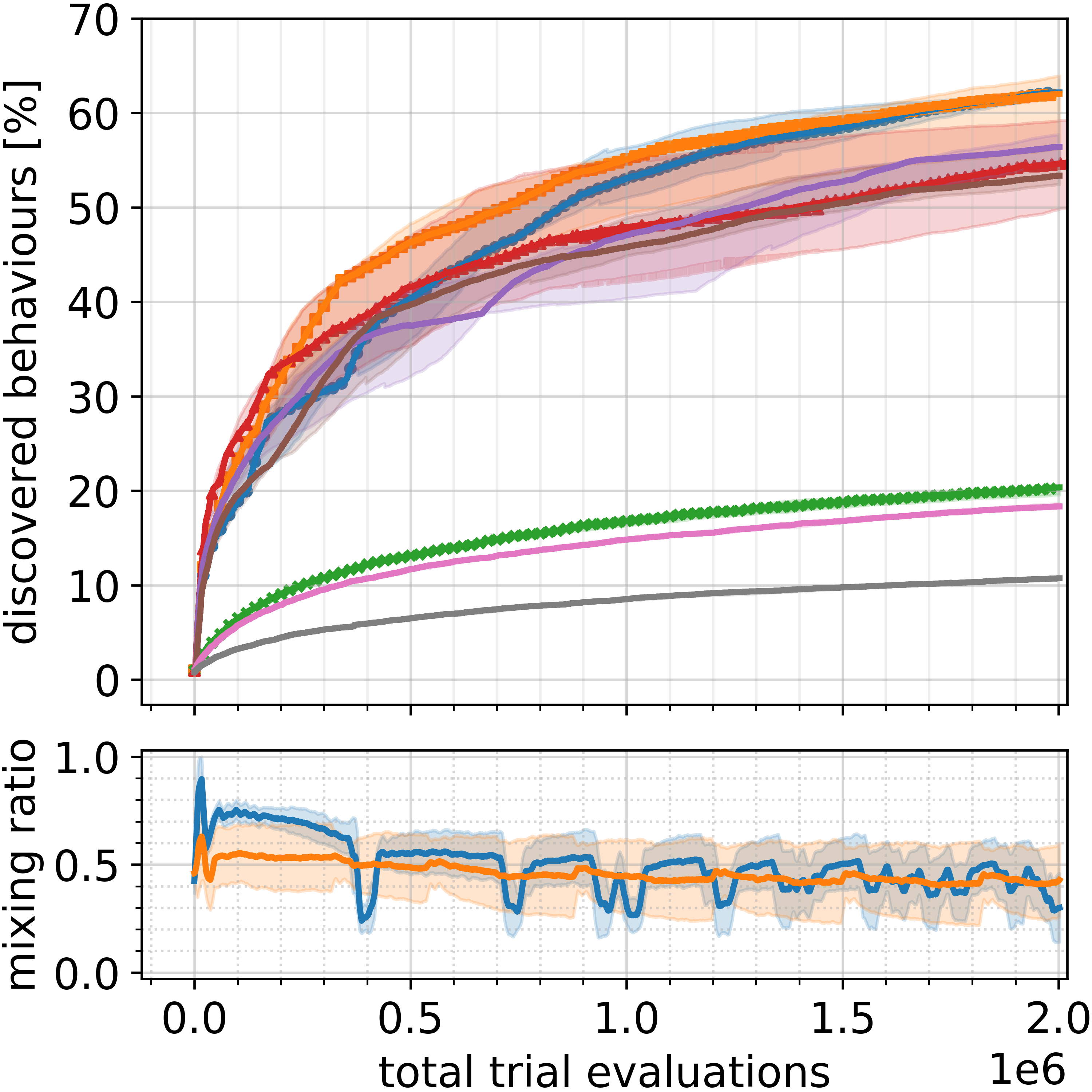}
        \label{fig:res_plots:biped_kick}} 
        \subfloat[Striker]{\includegraphics[width=0.32\textwidth]
        {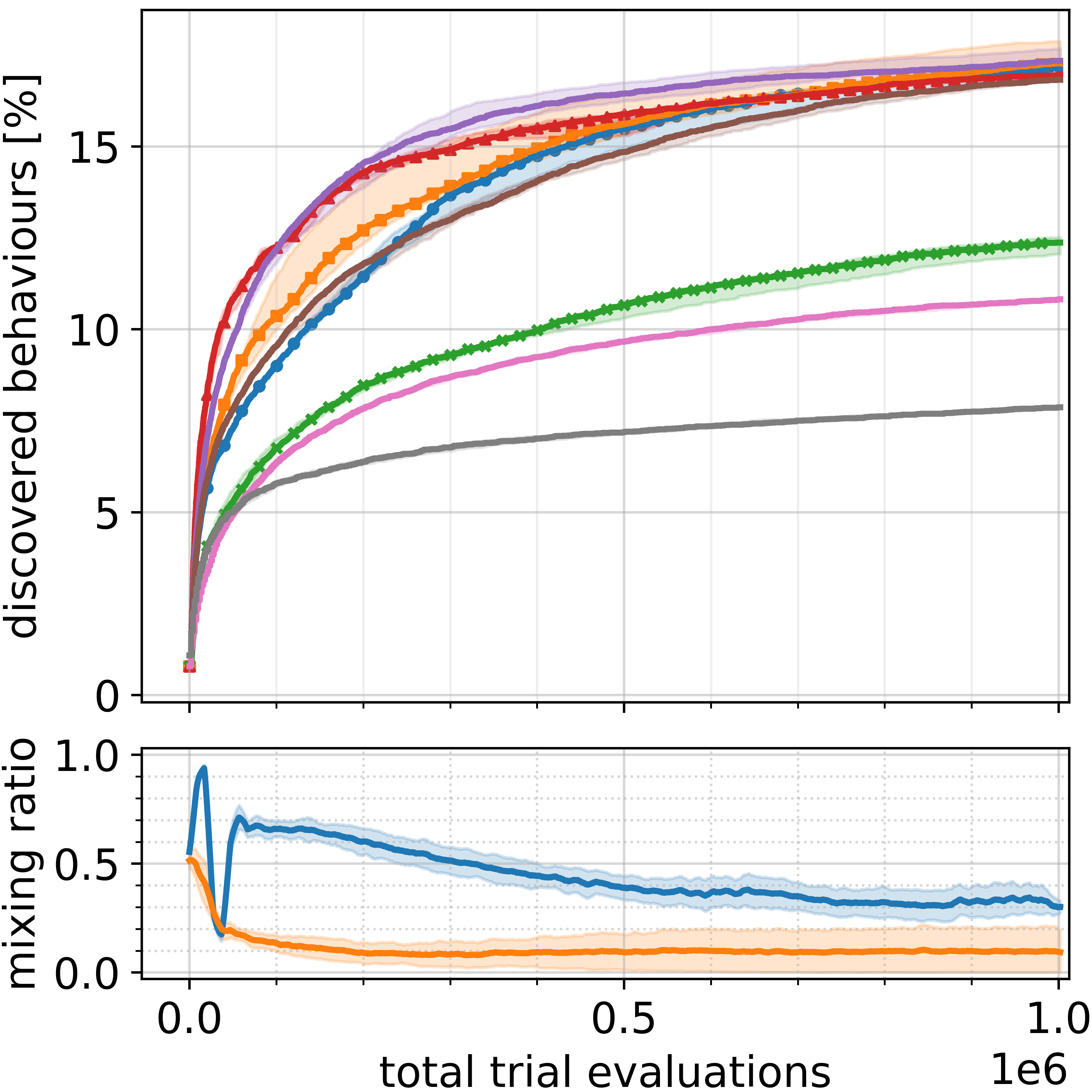}
        \label{fig:res_plots:striker}} 
        \\
        \subfloat[Bipedal-Walker mixed scale]{\includegraphics[width=0.32\textwidth]
        {Images/results/bd_coverage__nolegend_bdpercent_BIPEDAL_WALKER_AUGMENTED_MIX_SCALE.png}
        \label{fig:res_plots:biped_walk_mix}} 
        \subfloat[Bipedal-Kicker mixed scale]{\includegraphics[width=0.32\textwidth]
        {Images/results/bd_coverage__nolegend_bdpercent_BIPEDAL_KICKER_AUGMENTED.png}
        \label{fig:res_plots:biped_kick_mix}} 
        \subfloat[Striker mixed scale]{\includegraphics[width=0.32\textwidth]
        {Images/results/bd_coverage__nolegend_bdpercent_STRIKER_AUGMENTED_MIX_SCALE.png}
        \label{fig:res_plots:striker_mix}} 
    \caption{Behaviour coverage and mixing-ratio plots of the compared approaches achieved in three continuous control environments. The markers on lines corresponding to approaches using a latent representation show points at which the latent representation is updated.
    }
    \label{fig:res_plots}
    \end{figure}
    
    \subsection{Discussion}
    
        Below, we discuss the behaviour coverage results from Fig.\ref{fig:res_plots} achieved by the compared algorithms in more detail.
        
        \textbf{A1. Learned latent space vs original parameter space search.}
        The first conclusion we note from the experiments, is that the standard MAP-Elites algorithm (MAPE-Iso) is competitive in high-dimensional parameter space problems, which has not been sufficiently investigated in previous work.
        Comparing the proposed PoMS approach with MAPE, PoMS systematically achieves higher behaviour coverage across the tasks. The only exception is the Striker case (Fig. \ref{fig:res_plots:striker} and \ref{fig:res_plots:striker_mix}), where methods converge to the same performance.
        This can be attributed to the simplicity of the task, as the policy outputs can directly influence the planar movement of the striker and by extension the puck, while this connection is more complex within bipedal locomotion.
        Even though MAPE-IsoLineDD operates in the parameter space, it exploits the notion of a hyper-volume of elites which makes it more efficient than MAPE-Iso. However, besides the Striker tasks where it reaches equal asymptotic performance, it converges to a lower behaviour coverage compared to PoMS.
        By definition of the line mutation, the MAPE-IsoLineDD usually performs well when the hyper-volume is convex. When this assumption does not hold, a more involved transformation is needed which PoMS realises via manifold learning.
        These examples validate the benefits of using a learned latent representation of the parameter space for policy parameter search.
        %
        %
        %
            
        \textbf{A2. Contribution of using the Jacobian of the decoder.}
        By comparing PoMS and PoMS-no-jacobian, we can clearly see that generating new policies through random parameter perturbation in the latent space and reconstruction, not accounting for the Jacobian, achieves significantly worse behaviour diversity.
        %
        %
        As hypothesised, this leads to a search that is almost random, which explains the results where PoMS-no-jacobian shows similar performance as random search (Fig. \ref{fig:res_plots:biped_kick}, \ref{fig:res_plots:biped_kick_mix}, \ref{fig:res_plots:striker_mix}).
        
        \textbf{A3. Linear vs non-linear representations.}
        The difference in performance between PoMS and PoMS-PCA, speaks mostly about the intrinsic complexity of the given task control problem. 
        Tasks in which locomotion is involved have an intrinsic non-linearity in the mapping of the policy outputs and actual motions contained in the observation vector, which is used to determine behaviour descriptors.
        This explains the performance gap in tasks other than Striker. 
        However, performance equal to PoMS seen in Bipedal-Kicker (Fig.~\ref{fig:res_plots:biped_kick}) is due to the observations being normalized which keep the parameters well behaved and easier to obtain a useful linear projection, while this does not hold in Striker-Kicker mix-scale (Fig.~\ref{fig:res_plots:biped_kick_mix}) which explains the drop in performance
        
        \textbf{A4. State-of-the-art performance comparison.}
        The top performing methods are PoMS, PoMS-PCA, DDE and MAPE-IsoLineDD, where the characteristic they have in common is focusing the search in the hyper-volume of elites or the learned manifold. This further solidifies claims from A1.
        Moreover, comparing `normal' and mix-scale versions of the environments (top and bottom row of Fig.~\ref{fig:res_plots}), we can see the robustness of the approaches using some form of a manifold for search. As we introduce a large variation of scale among the elements of the observation vector, this causes a reduction in the overall achieved behaviour diversity. This drop in performance is much higher in purely parameter space search methods such as MAPE-Iso, ps-uniform and ps-normal.
        The proposed PoMS approach outperforms the next best state-of-the-art approach by up to 5\%, except in the Striker tasks where it achieves on-par performance at convergence, and Bipedal-Kicker where PoMS and PoMS-PCA have similar performance which is ~5\% better than the next best MAPE-IsoLineDD.
    
        \textbf{Mixing ratio}
        Below the corresponding behaviour coverage plots, we show the mixing ratio plots.
        For the PoMS and PoMS-PCA, the plots represent the averaged ratio of samples generated in the latent versus the parameter space during the training. Mixing ratio of 1 means that all of the samples are taken in the parameter space, and vice versa.
        The first loop of the algorithm draws with a mixing ration of 0.5 and subsequently the ratio changes based on the mean reconstruction error as explained in the Method section.
        From mixing ratios we can see that in the beginning there is usually a spike to take more parameter space samples. This is due to the fact that there are many sampled points with a high reconstruction error in the beginning, because initially the AE was fitted on the small amount of data and needs more data diversity - thus it `explores'.
        The mixing ratio slowly decreases in favor of the latent space samples, with several salient `dips' which slightly correlate to rises in behaviour discovery. This can be interpreted as the algorithm `exploiting' the latent representation. 
        This is not evident with PCA as its representations tend to be more rigid and do not change often with new data. 
    

\section{Achieved behaviour diversity}
    In Tables \ref{tab:striker}-\ref{tab:bipedal_kicker_mix}, we show the achieved diversity of trajectories and policy collection coverage during the experiments, for each of the compared algorithms. We are showing the data corresponding to the median curves from the main results graph (Figure 3. in the main manuscript).
    The colouring of trajectories is based on a specific dimension of the policy collection cell-grid:
    
    For Striker and Striker mix-scale experiments (Tables~\ref{tab:striker} and~\ref{tab:striker_mix}), each colour represents the index of D3, which describes the wall(s) with which the puck collided. The corresponding grid plot separates the x-y area based on this dimension.
    
    For Bipedal-Walker and Bipedal-Walker mix-scale experiments (Tables~\ref{tab:bipedal_walker} and~\ref{tab:bipedal_walker_mix}), each colour represents the index of D3, which describes left leg's duty factor, i.e. proportion of time the left leg spent in contact with the ground.
    
    For Bipedal-Kicker and Bipedal-Kicker mix-scale experiments (Tables~\ref{tab:bipedal_kicker} and~\ref{tab:bipedal_kicker_mix}), each colour represents the index of D2, which describes the max y-coordinate of the ball achieved during an episode.

    \begin{table}[h!]
    \centering
    \caption{Puck trajectories and policy collections for Striker experiments.}
    \label{tab:striker}
        \begin{tabular}{ | m{0.22\textwidth} m{0.215\textwidth} || m{0.22\textwidth} m{0.215\textwidth}| }
            \hline
            \multicolumn{4}{|l|}{\textbf{Algorithm}} \\
            \multicolumn{4}{|c|}{\textbf{Striker}}  \\
            \hline
            \hline
            \multicolumn{2}{|l||}{\textbf{PoMS}}  &  \multicolumn{2}{l|}{\textbf{MAPE-IsoLineDD}} \\       
            \includegraphics[width=0.249\textwidth]{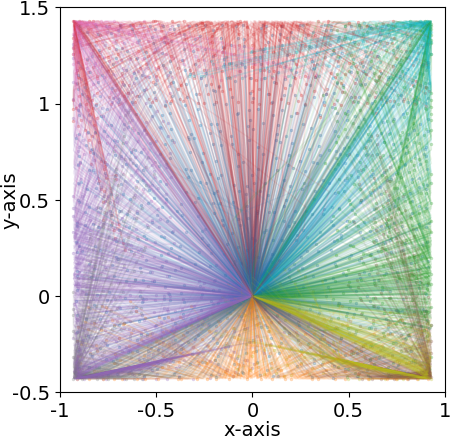}
            & 
            \includegraphics[width=0.225\textwidth]{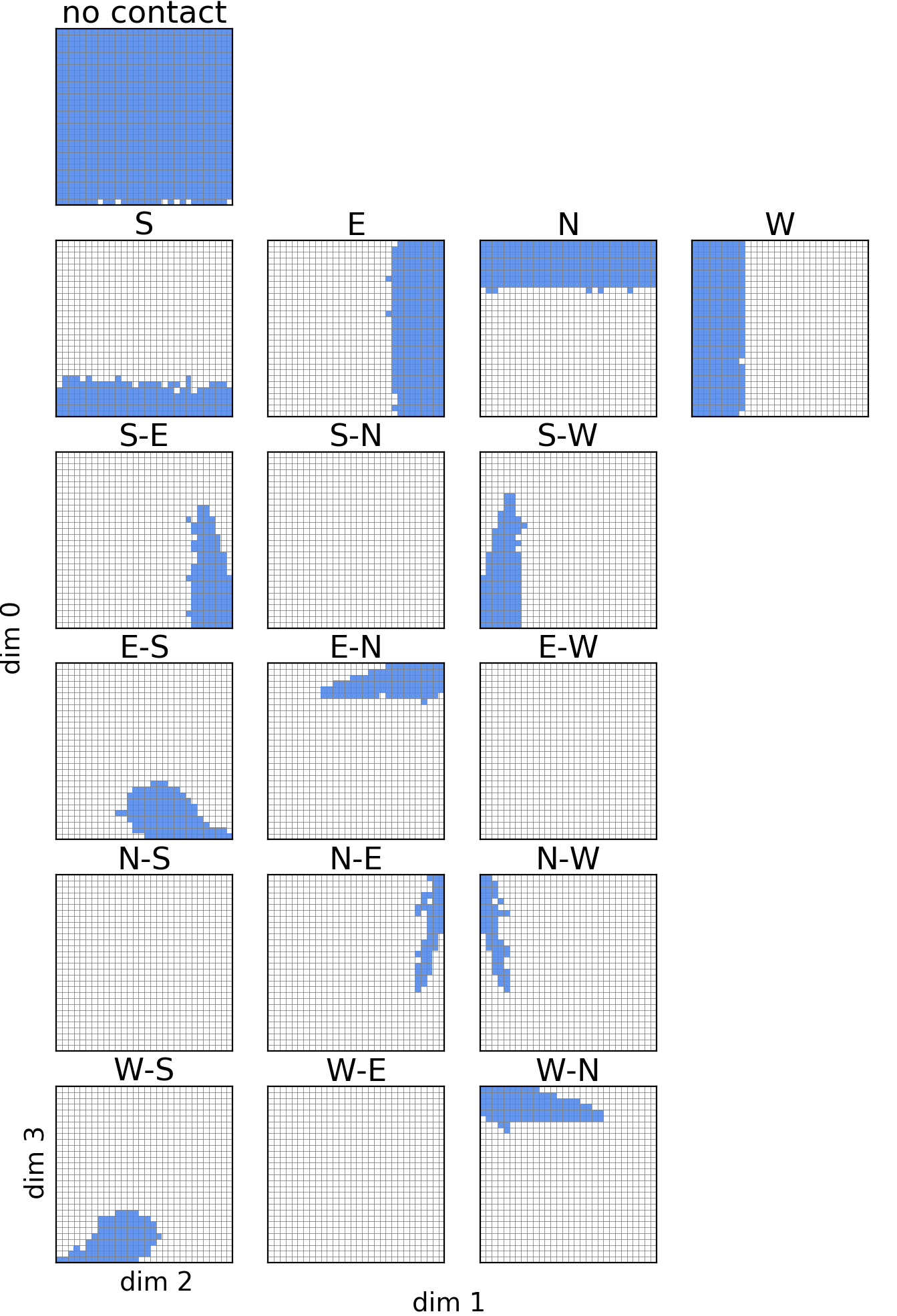}
            & 
            \includegraphics[width=0.249\textwidth]{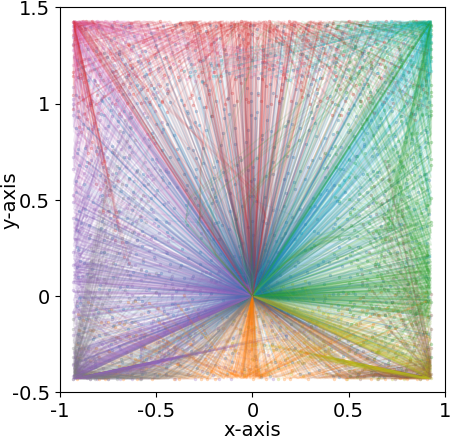}
            &
            \includegraphics[width=0.225\textwidth]{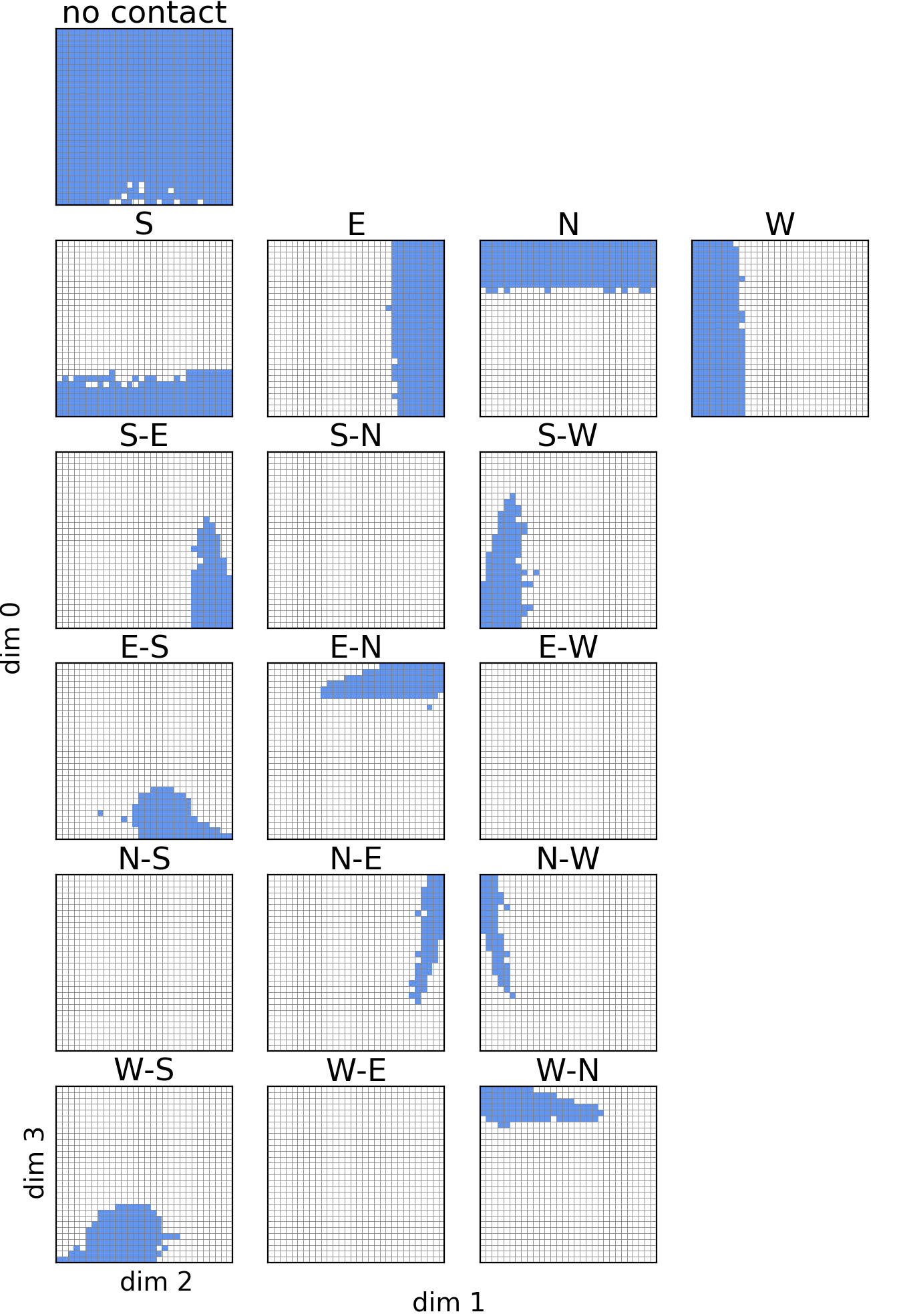}
            \\
            \hline 
            \multicolumn{2}{|l||}{\textbf{PoMS-PCA}}  &  \multicolumn{2}{l|}{\textbf{MAPE-Iso}} \\       
            \includegraphics[width=0.249\textwidth]{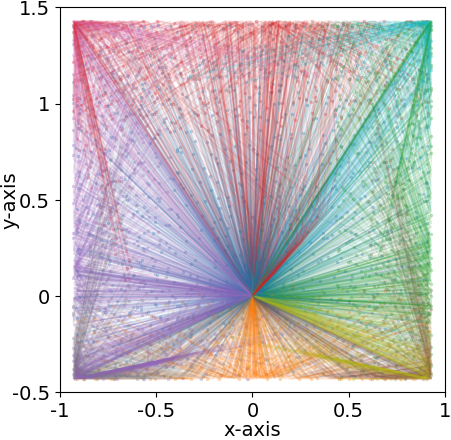}
            &
            \includegraphics[width=0.225\textwidth]{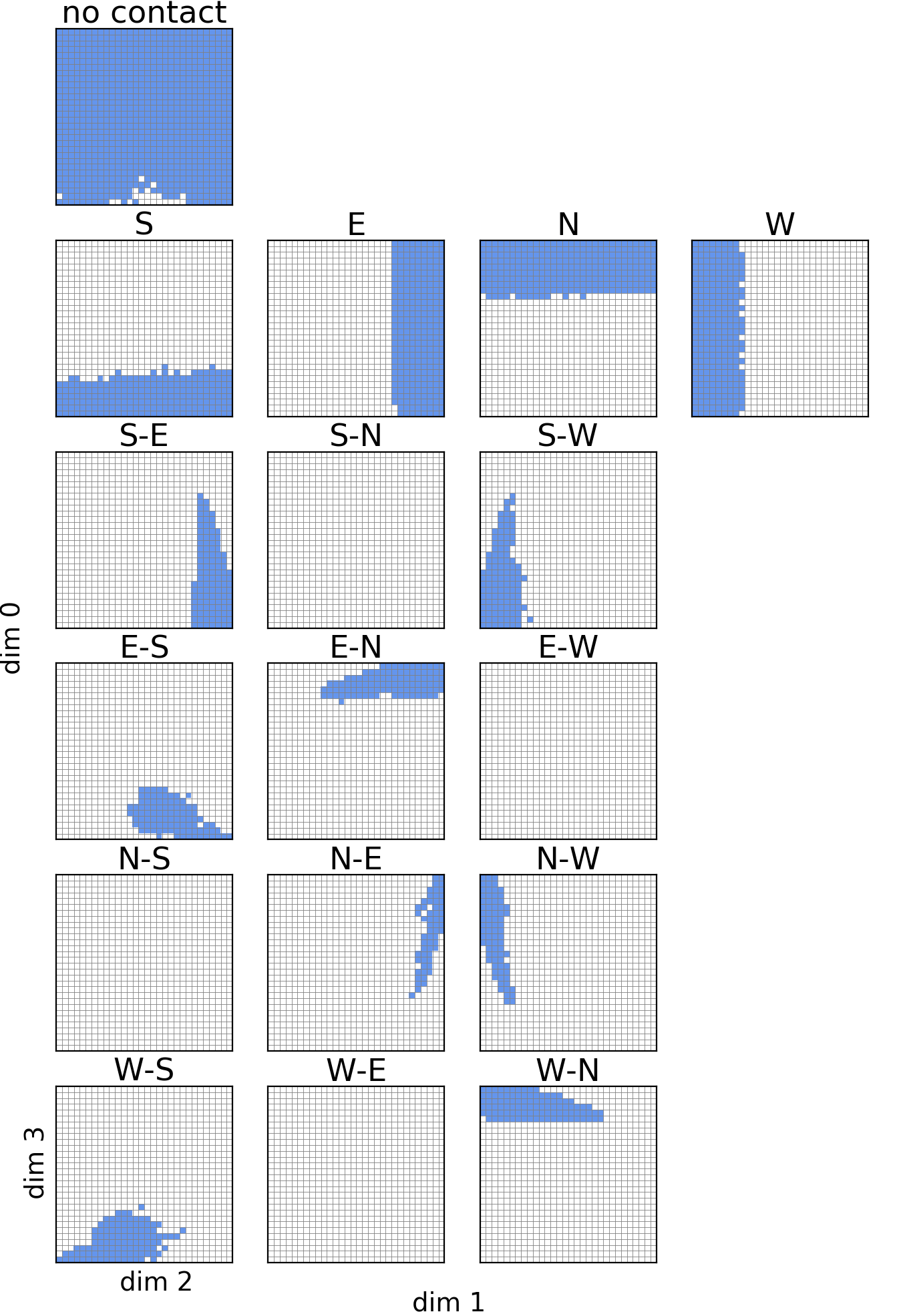}
            & 
            \includegraphics[width=0.249\textwidth]{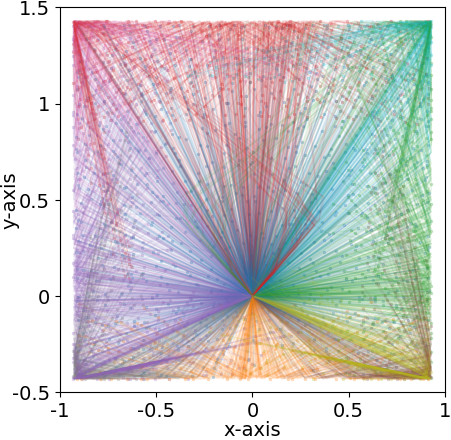}                  & 
            \includegraphics[width=0.225\textwidth]{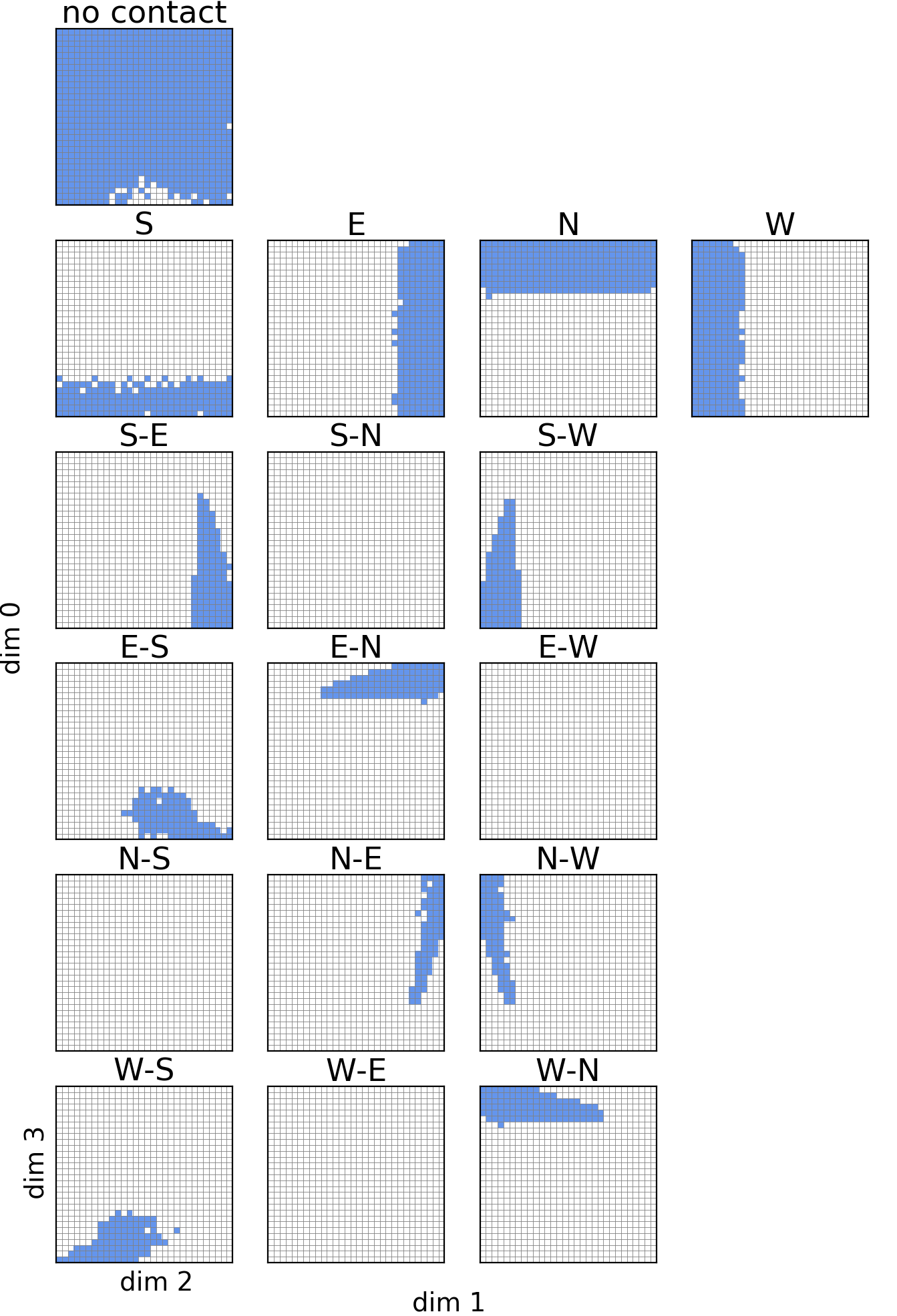}
            \\
            \hline 
            \multicolumn{2}{|l||}{\textbf{PoMS-no-jacobian}}  &  \multicolumn{2}{l|}{\textbf{ps-uniform}} \\       
            \includegraphics[width=0.249\textwidth]{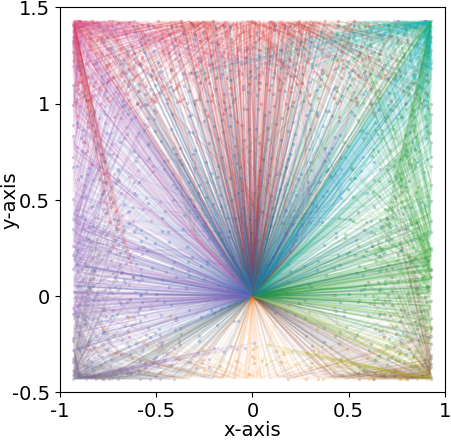}
            &
            \includegraphics[width=0.225\textwidth]{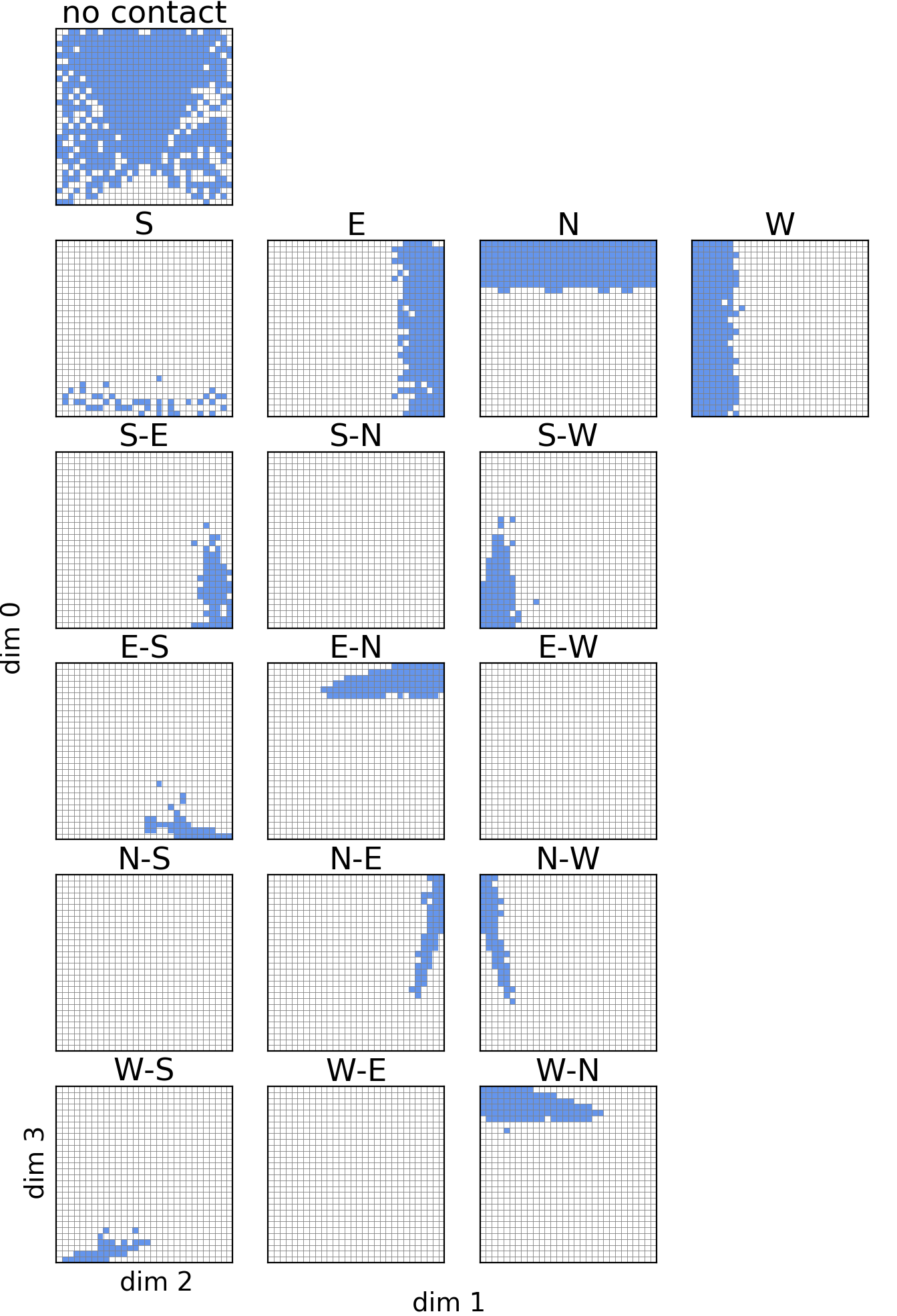}
            &
            \includegraphics[width=0.249\textwidth]{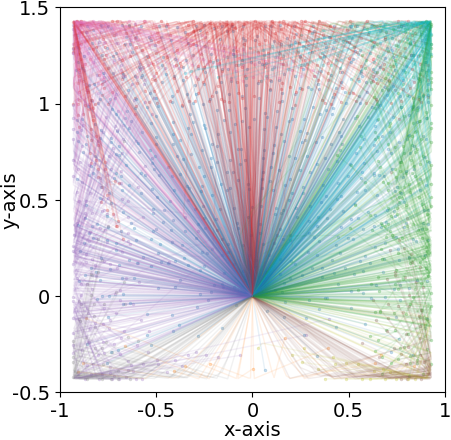}            &
            \includegraphics[width=0.225\textwidth]{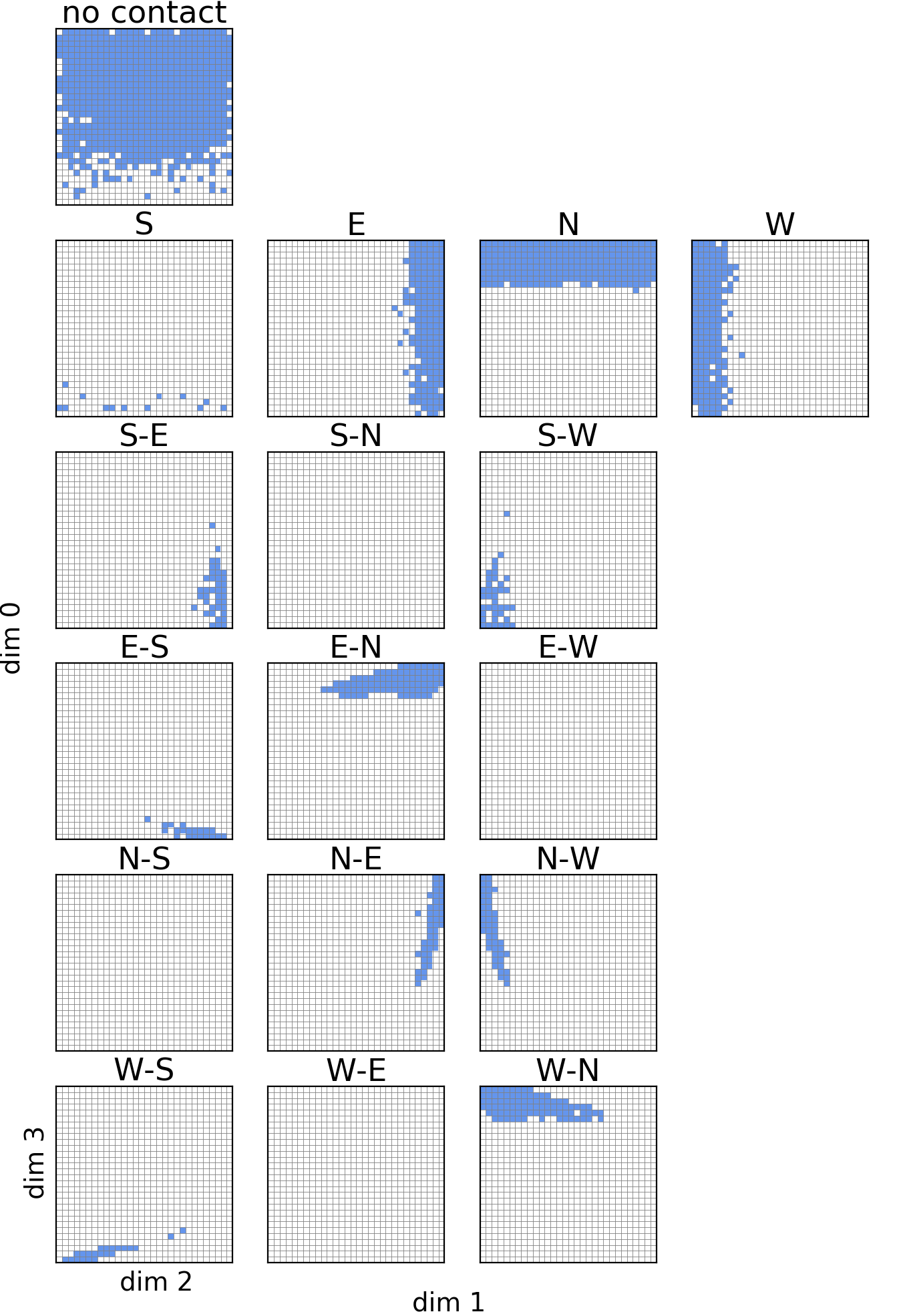}
            \\
            \hline 
            \multicolumn{2}{|l||}{\textbf{DDE}}  &  \multicolumn{2}{l|}{\textbf{ps-glorot}} \\       
            \includegraphics[width=0.249\textwidth]{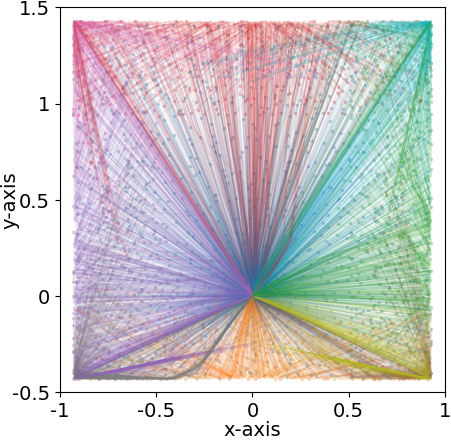}           &
            \includegraphics[width=0.225\textwidth]{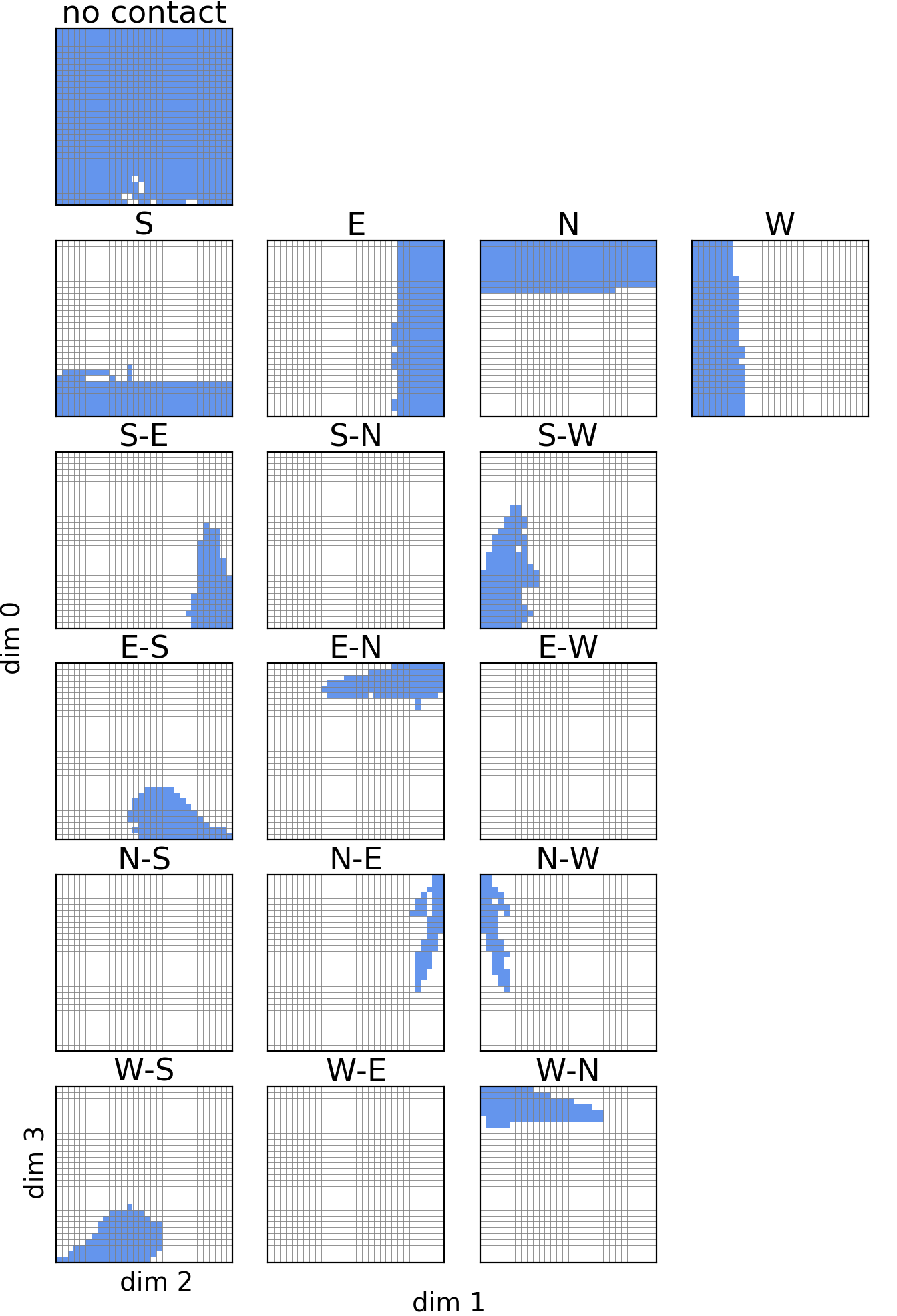}
            &
            \includegraphics[width=0.249\textwidth]{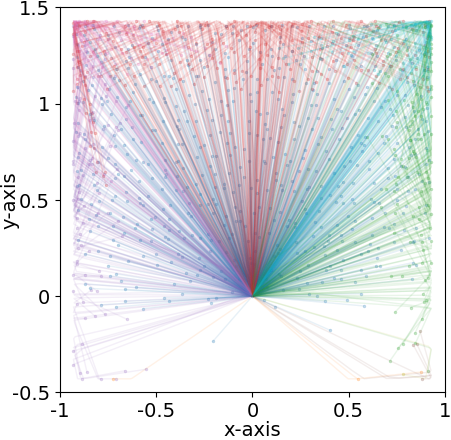}             & 
            \includegraphics[width=0.225\textwidth]{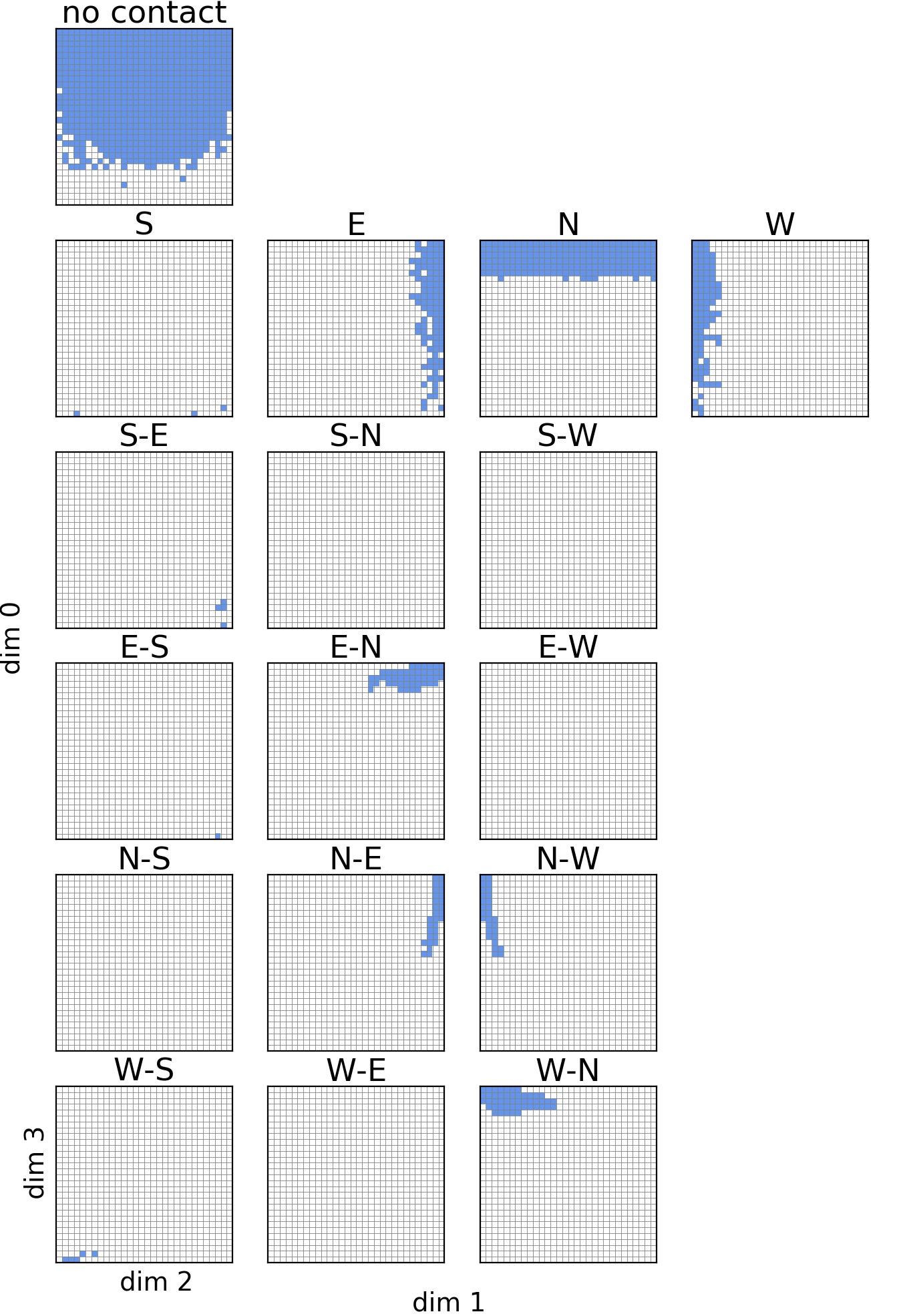}
            \\
            \hline
        \end{tabular}
    \end{table}
\newpage

    \begin{table}[h!]
    \centering
    \caption{Puck trajectories and policy collections for Striker mix-scale experiments.}
    \label{tab:striker_mix}
        \begin{tabular}{ | m{0.22\textwidth} m{0.215\textwidth} || m{0.22\textwidth} m{0.215\textwidth}| }
            \hline
            \multicolumn{4}{|l|}{\textbf{Algorithm}} \\
            \multicolumn{4}{|c|}{\textbf{Striker mix-scale}}  \\
            \hline
            \hline
            \multicolumn{2}{|l||}{\textbf{PoMS}}  &  \multicolumn{2}{l|}{\textbf{MAPE-IsoLineDD}} \\       
            \includegraphics[width=0.249\textwidth]{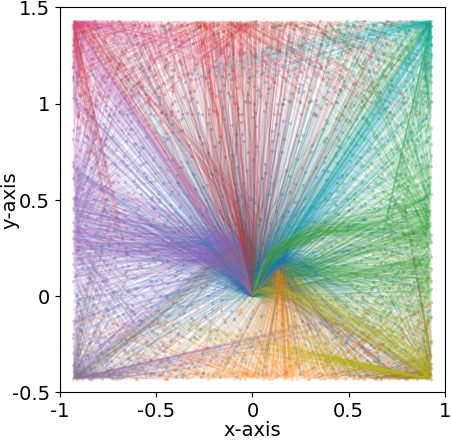}
            & 
            \includegraphics[width=0.225\textwidth]{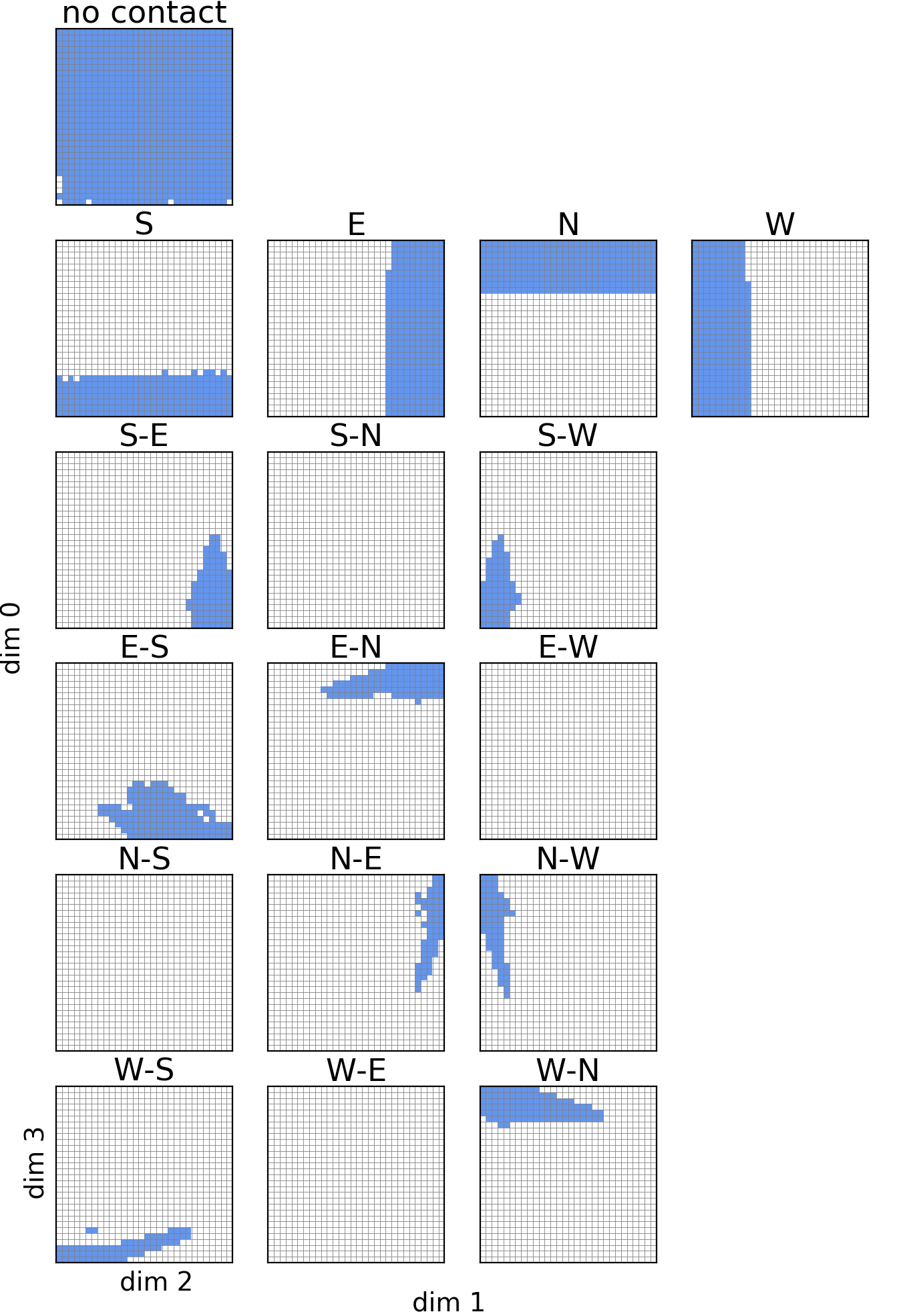}
            & 
            \includegraphics[width=0.249\textwidth]{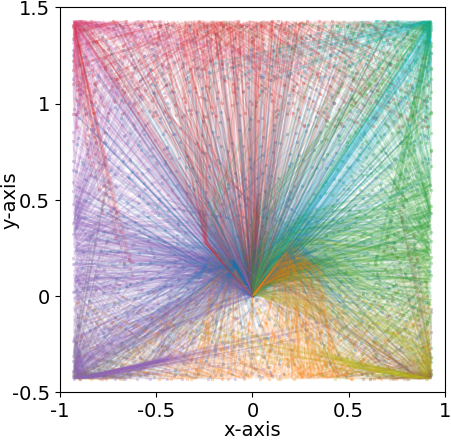}
            &
            \includegraphics[width=0.225\textwidth]{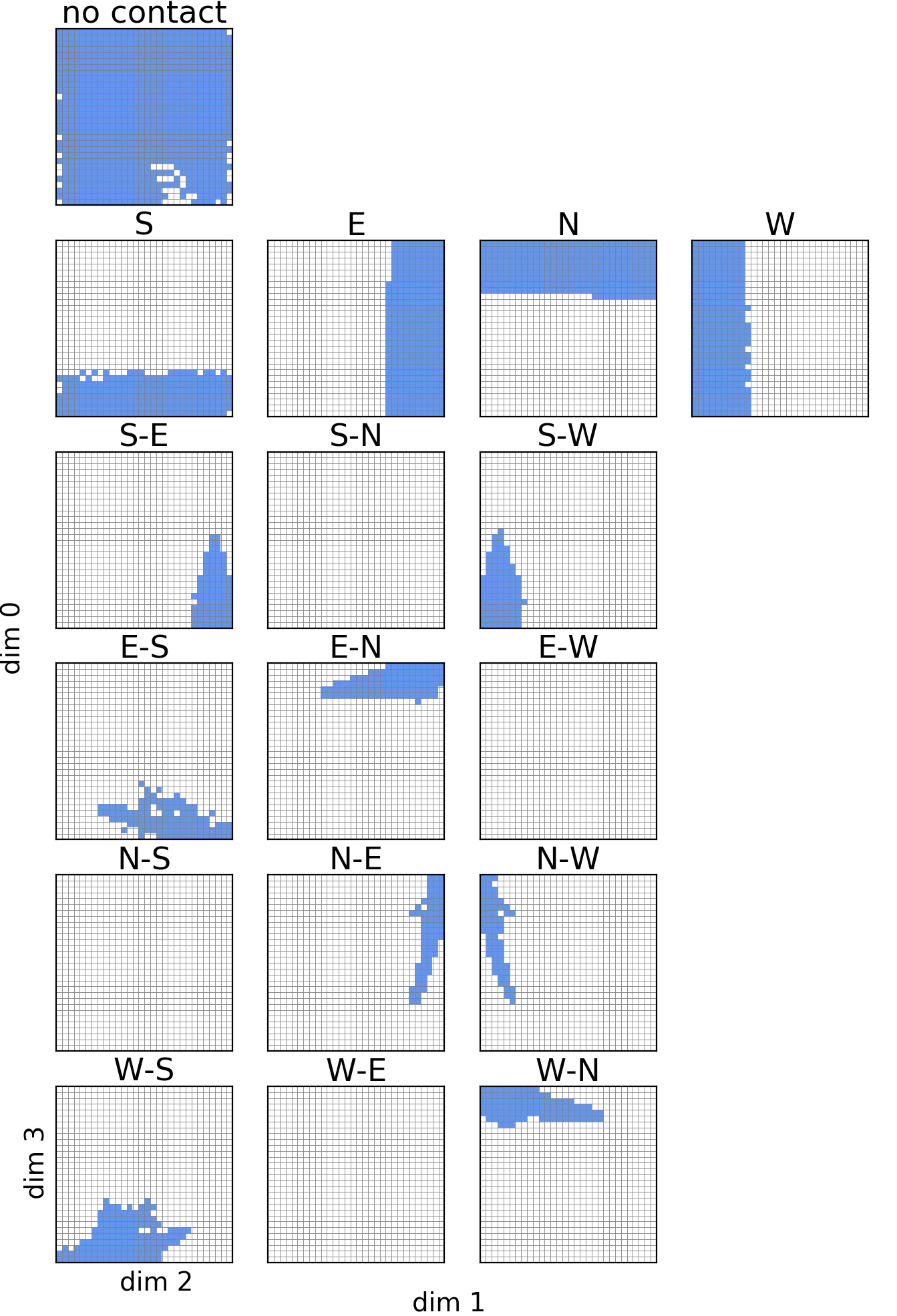}
            \\
            \hline 
            \multicolumn{2}{|l||}{\textbf{PoMS-PCA}}  &  \multicolumn{2}{l|}{\textbf{MAPE-Iso}} \\       
            \includegraphics[width=0.249\textwidth]{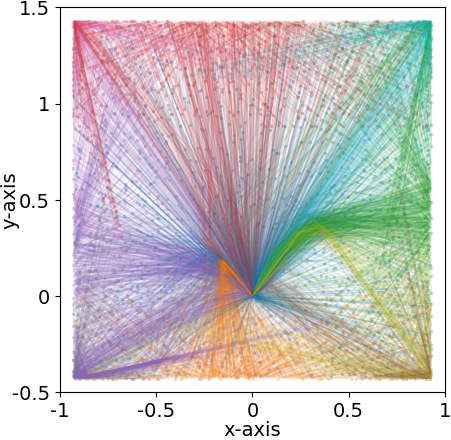}
            &
            \includegraphics[width=0.225\textwidth]{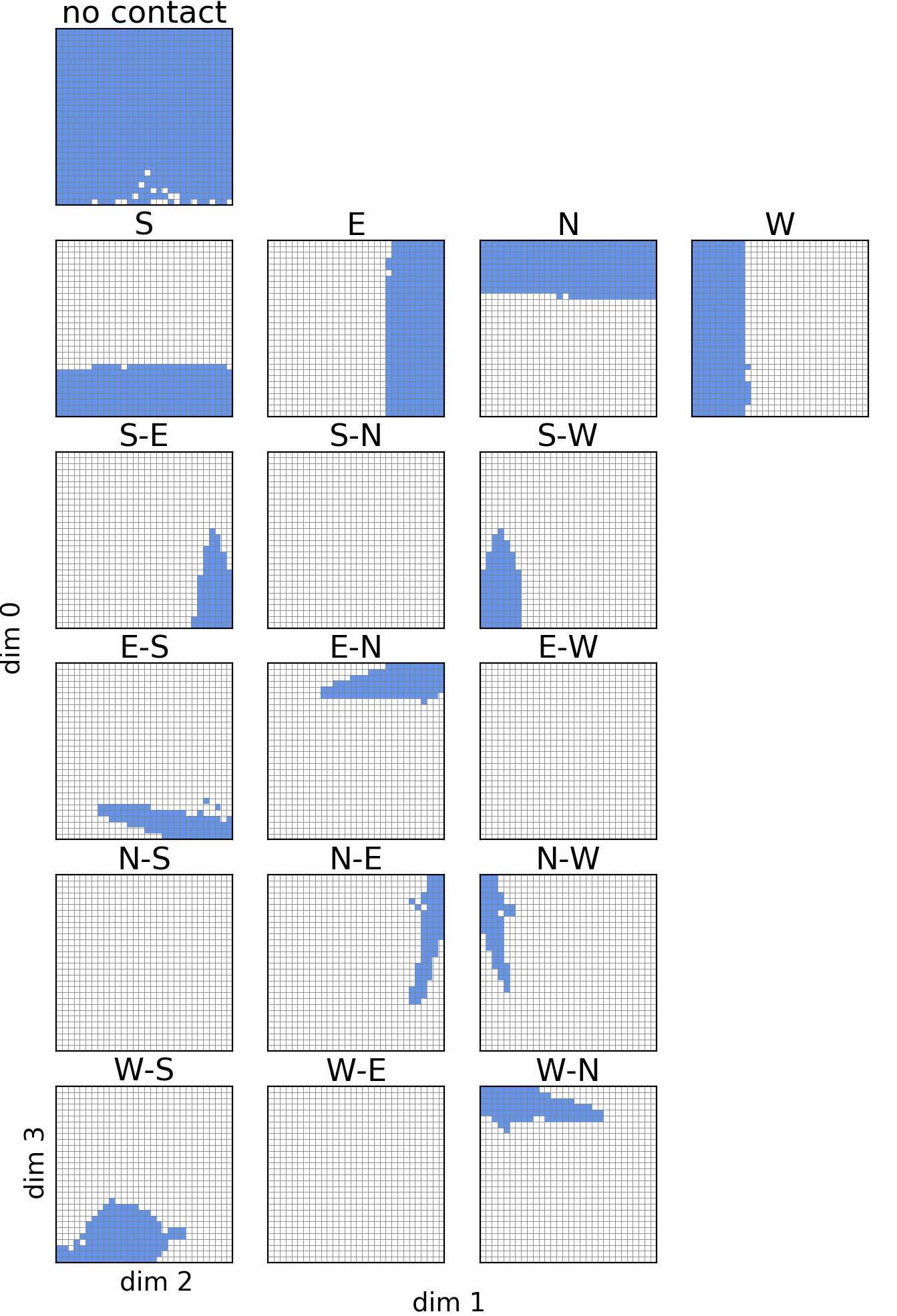}
            & 
            \includegraphics[width=0.249\textwidth]{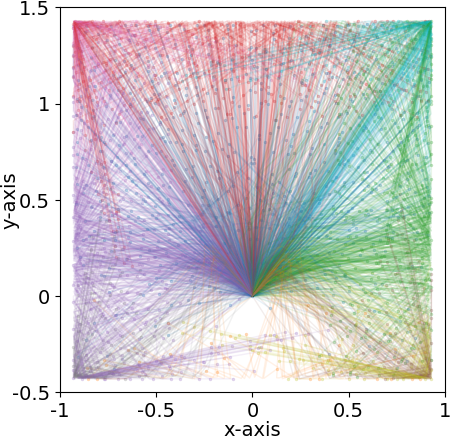}     
            & 
            \includegraphics[width=0.225\textwidth]{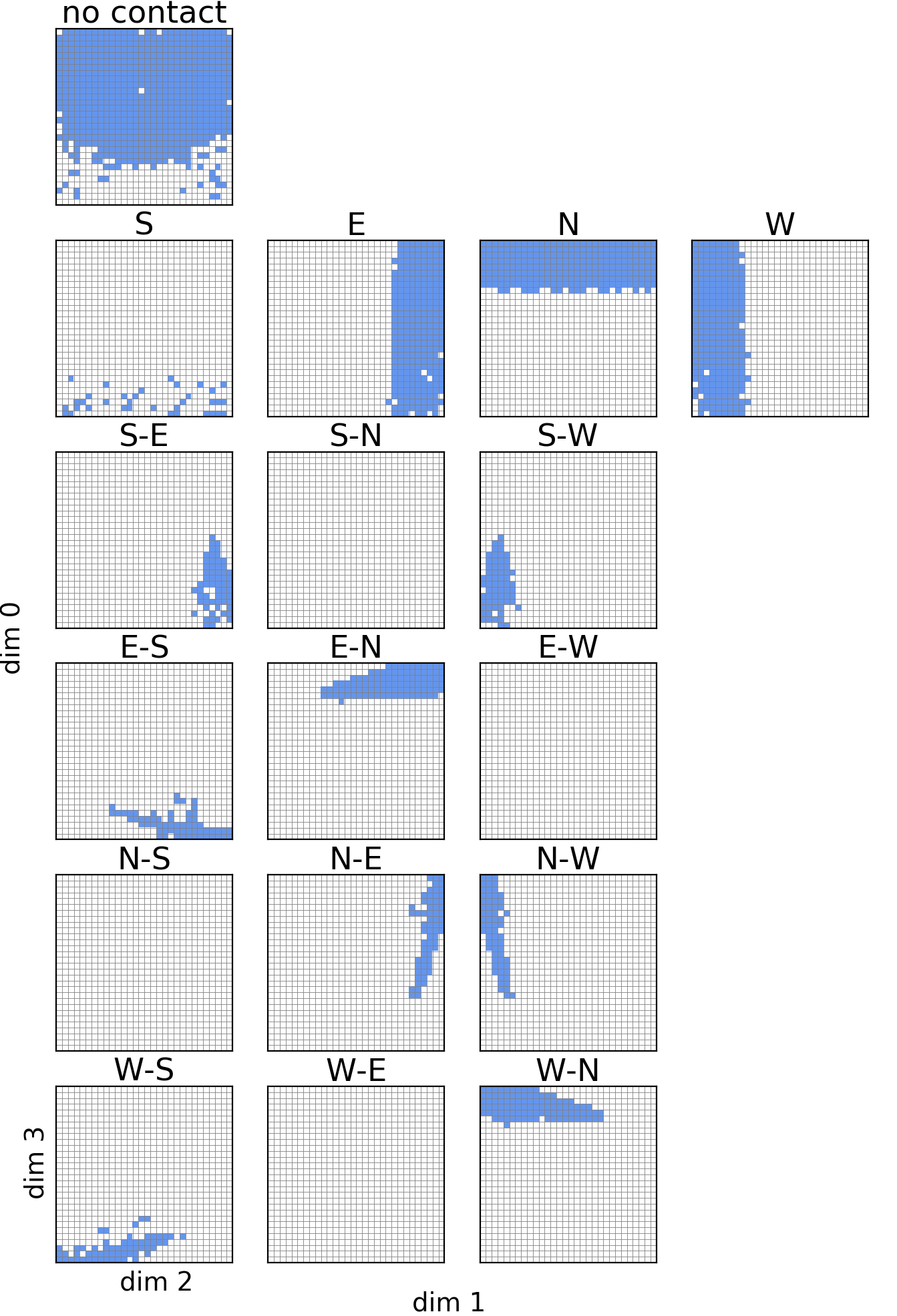}
            \\
            \hline 
            \multicolumn{2}{|l||}{\textbf{PoMS-no-jacobian}}  &  \multicolumn{2}{l|}{\textbf{ps-uniform}} \\       
            \includegraphics[width=0.249\textwidth]{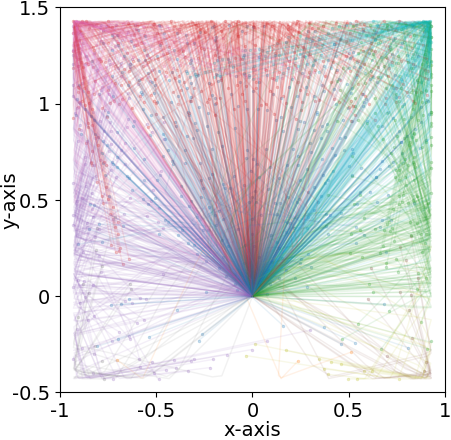}
            &
            \includegraphics[width=0.225\textwidth]{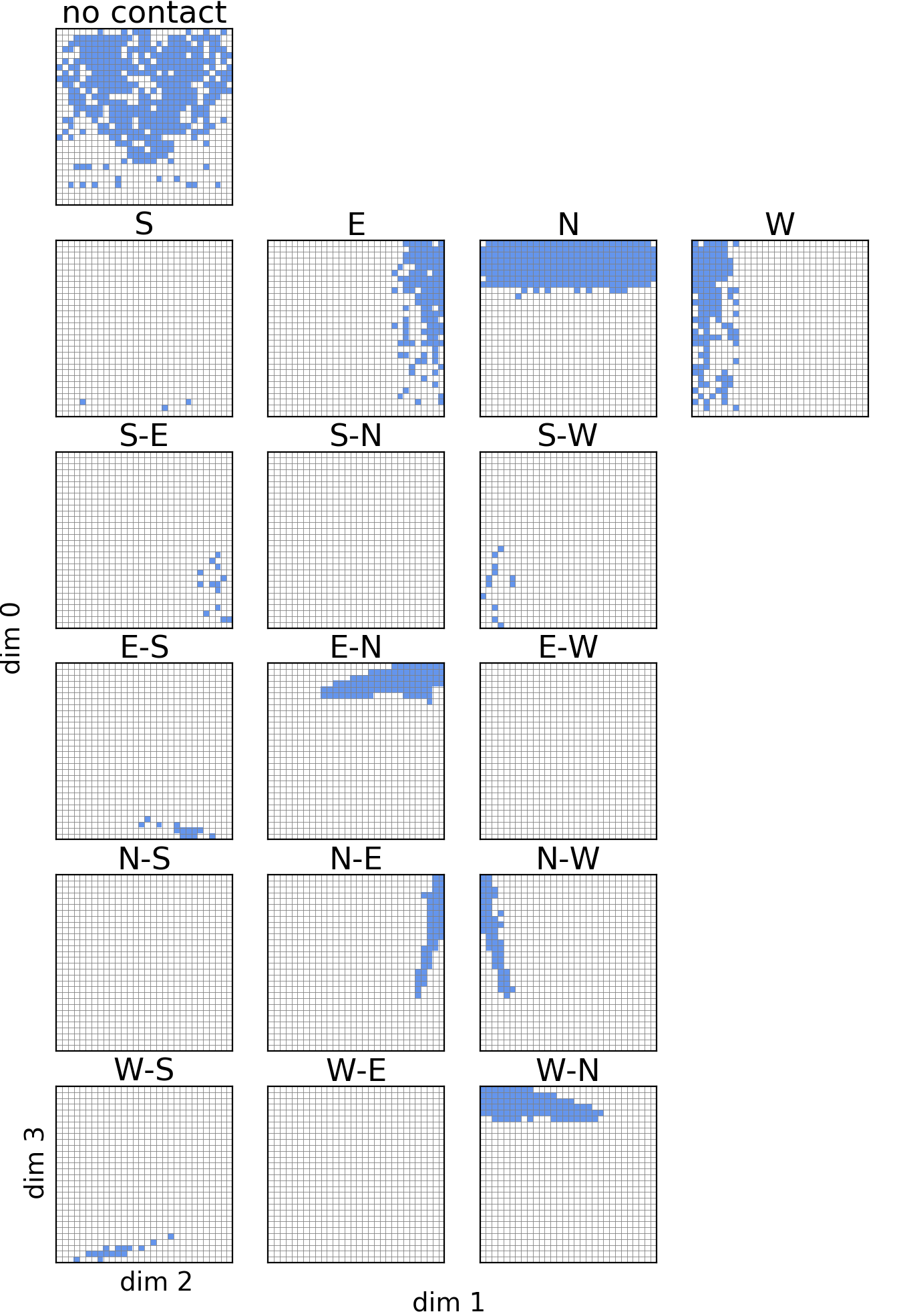}
            &
            \includegraphics[width=0.249\textwidth]{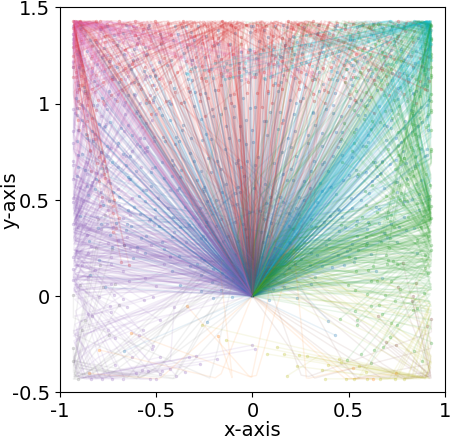}
            &
            \includegraphics[width=0.225\textwidth]{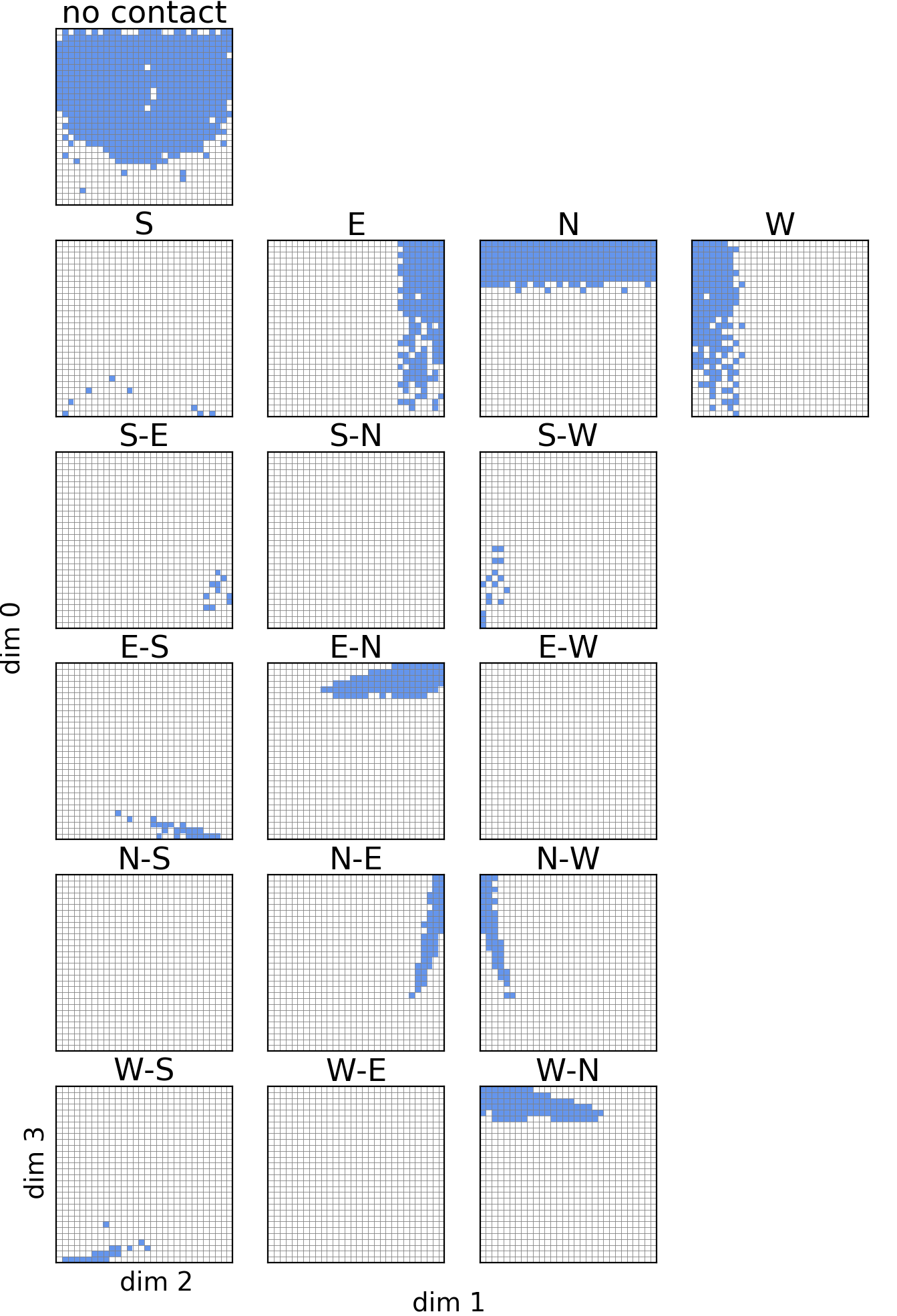}
            \\
            \hline 
            \multicolumn{2}{|l||}{\textbf{DDE}}  &  \multicolumn{2}{l|}{\textbf{ps-glorot}} \\       
            \includegraphics[width=0.249\textwidth]{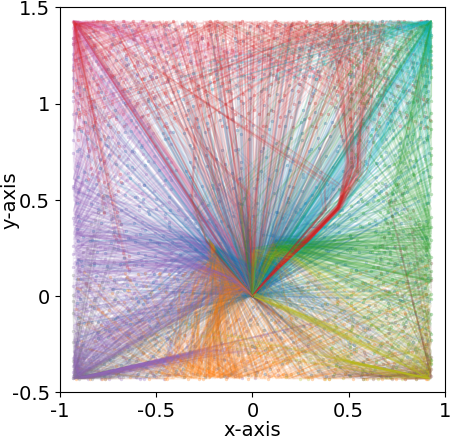}
            &
            \includegraphics[width=0.225\textwidth]{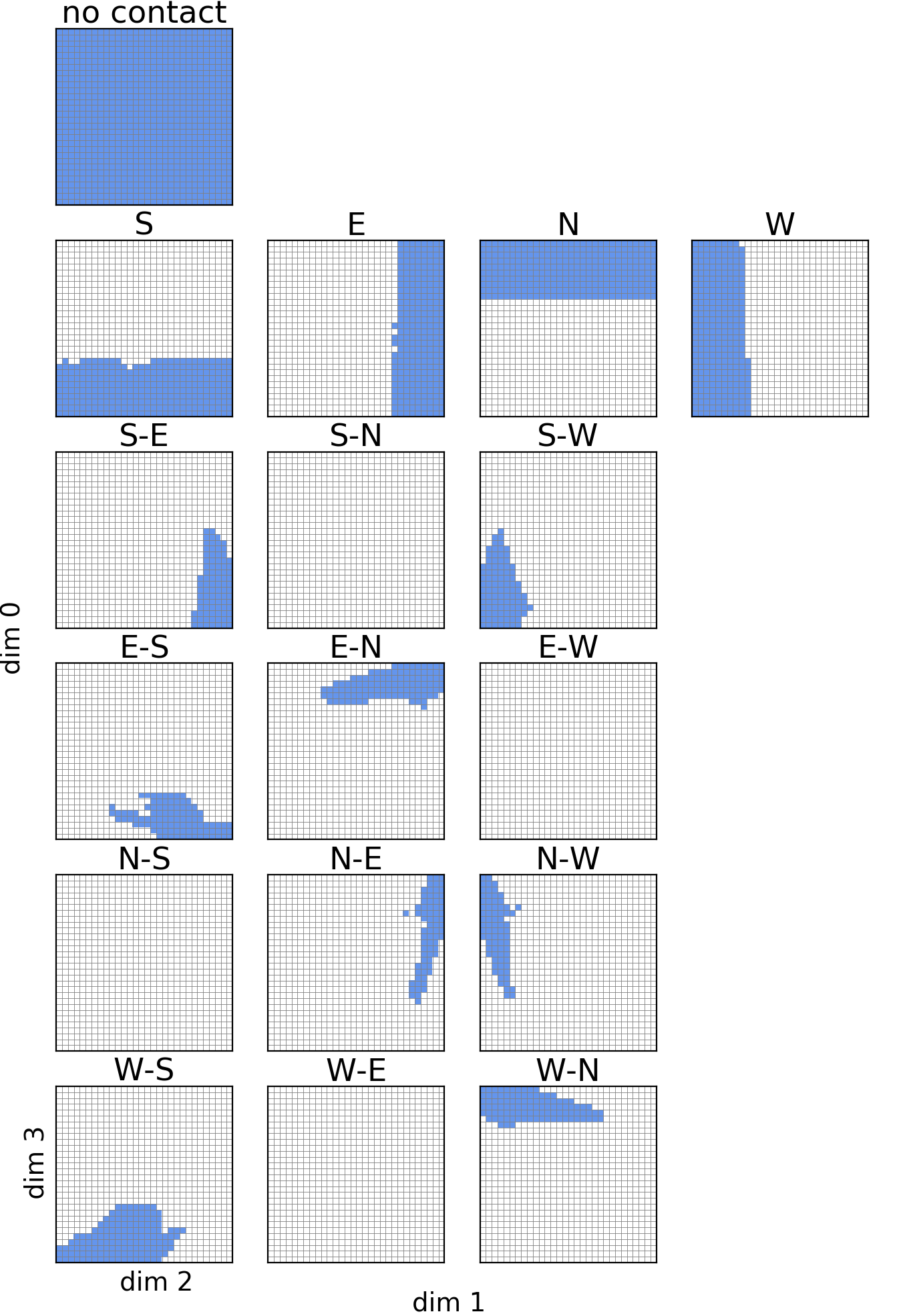}
            &
            \includegraphics[width=0.249\textwidth]{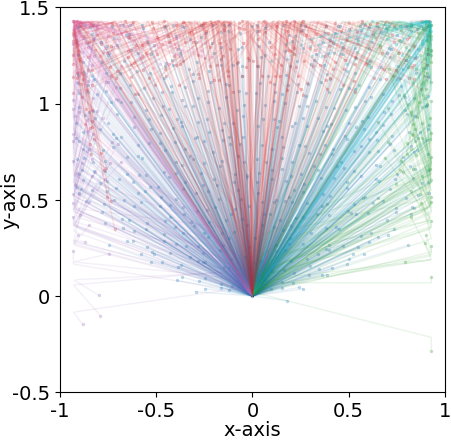} 
            & 
            \includegraphics[width=0.225\textwidth]{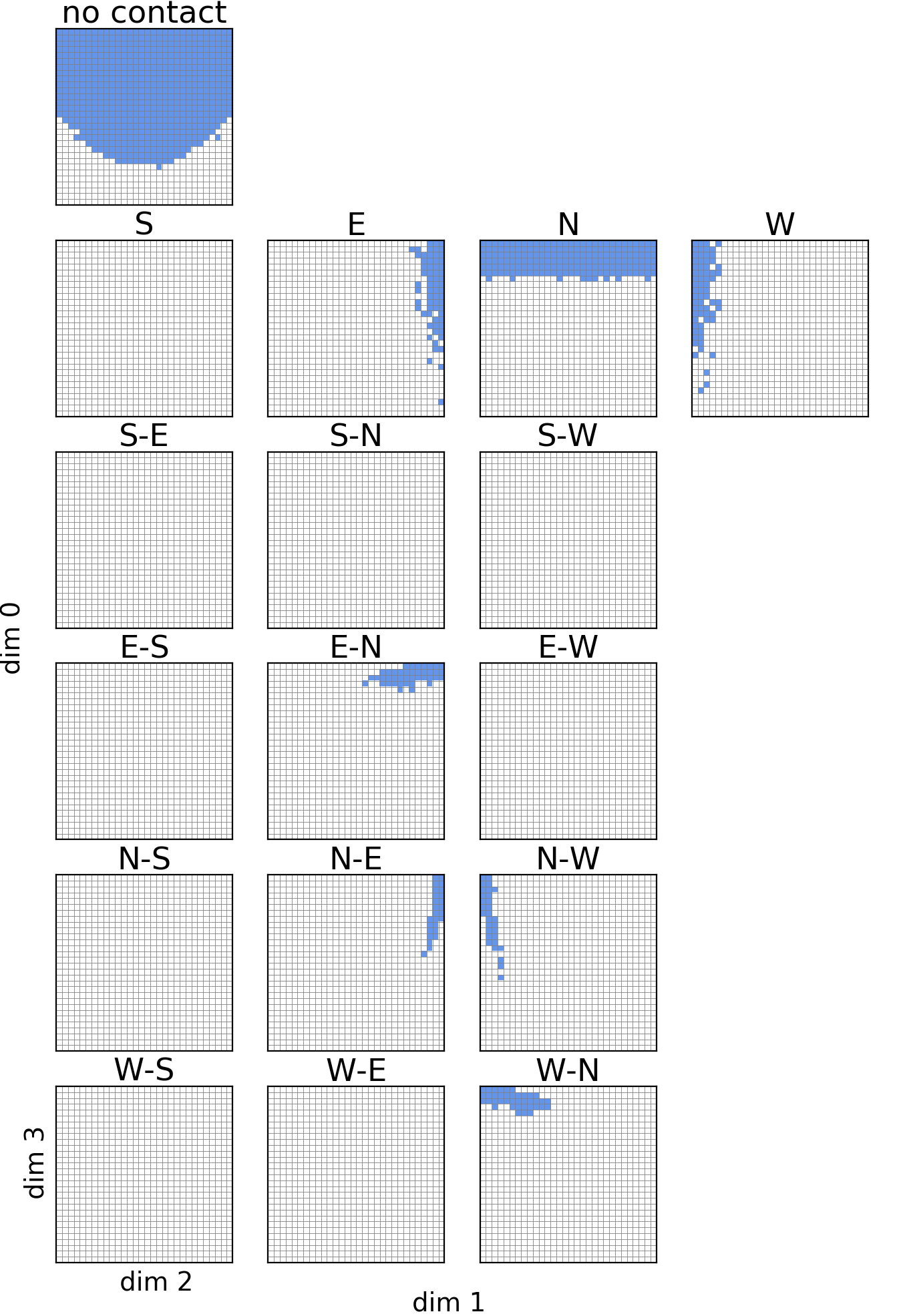}
            \\
            \hline
        \end{tabular}
    \end{table}
\newpage

     
    \begin{table}[h!]
    \centering
    \caption{Robot trajectories and policy collections for Bipedal-Walker experiments.}
    \label{tab:bipedal_walker}
        \begin{tabular}{ | m{0.2\textwidth} | m{0.7\textwidth} | }
            \hline
            \textbf{Algorithm}  & \multicolumn{1}{c|}{\textbf{Bipedal-Walker}}   \\
            \hline
            \hline
            \multirow{2}{*}{\textbf{PoMS}}
            & 
            \includegraphics[width=0.658\textwidth]{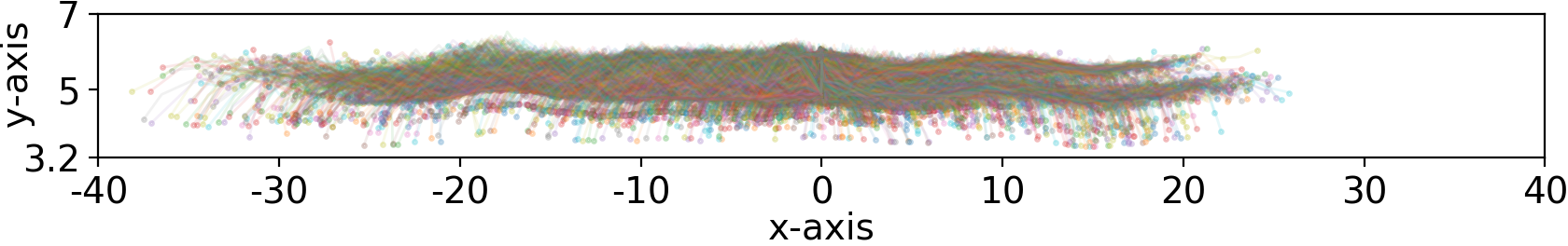} \\ 
            & 
            \includegraphics[width=0.658\textwidth]{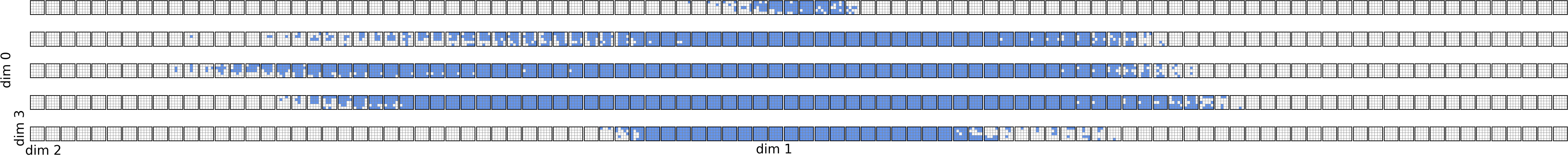}\\
            \hline
            \multirow{2}{*}{\textbf{PoMS-PCA}}
            & 
            \includegraphics[width=0.658\textwidth]{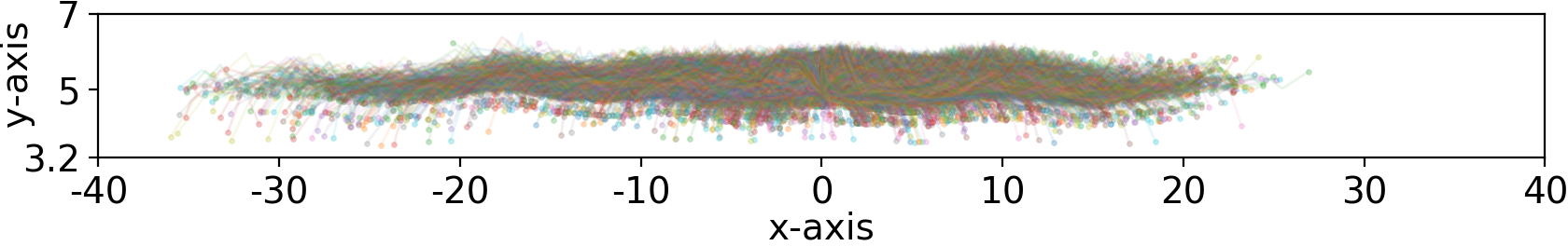}\\ 
            & 
            \includegraphics[width=0.658\textwidth]{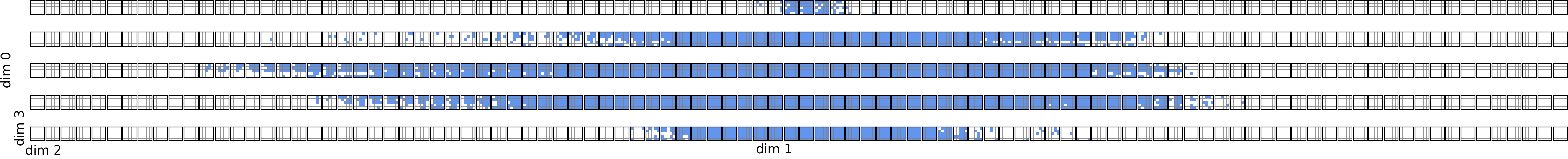}\\
            \hline
            %
            %
            \multirow{2}{*}{\textbf{PoMS-no-jacobian}}
            & 
            \includegraphics[width=0.658\textwidth]{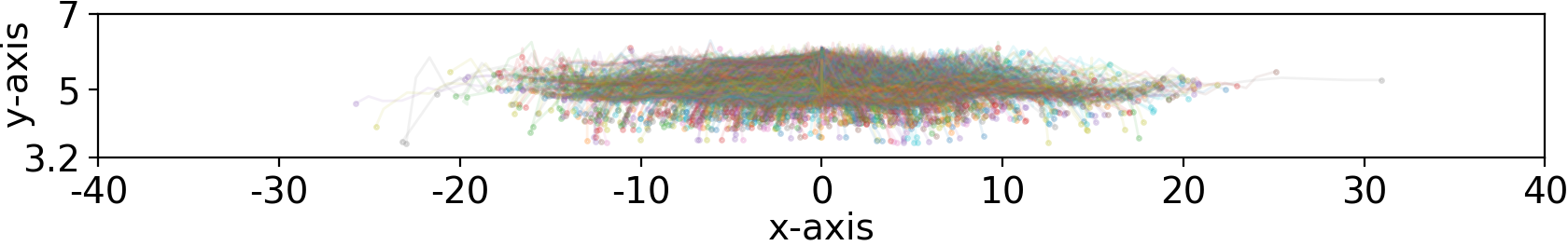}\\ 
            & 
            \includegraphics[width=0.658\textwidth]{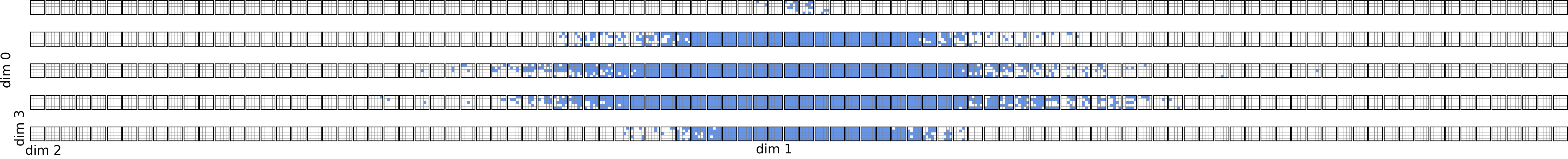}\\
            \hline
            \multirow{2}{*}{\textbf{DDE}}
            & 
            \includegraphics[width=0.658\textwidth]{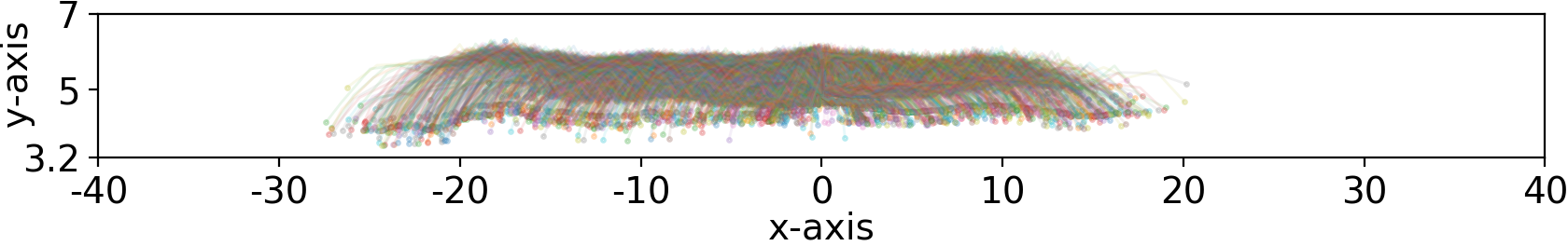} \\ 
            & 
            \includegraphics[width=0.658\textwidth]{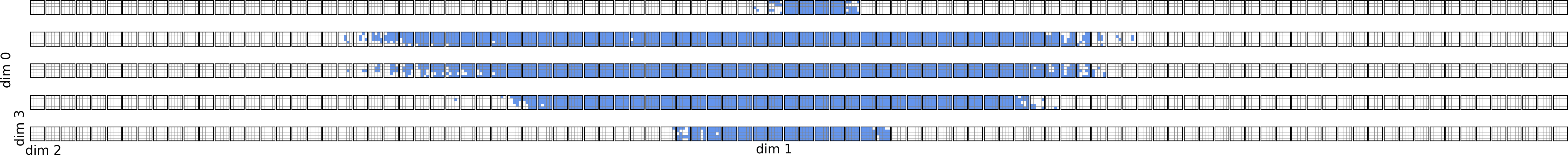}\\
            \hline
            %
            %
            \multirow{2}{*}{\textbf{MAPE-IsoLineDD}}
            & 
            \includegraphics[width=0.658\textwidth]{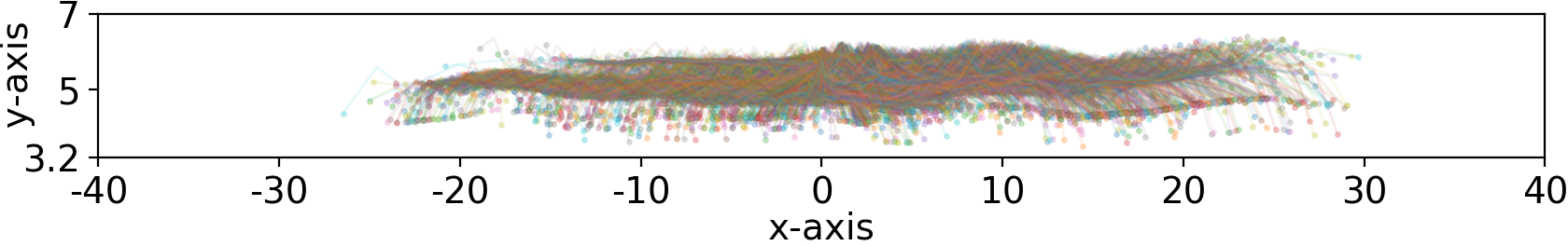}\\ 
            & 
            \includegraphics[width=0.658\textwidth]{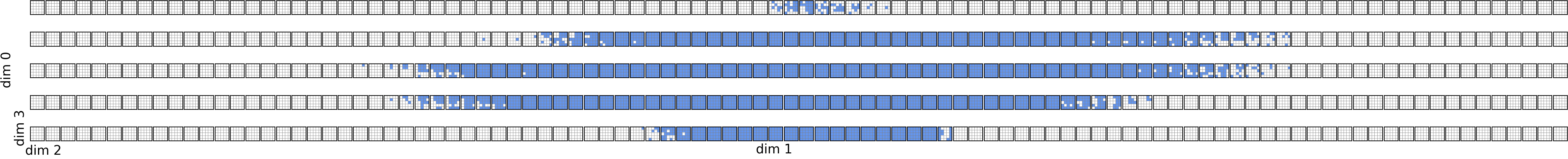} \\
            \hline
            \multirow{2}{*}{\textbf{MAPE-Iso}}
            & 
            \includegraphics[width=0.658\textwidth]{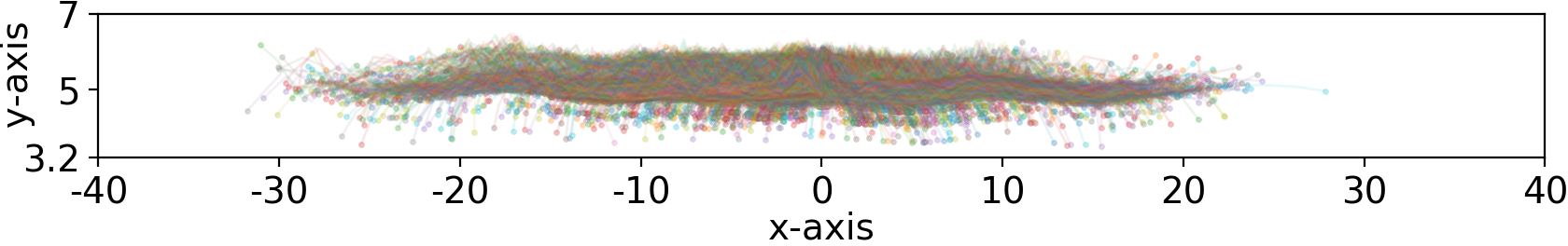}\\ 
            & 
            \includegraphics[width=0.658\textwidth]{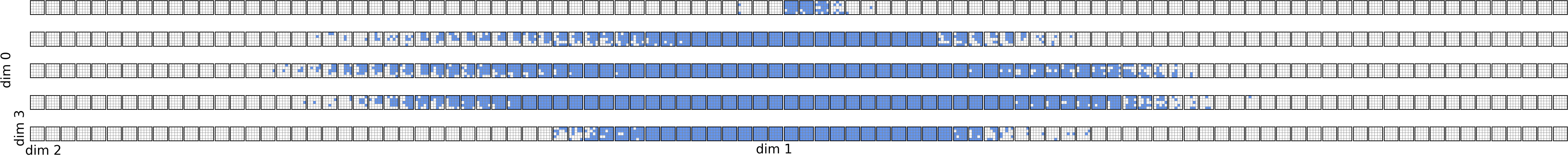}\\
            \hline
            \multirow{2}{*}{\textbf{ps-uniform}}
            & 
            \includegraphics[width=0.658\textwidth]{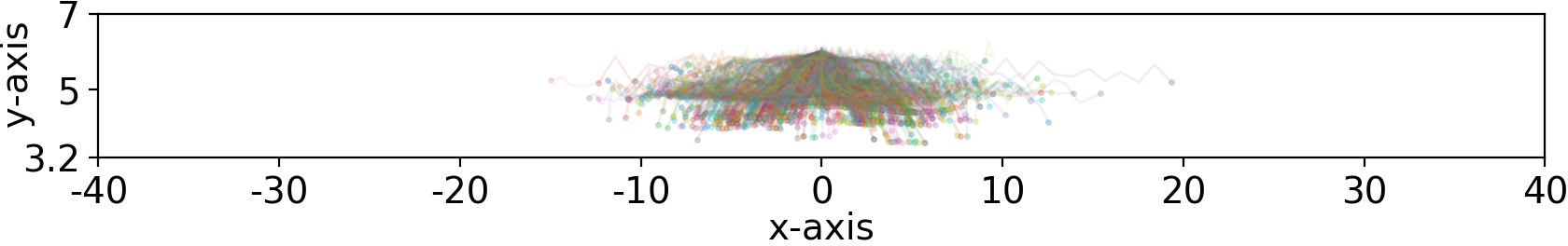} \\ 
            & 
            \includegraphics[width=0.658\textwidth]{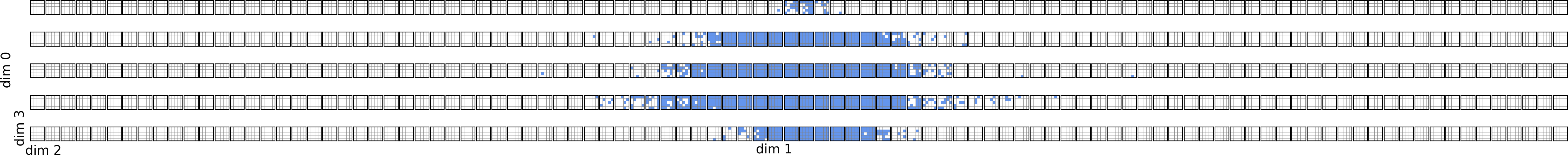}\\
            \hline
            \multirow{2}{*}{\textbf{ps-glorot}}
            & 
            \includegraphics[width=0.658\textwidth]{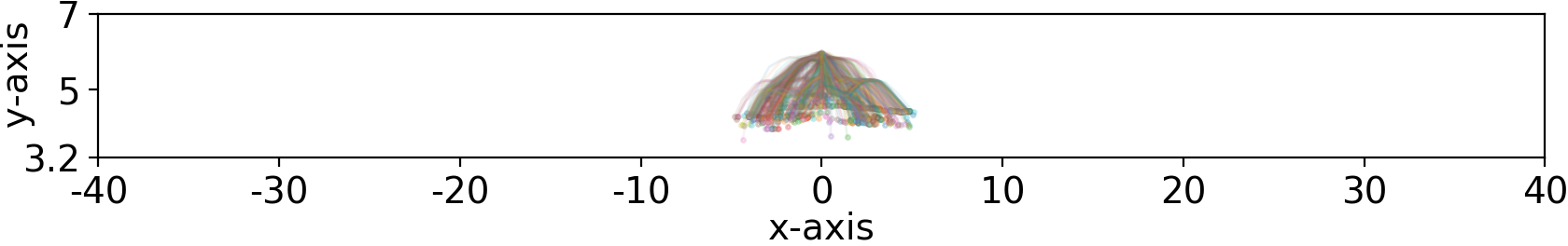}\\ 
            & 
            \includegraphics[width=0.658\textwidth]{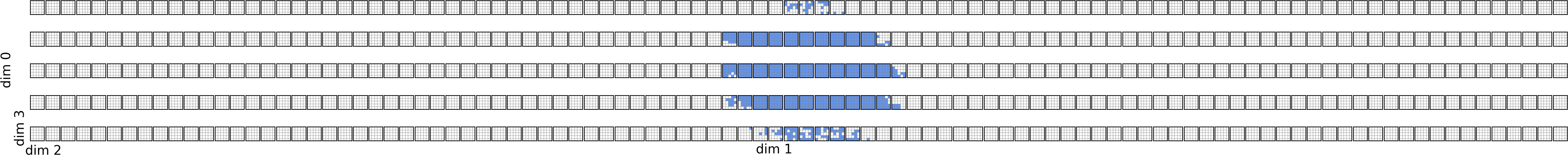} \\
            \hline
        \end{tabular}
    \end{table}
\newpage

    \begin{table}[h!]
    \centering
    \caption{Robot trajectories and policy collections for Bipedal-Walker mix-scale experiments.}
    \label{tab:bipedal_walker_mix}
        \begin{tabular}{ | m{0.2\textwidth} | m{0.7\textwidth} | }
            \hline
            \textbf{Algorithm}  & \multicolumn{1}{c|}{\textbf{Bipedal-Walker mix-scale}}   \\
            \hline
            \hline
            \multirow{2}{*}{\textbf{PoMS}}
            & 
            \includegraphics[width=0.658\textwidth]{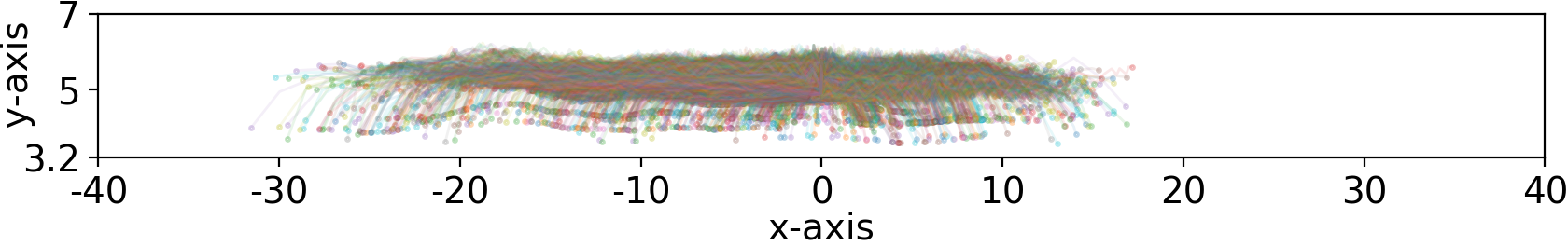} \\ 
            & 
            \includegraphics[width=0.658\textwidth]{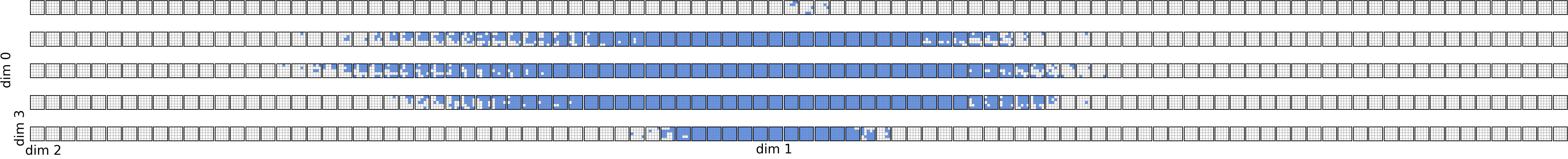} \\
            \hline
            \multirow{2}{*}{\textbf{PoMS-PCA}}
            & 
            \includegraphics[width=0.658\textwidth]{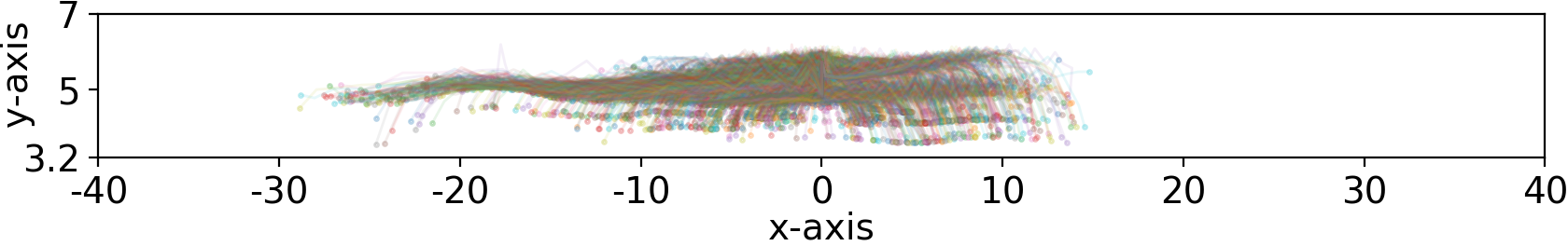} \\ 
            & 
            \includegraphics[width=0.658\textwidth]{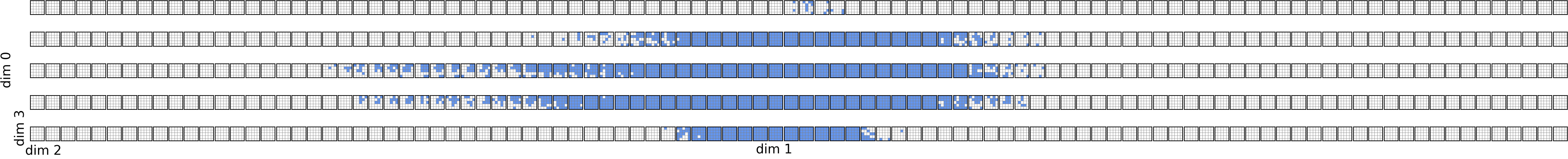} \\
            \hline
            %
            %
            \multirow{2}{*}{\textbf{PoMS-no-jacobian}}
            & 
            \includegraphics[width=0.658\textwidth]{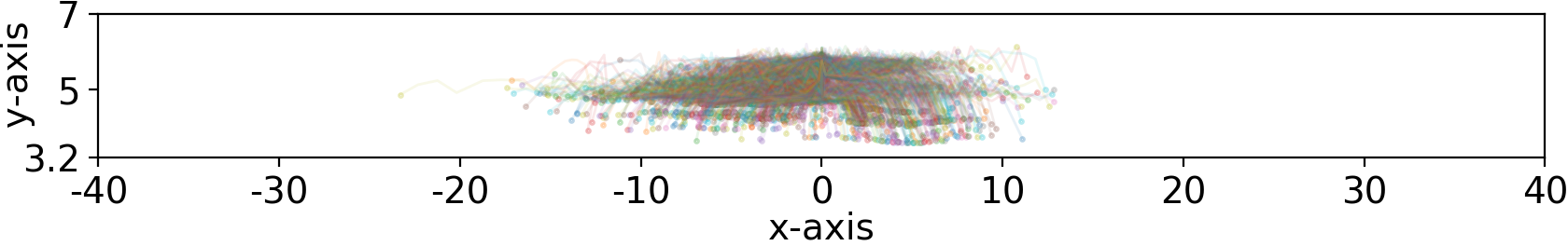} \\ 
            & 
            \includegraphics[width=0.658\textwidth]{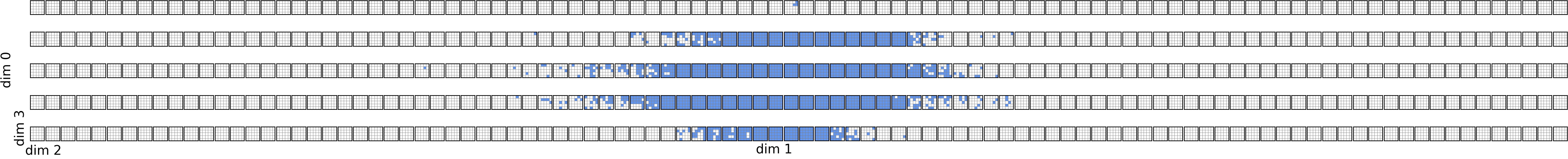} \\
            \hline
            \multirow{2}{*}{\textbf{DDE}}
            & 
            \includegraphics[width=0.658\textwidth]{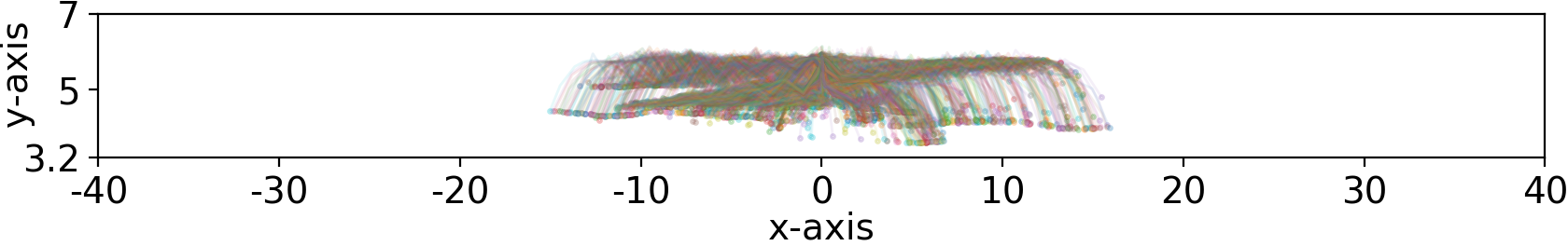} \\ 
            & 
            \includegraphics[width=0.658\textwidth]{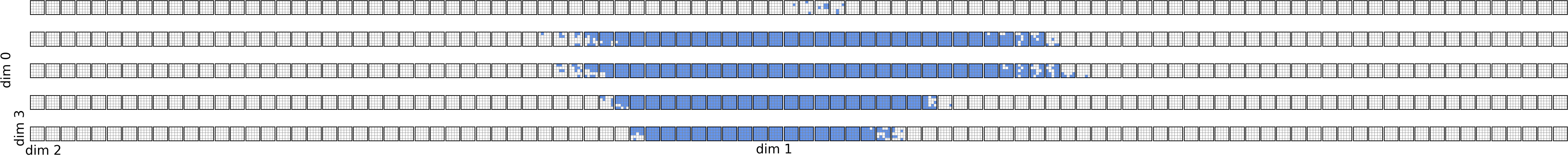}\\
            \hline
            %
            %
            \multirow{2}{*}{\textbf{MAPE-IsoLineDD}}
            & 
            \includegraphics[width=0.658\textwidth]{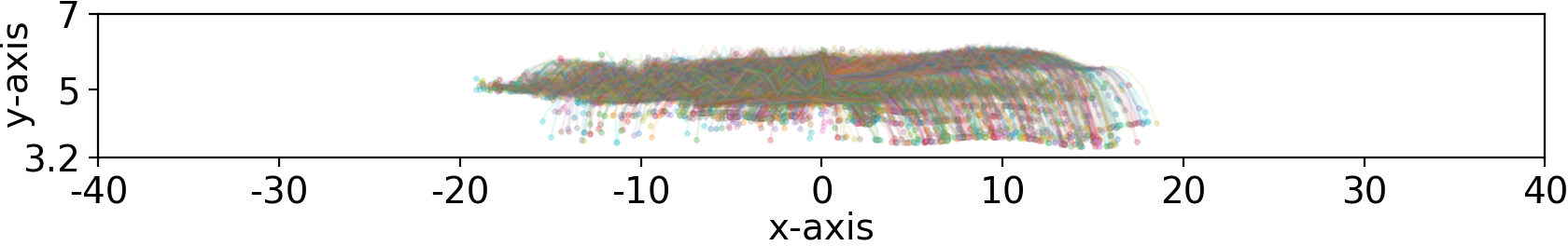} \\ 
            & 
            \includegraphics[width=0.658\textwidth]{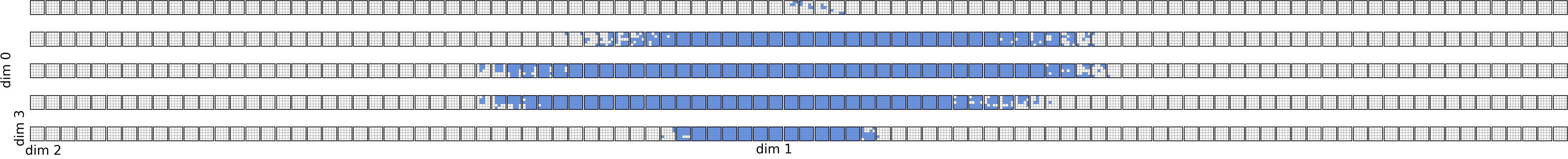} \\
            \hline
            \multirow{2}{*}{\textbf{MAPE-Iso}}
            & 
            \includegraphics[width=0.658\textwidth]{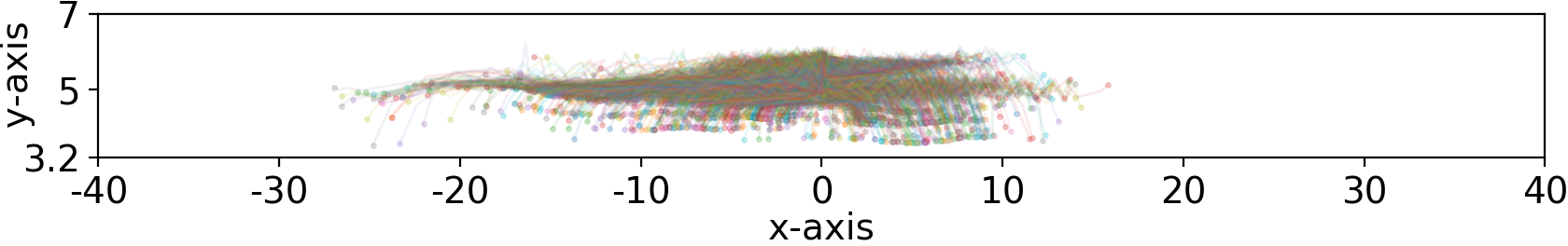} \\ 
            & 
            \includegraphics[width=0.658\textwidth]{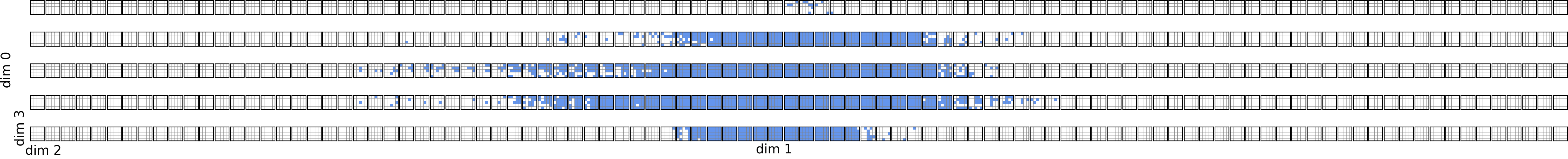} \\
            \hline
            \multirow{2}{*}{\textbf{ps-uniform}}
            & 
            \includegraphics[width=0.658\textwidth]{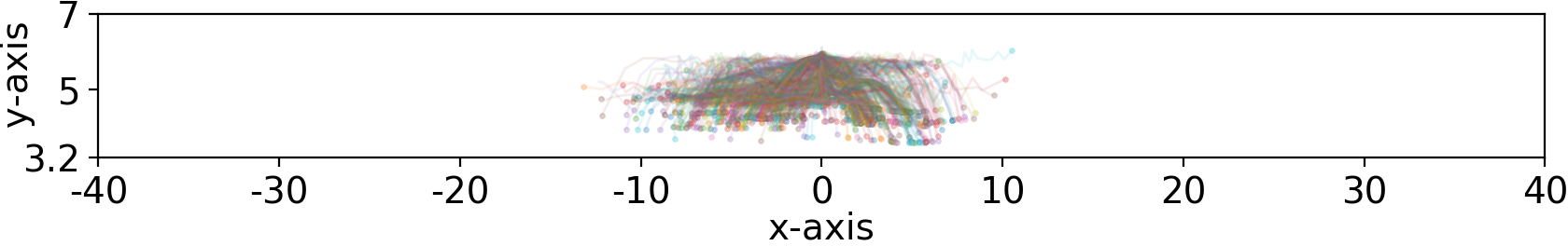} \\ 
            & 
            \includegraphics[width=0.658\textwidth]{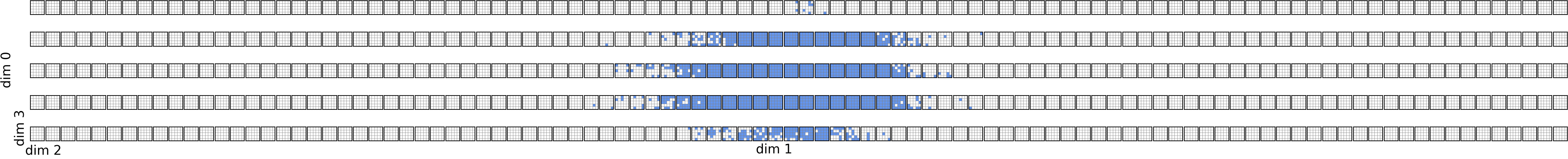} \\
            \hline
            \multirow{2}{*}{\textbf{ps-glorot}}
            & 
            \includegraphics[width=0.658\textwidth]{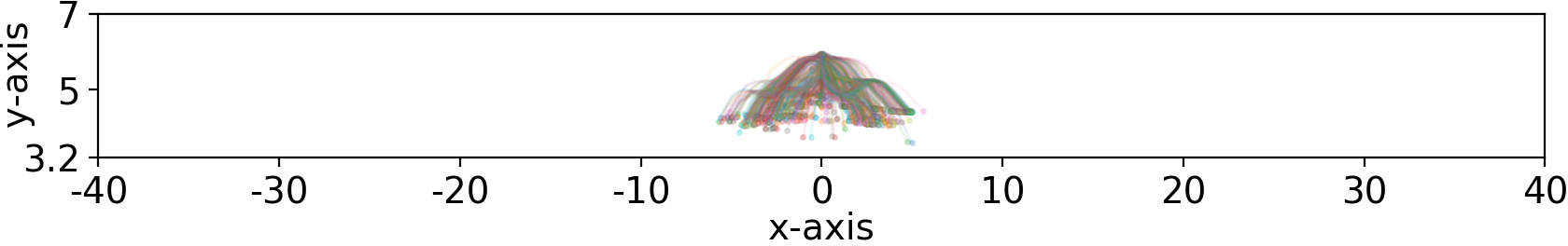} \\ 
            & 
            \includegraphics[width=0.658\textwidth]{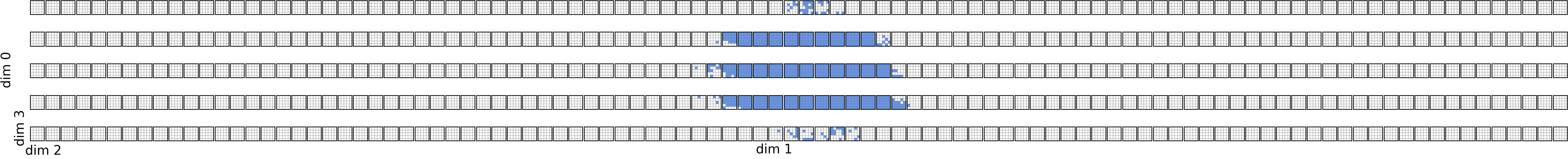} \\
            \hline
        \end{tabular}
    \end{table}
\newpage    
     
    
    \begin{table}[h!]
    \centering
    \caption{Ball trajectories and policy collections for Bipedal-Kicker experiments.}
    \label{tab:bipedal_kicker}
        \begin{tabular}{ |m{0.465\textwidth}  m{0.473\textwidth} | }
            \hline
            \multicolumn{1}{|l}{\textbf{Algorithm}} 
            & \\
            \multicolumn{2}{|c|}{\textbf{Bipedal-Kicker}}  \\
            \hline
            \hline
            \textbf{PoMS}
            & \\
            \includegraphics[width=0.49\textwidth]{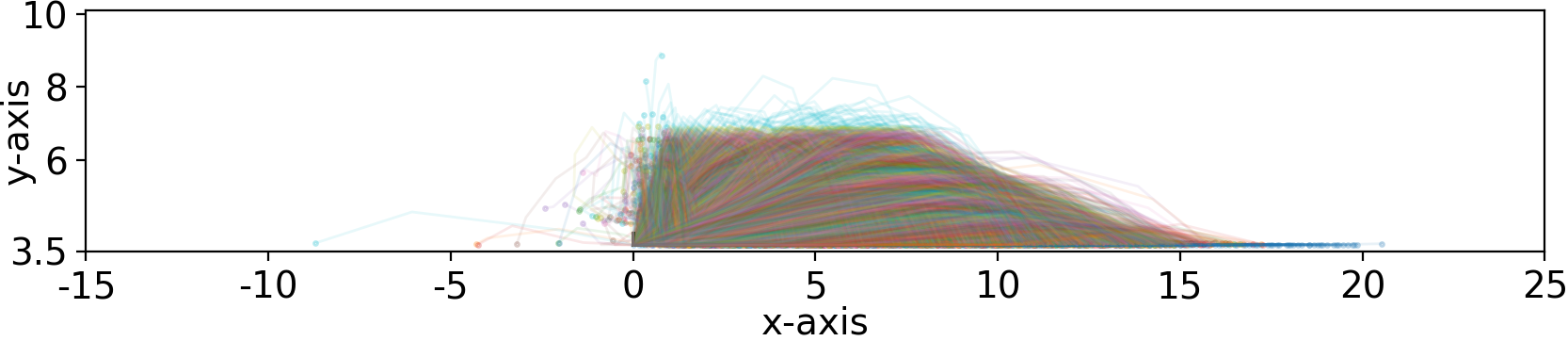}
            & 
            \includegraphics[width=0.483\textwidth]{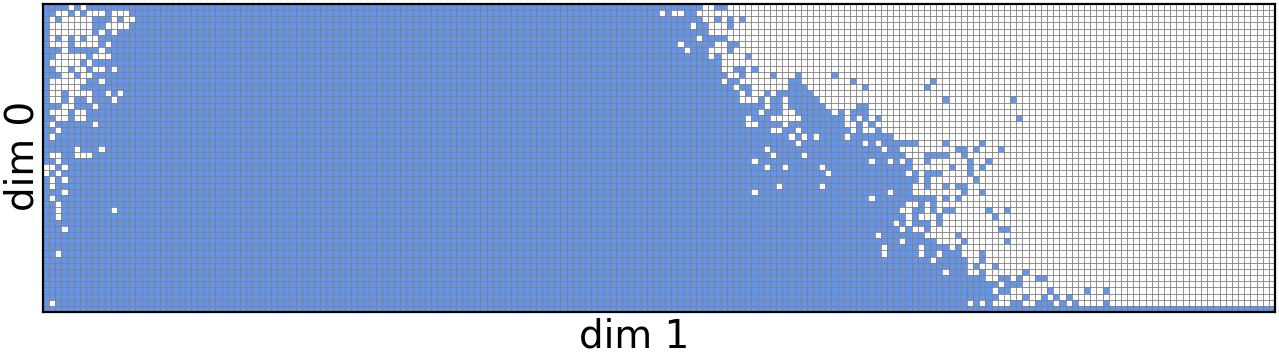}\\
            \hline
            \textbf{PoMS-PCA}
            & \\
            \includegraphics[width=0.49\textwidth]{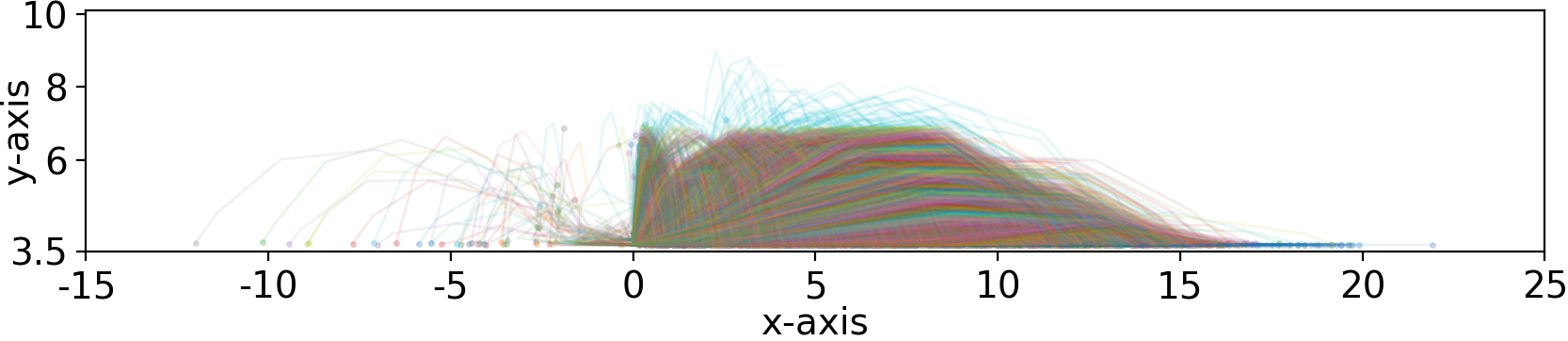}
            & 
            \includegraphics[width=0.483\textwidth]{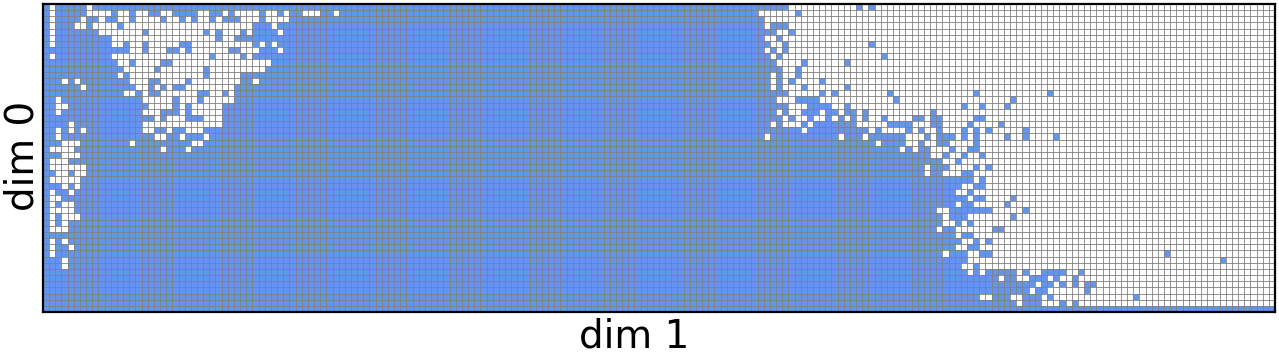}\\
            \hline
            %
            %
            \textbf{PoMS-no-jacobian}
            & \\
            \includegraphics[width=0.49\textwidth]{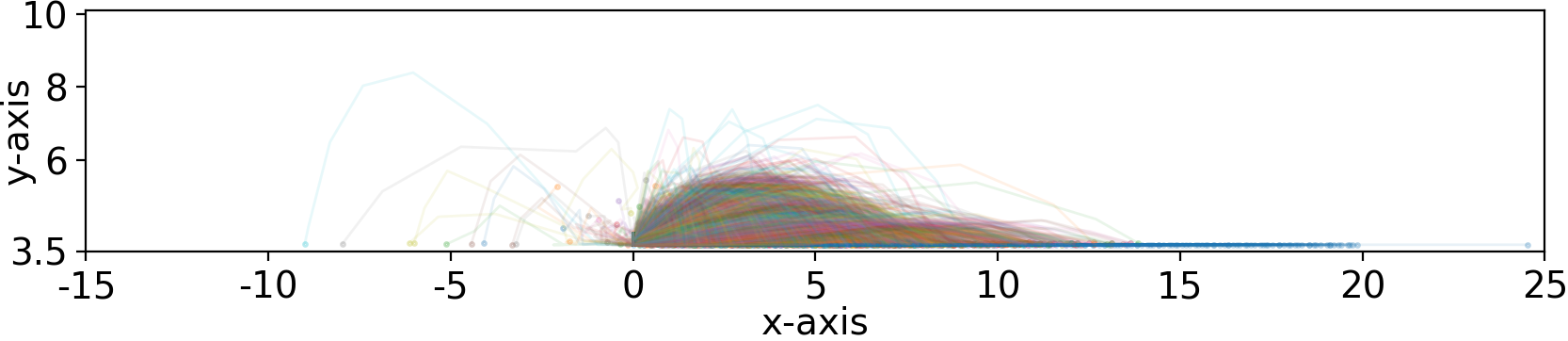}
            & 
            \includegraphics[width=0.483\textwidth]{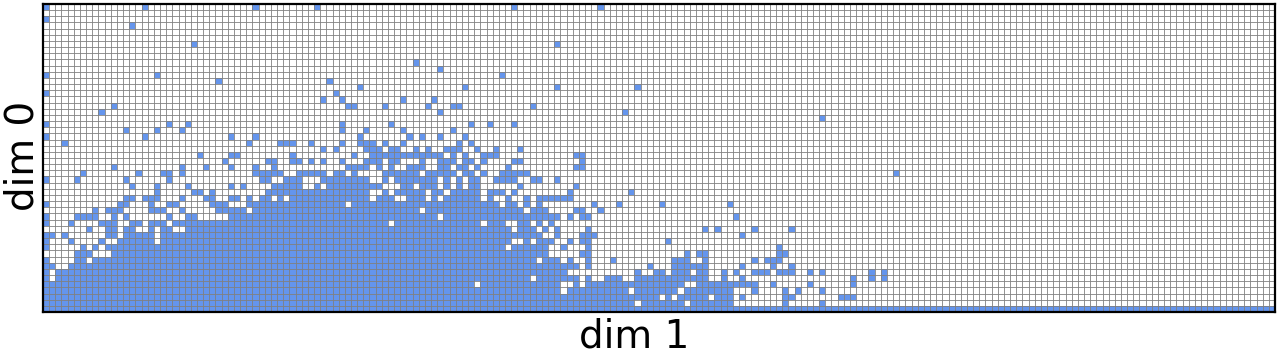} \\
            \hline
            \textbf{DDE}
            & \\
            \includegraphics[width=0.49\textwidth]{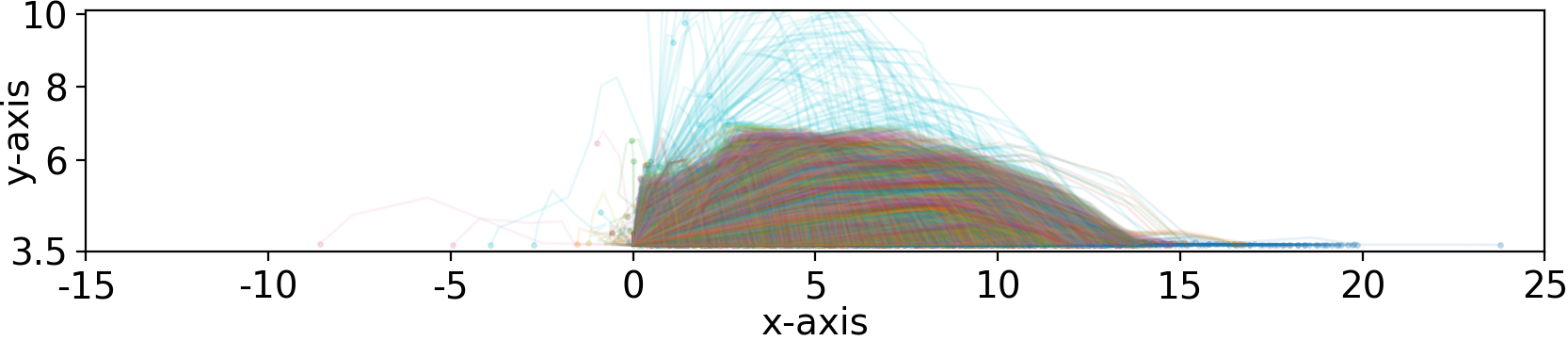}
            & 
            \includegraphics[width=0.483\textwidth]{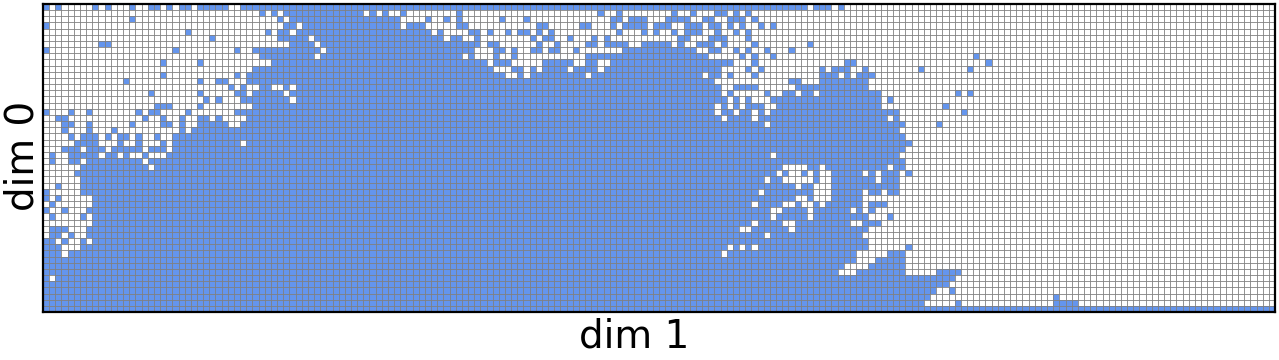}\\
            \hline
            %
            %
            \textbf{MAPE-IsoLineDD}
            & \\
            \includegraphics[width=0.49\textwidth]{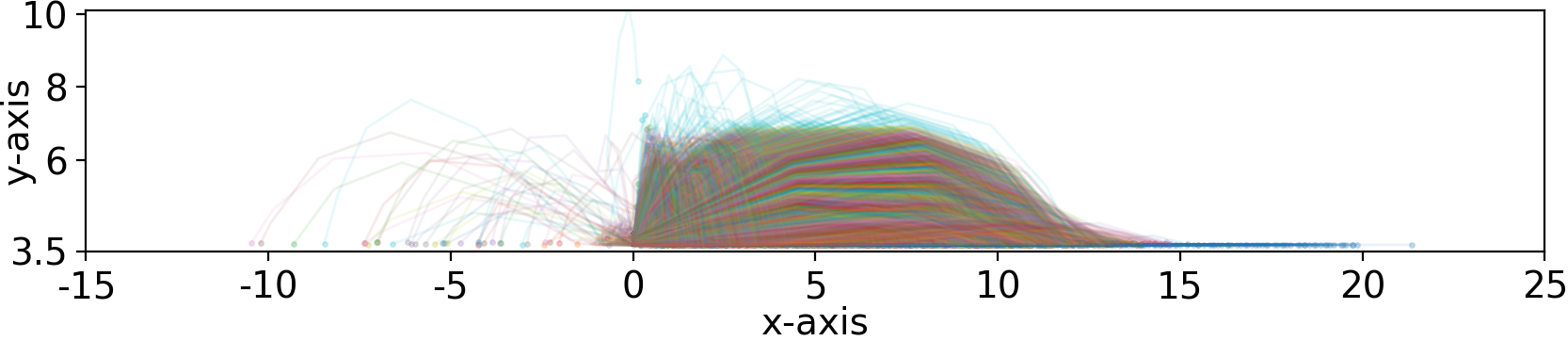} 
            & 
            \includegraphics[width=0.483\textwidth]{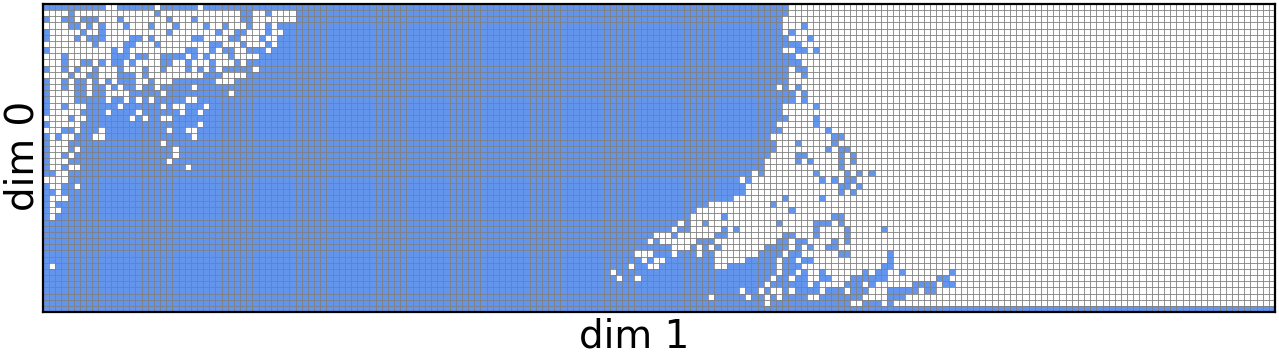}\\
            \hline
            \textbf{MAPE-Iso}
            & \\
            \includegraphics[width=0.49\textwidth]{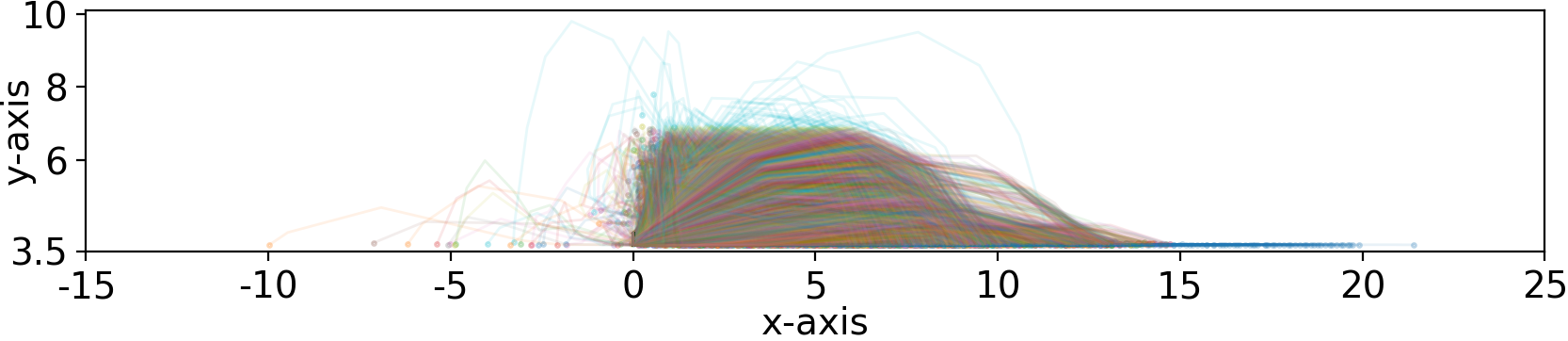}
            & 
            \includegraphics[width=0.483\textwidth]{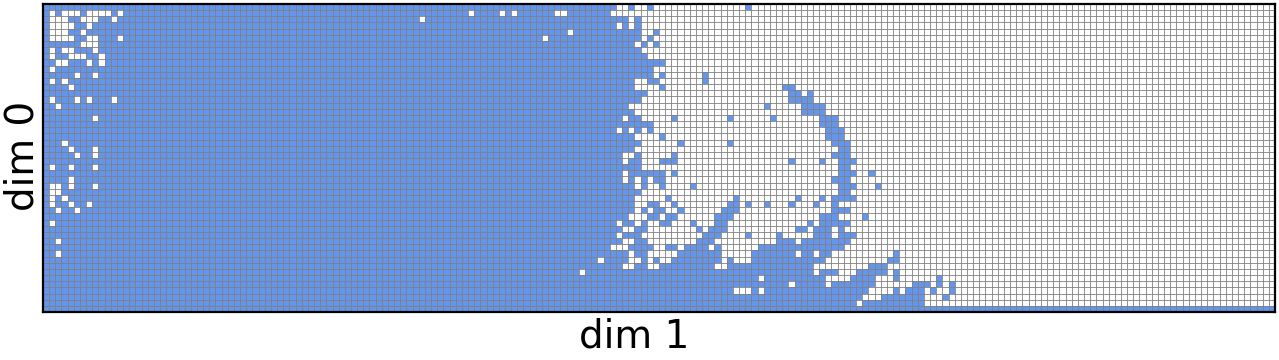} \\
            \hline
            \textbf{ps-uniform}
            & \\
            \includegraphics[width=0.49\textwidth]{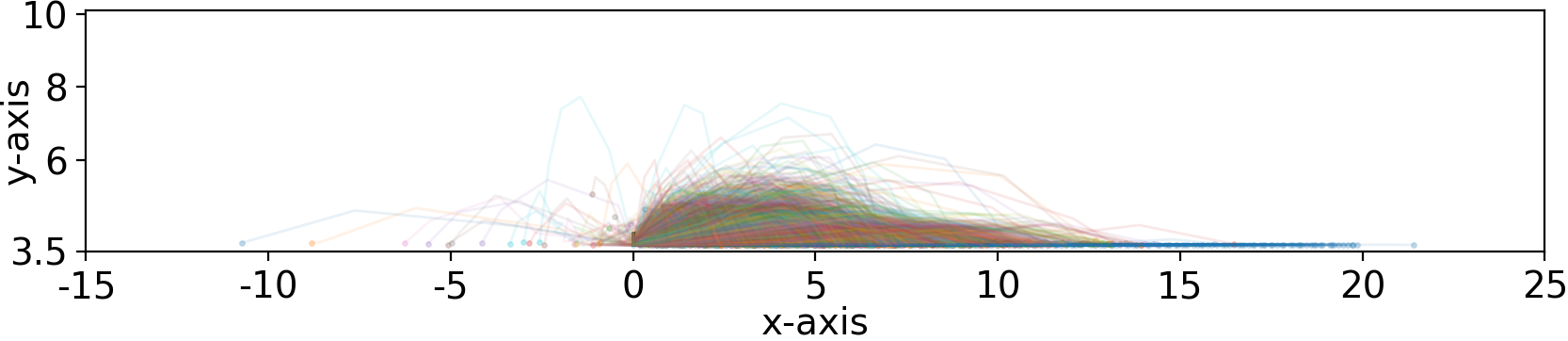}
            & 
            \includegraphics[width=0.483\textwidth]{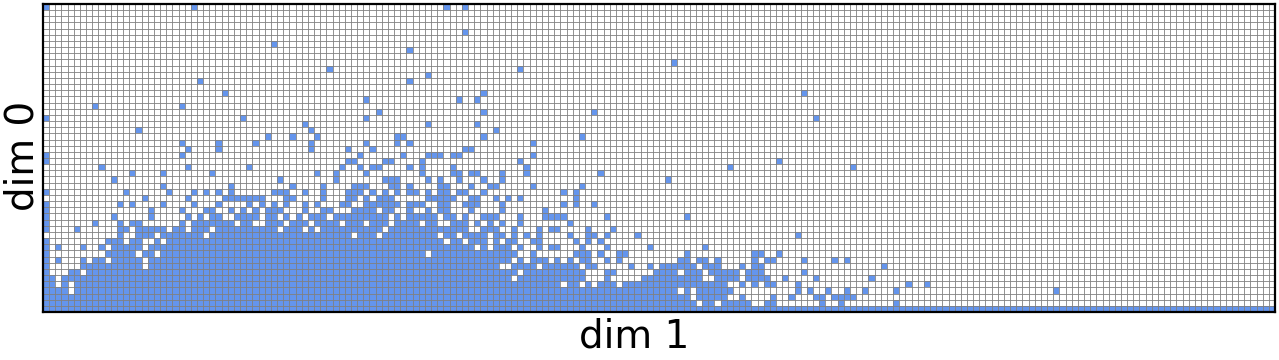} \\
            \hline
            \textbf{ps-glorot}
            & \\
            \includegraphics[width=0.49\textwidth]{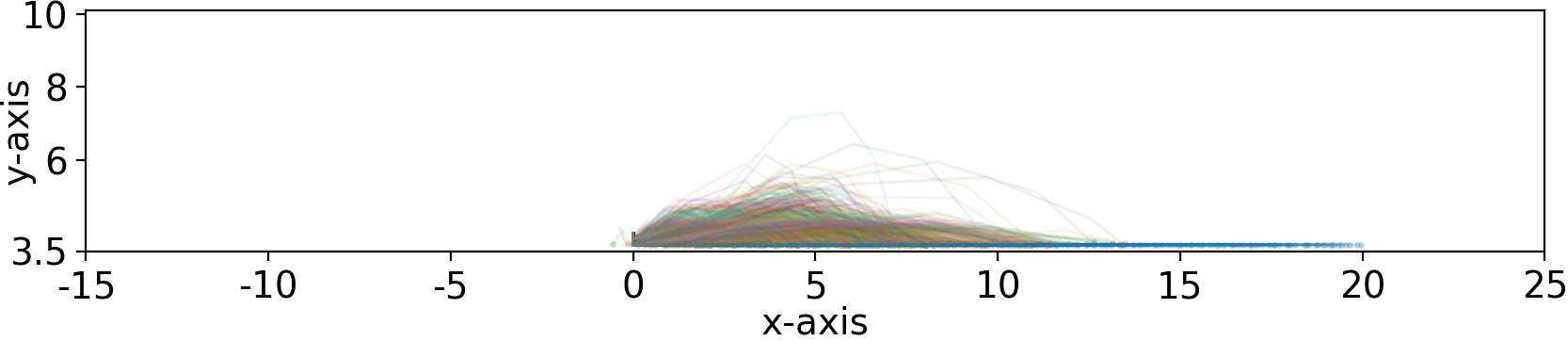}
            & 
            \includegraphics[width=0.483\textwidth]{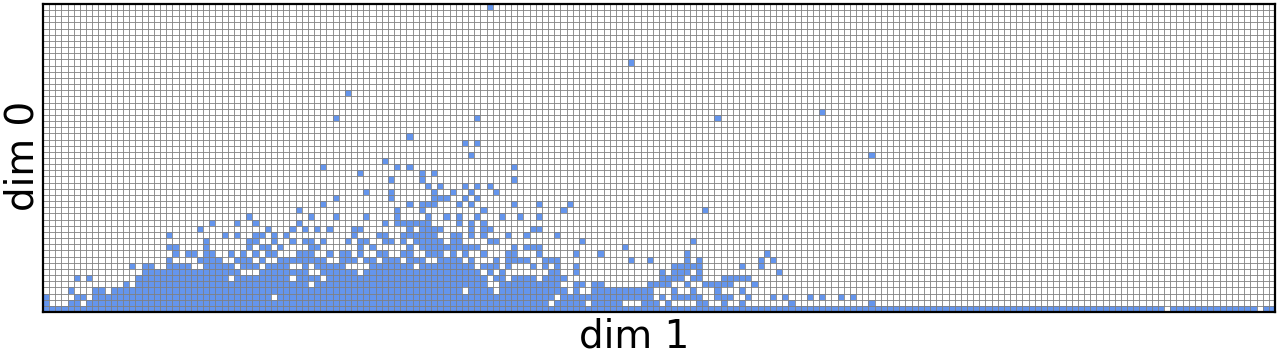} \\
            \hline
        \end{tabular}
    \end{table}
\newpage

    \begin{table}[h!]
    \centering
    \caption{Ball trajectories and policy collections for Bipedal-Kicker mix-scale experiments.}
    \label{tab:bipedal_kicker_mix}
        \begin{tabular}{ |m{0.465\textwidth}  m{0.473\textwidth} | }
            \hline
            \multicolumn{1}{|l}{\textbf{Algorithm}} 
            & \\
            \multicolumn{2}{|c|}{\textbf{Bipedal-Kicker mix-scale}}  \\
            \hline
            \hline
            \textbf{PoMS}
            & \\
            \includegraphics[width=0.49\textwidth]{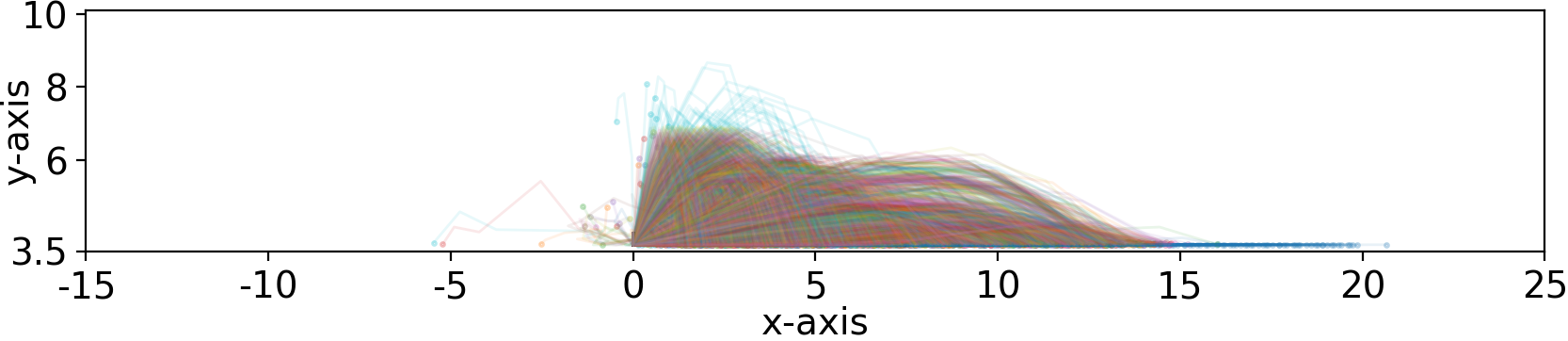} 
            & 
            \includegraphics[width=0.483\textwidth]{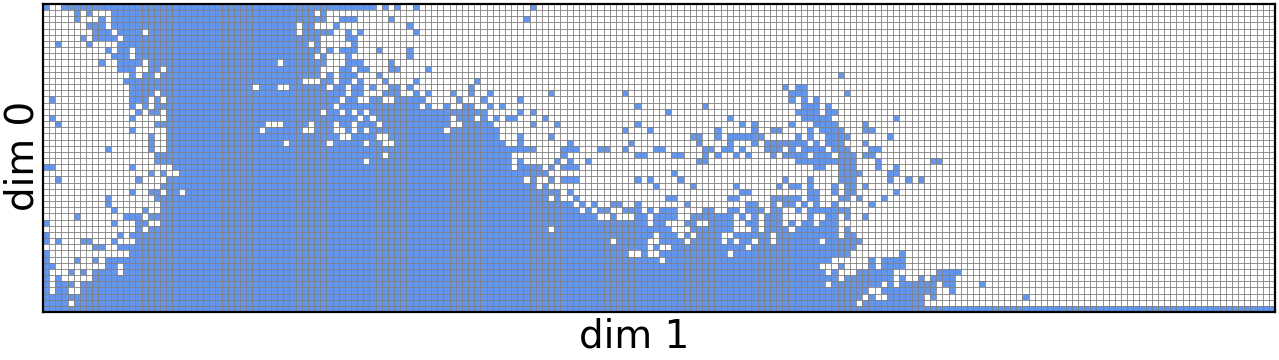} \\
            \hline
            \textbf{PoMS-PCA}
            & \\
            \includegraphics[width=0.49\textwidth]{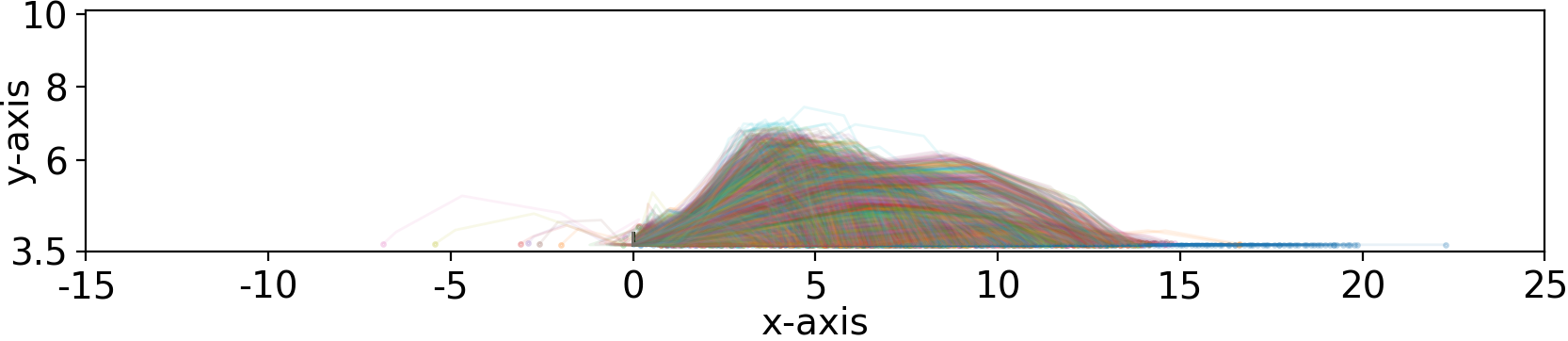} 
            & 
            \includegraphics[width=0.483\textwidth]{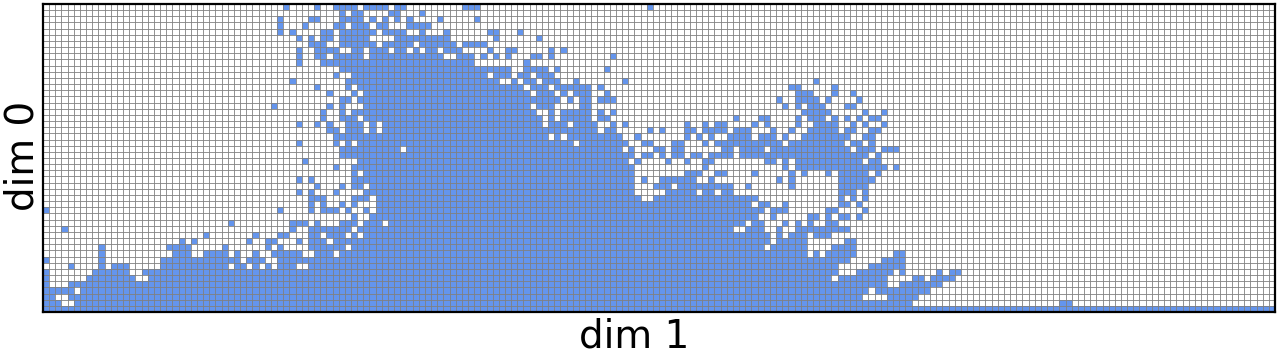} \\
            \hline
            %
            %
            \textbf{PoMS-no-jacobian}
            & \\
            \includegraphics[width=0.49\textwidth]{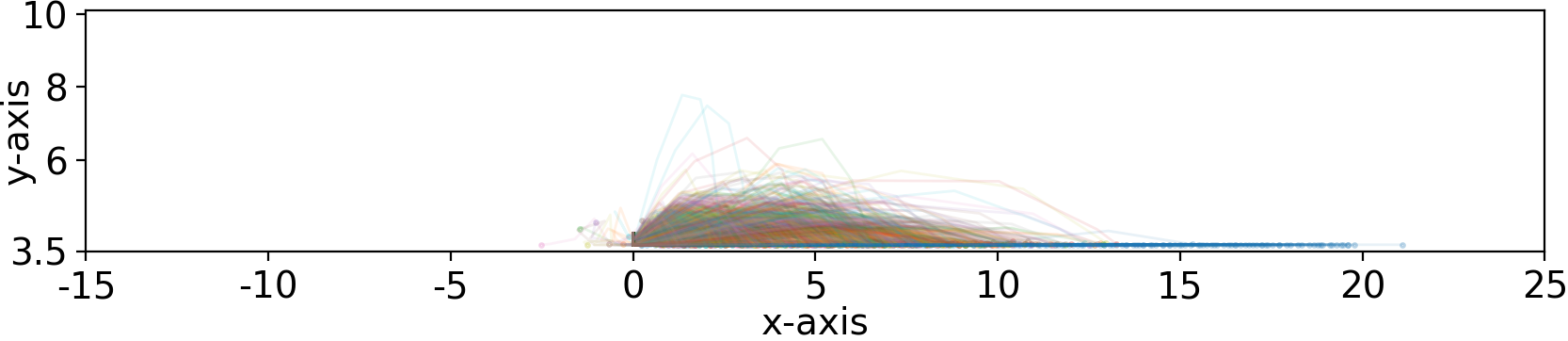}
            & 
            \includegraphics[width=0.483\textwidth]{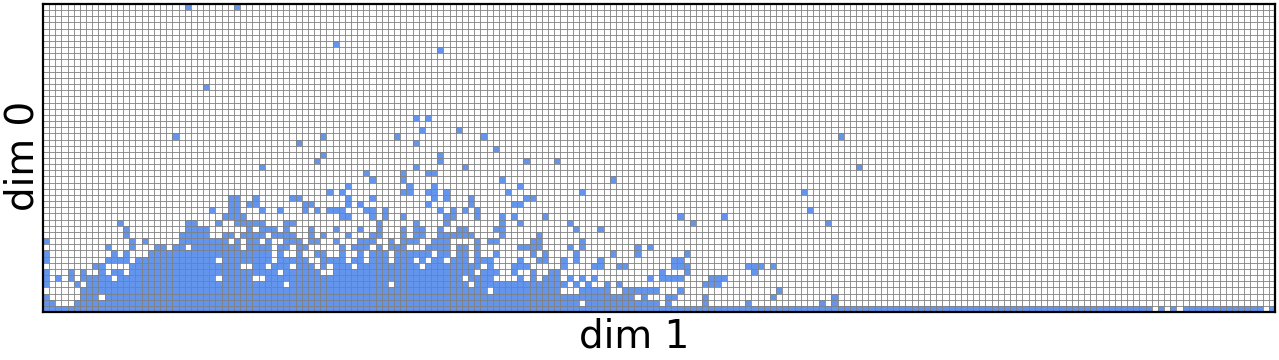} \\
            \hline
            \textbf{DDE}
            & \\
            \includegraphics[width=0.49\textwidth]{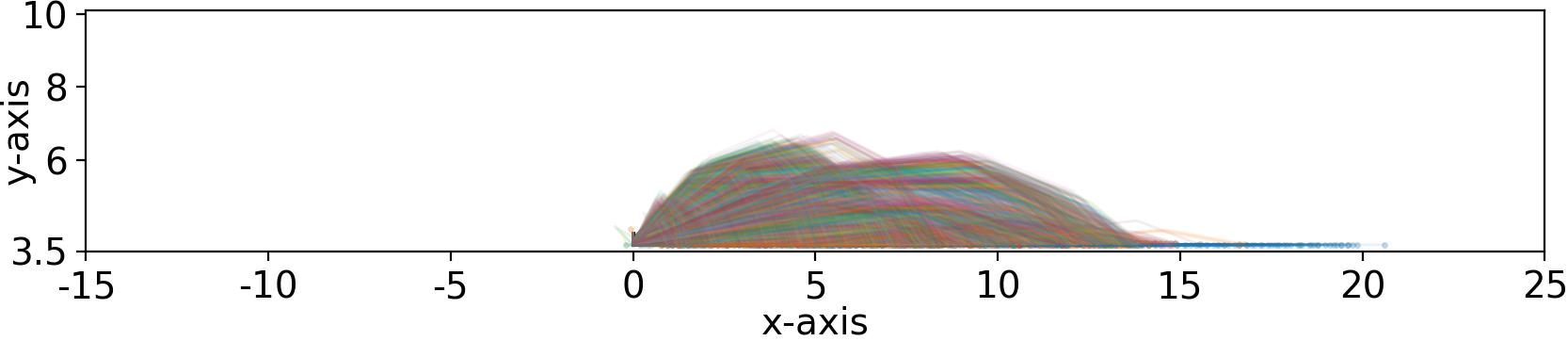}
            & 
            \includegraphics[width=0.483\textwidth]{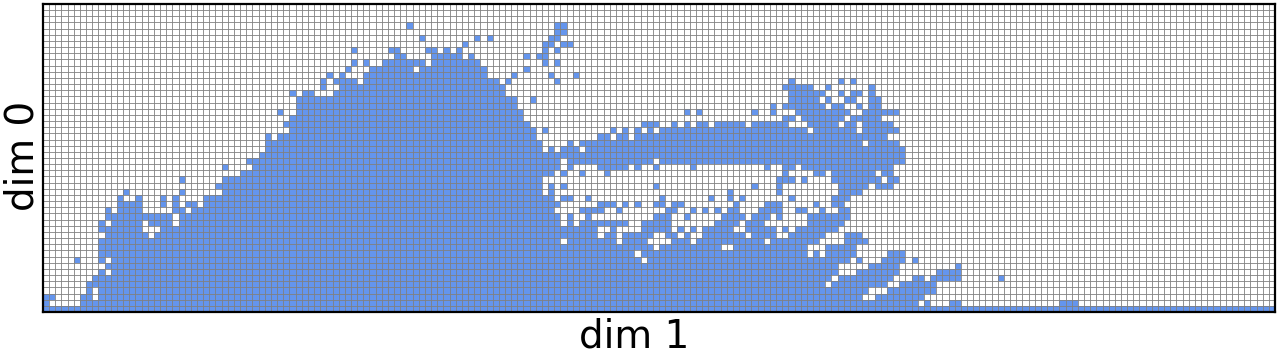} \\
            \hline
            %
            %
            \textbf{MAPE-IsoLineDD}
            & \\
            \includegraphics[width=0.49\textwidth]{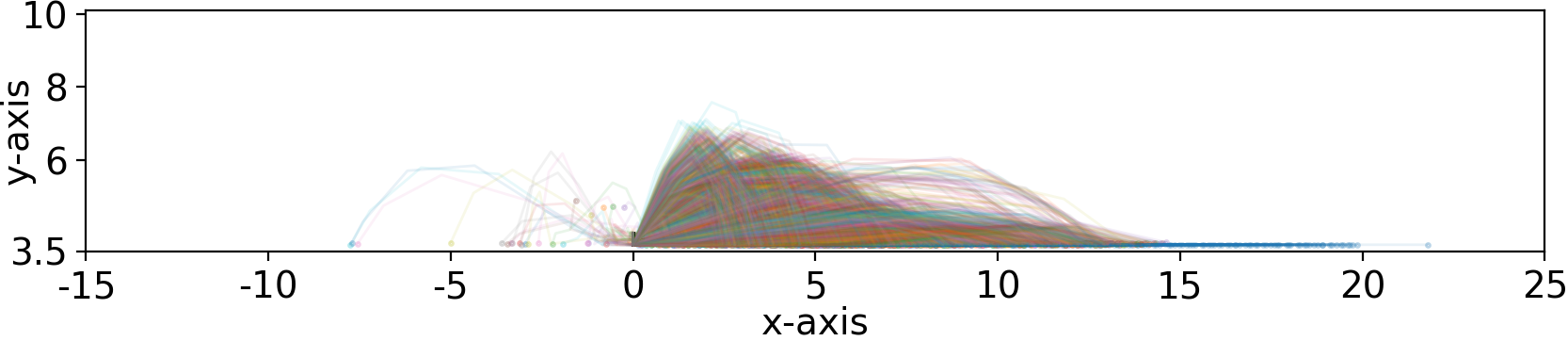} 
            & 
            \includegraphics[width=0.483\textwidth]{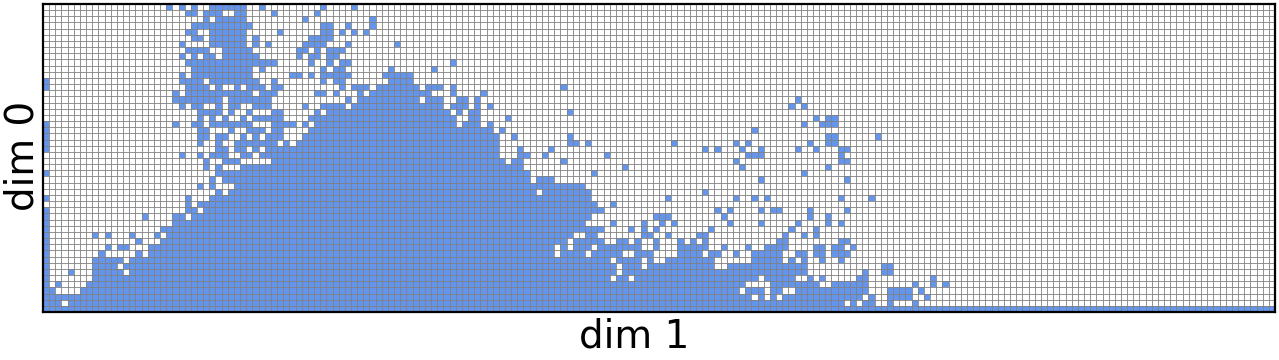} \\
            \hline
            \textbf{MAPE-Iso}
            & \\
            \includegraphics[width=0.49\textwidth]{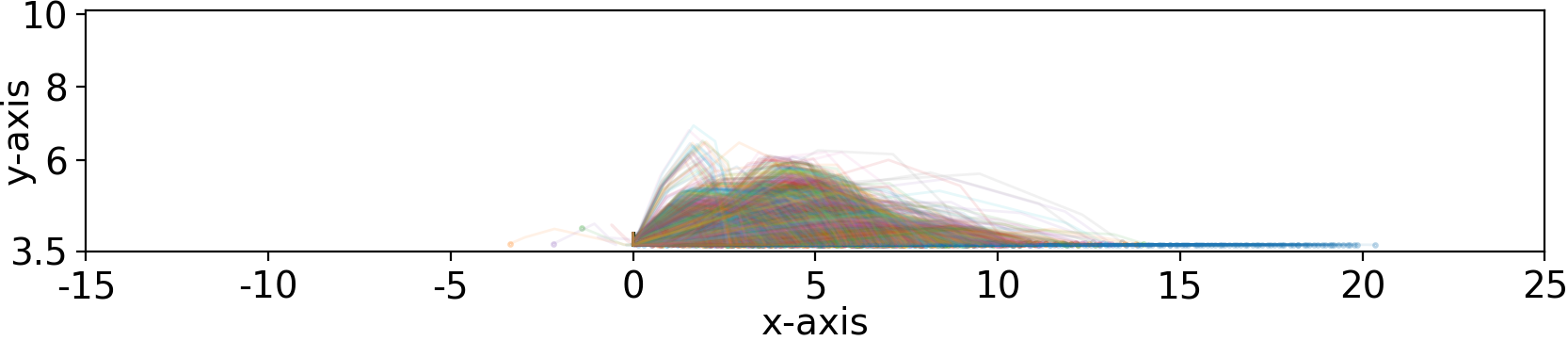}
            & 
            \includegraphics[width=0.483\textwidth]{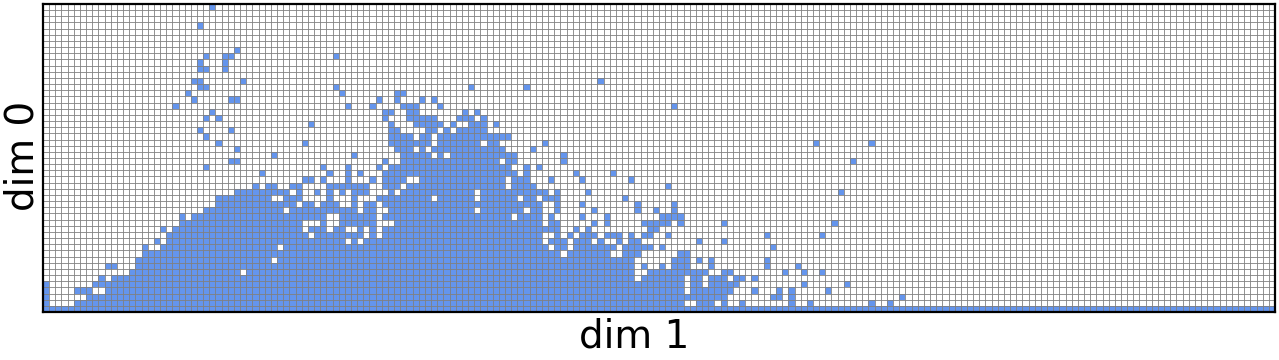} \\
            \hline
            \hline
            \textbf{ps-uniform}
            & \\
            \includegraphics[width=0.49\textwidth]{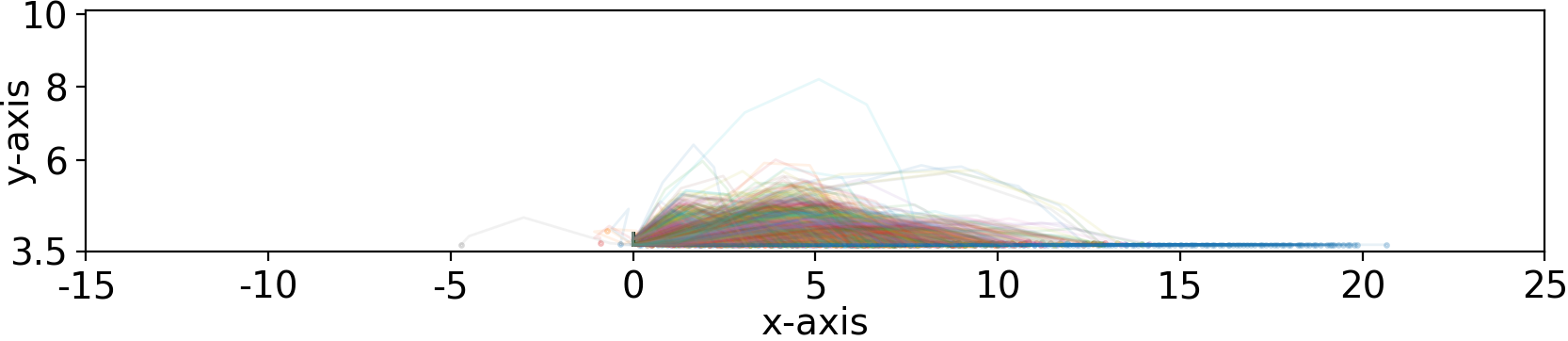}
            & 
            \includegraphics[width=0.483\textwidth]{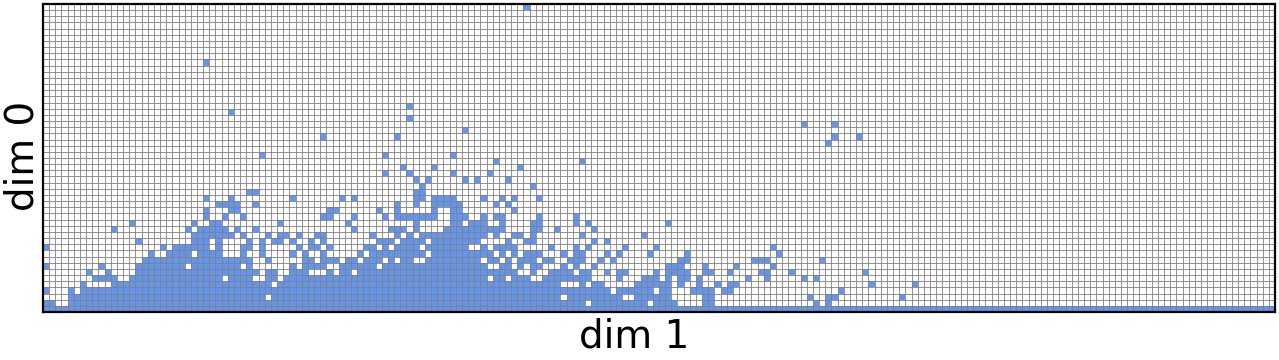} \\
            \hline
            \textbf{ps-glorot}
            & \\
            \includegraphics[width=0.49\textwidth]{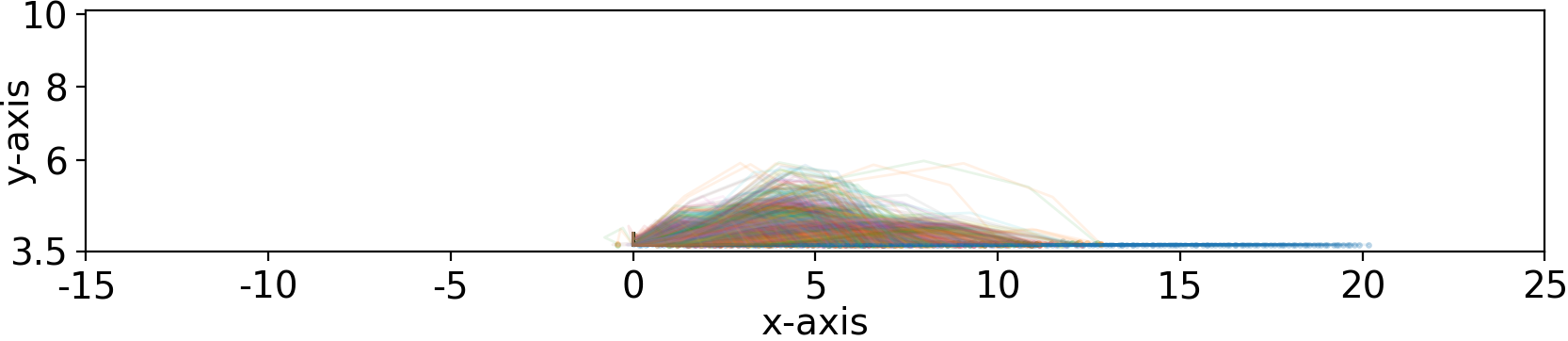}
            & 
            \includegraphics[width=0.483\textwidth]{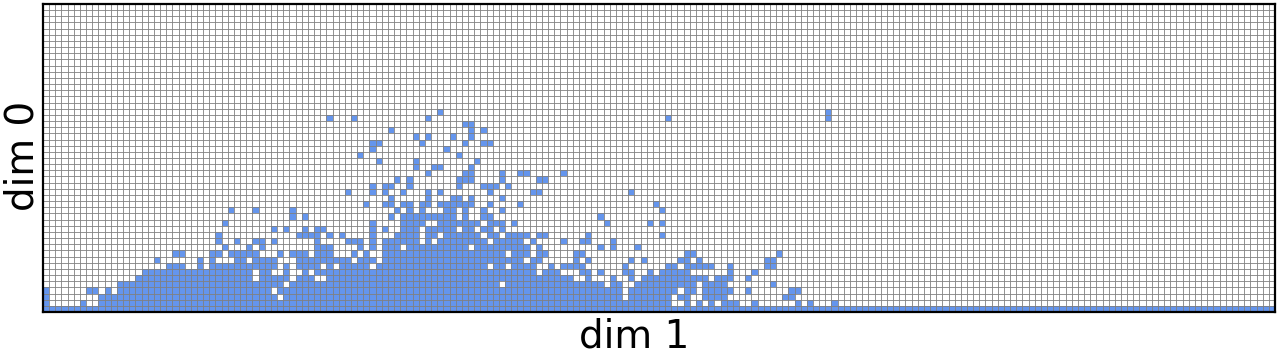} \\
            \hline
        \end{tabular}
    \end{table}
\newpage

\subsection{Parameter distributions and learned representations}
\begin{figure}[h!]
            \subfloat[PoMS]{\includegraphics[width=0.475\textwidth]
            {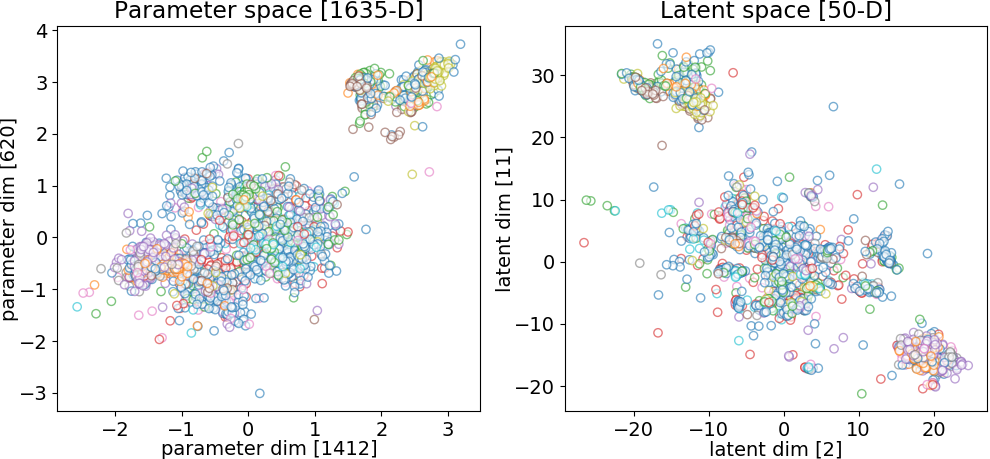}
            \label{fig:rep:striker:poms}} 
            \hfill
            \subfloat[PoMS-no-jacobian]{\includegraphics[width=0.5\textwidth]
            {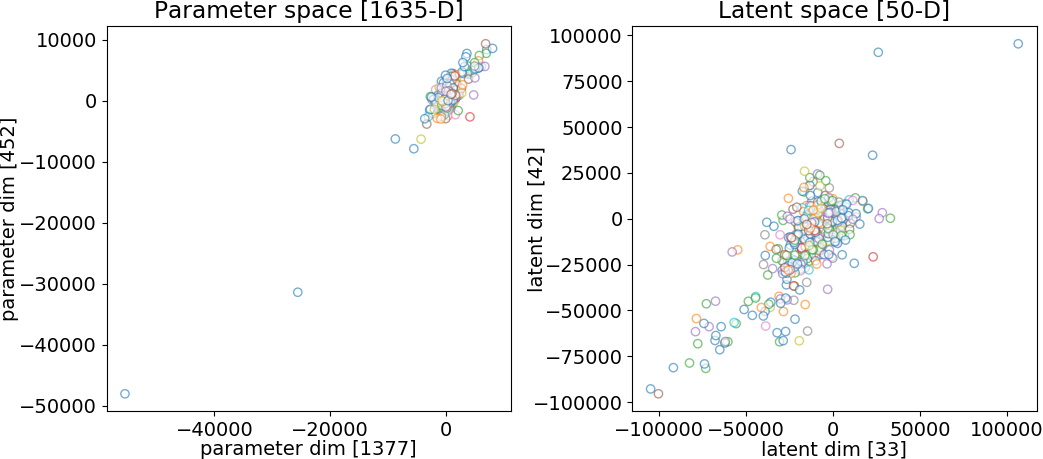}
            \label{fig:rep:striker:poms_no}} 
    \caption{Comparison of parameter and latent spaces for Striker experiment.}
    \label{fig:rep:striker}
    \end{figure}

\begin{figure}[h!]
            \subfloat[PoMS]{\includegraphics[width=0.49\textwidth]
            {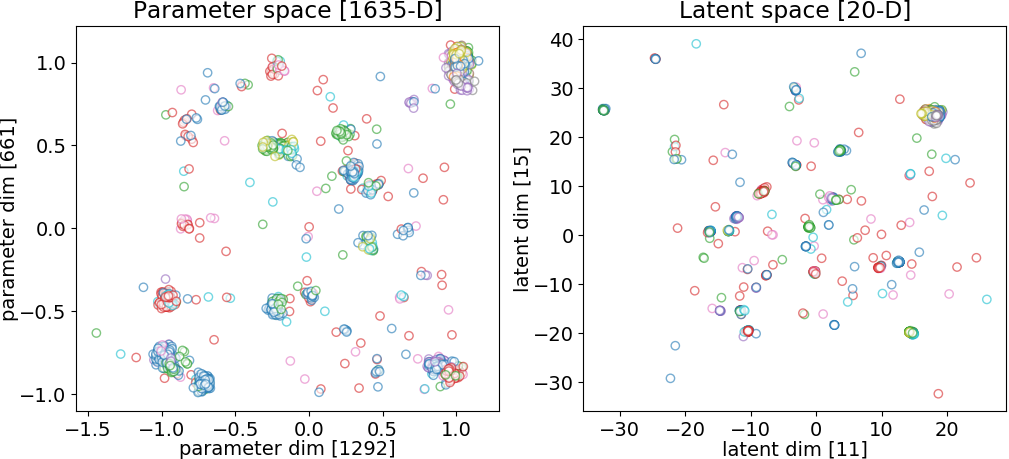}
            \label{fig:rep:striker_mix:poms}} 
            \hfill
            \subfloat[PoMS-no-jacobian]{\includegraphics[width=0.495\textwidth]
            {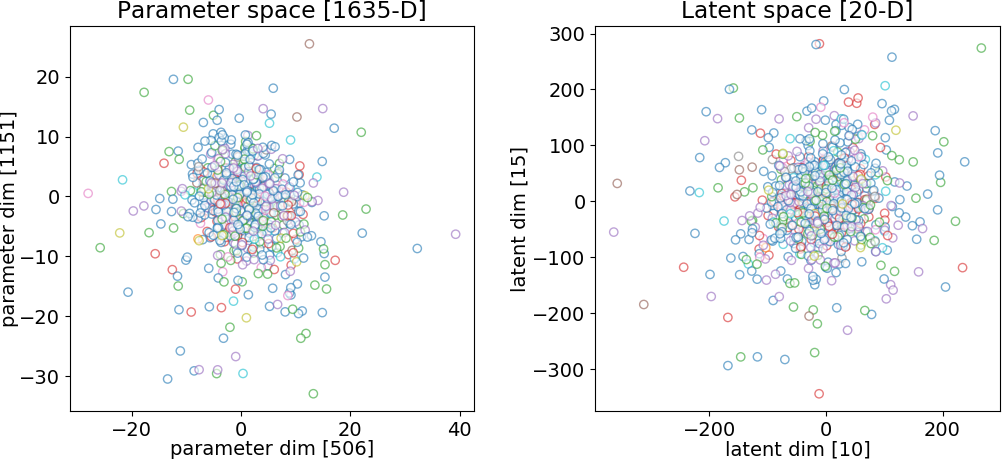}
            \label{fig:rep:striker_mix:poms_no}} 
    \caption{Comparison of parameter and latent spaces for Striker mix-scale experiment.}
    \label{fig:rep:striker_mix}
    \end{figure}

\begin{figure}[h!]
            \subfloat[PoMS]{\includegraphics[width=0.48\textwidth]
            {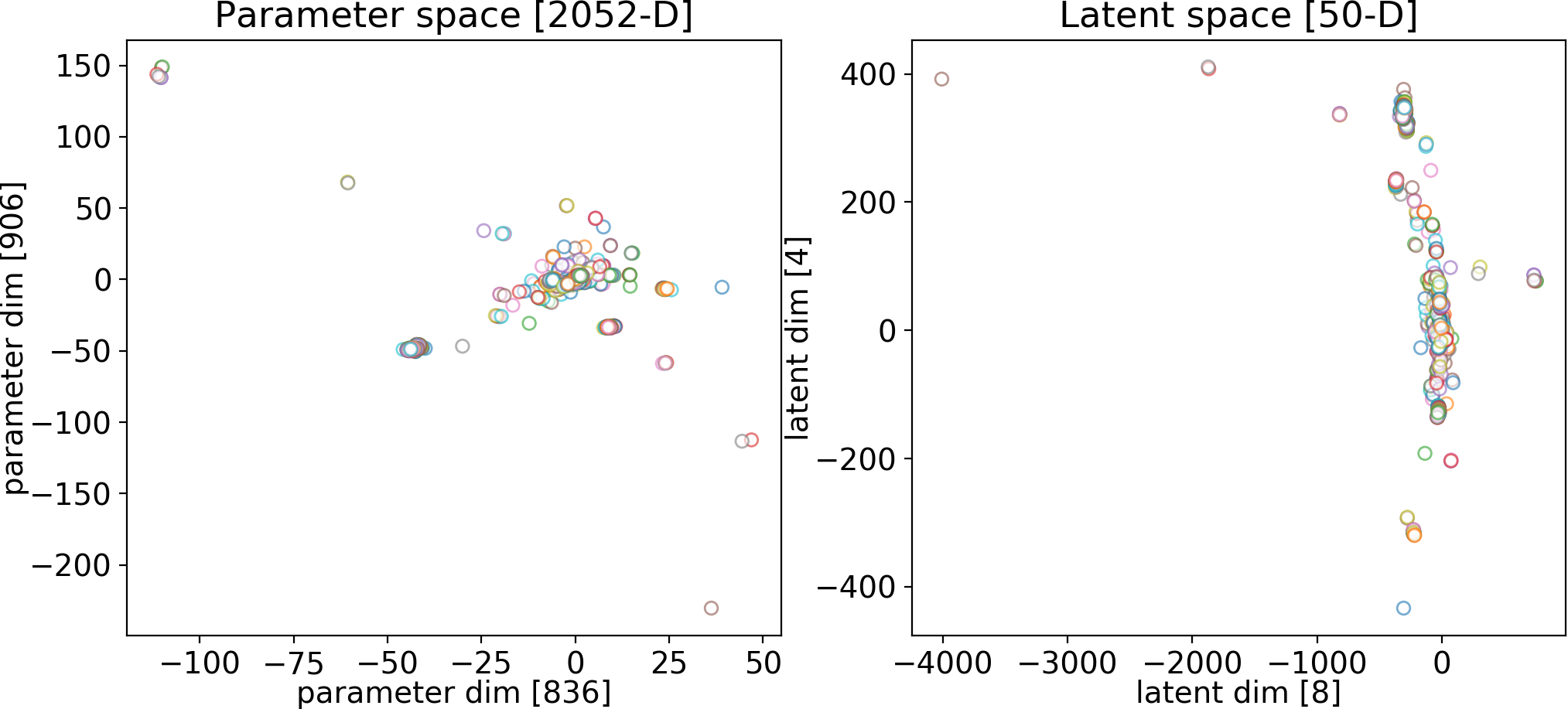}
            \label{fig:rep:bipedal_walker:poms}} 
            \hfill
            \subfloat[PoMS-no-jacobian]{\includegraphics[width=0.5\textwidth]
            {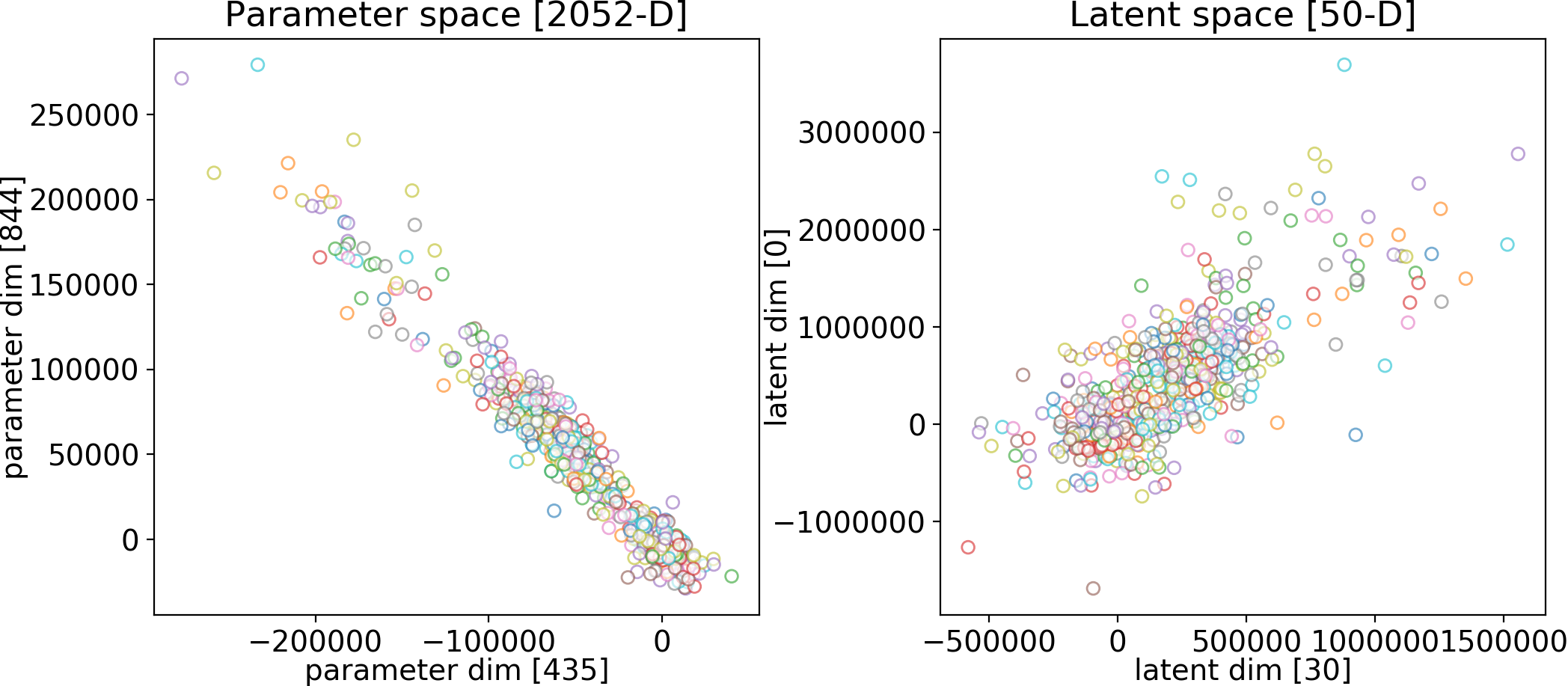}
            \label{fig:rep:bipedal_walker:poms_no}} 
    \caption{Comparison of parameter and latent spaces for Bipedal-Walker experiment.}
    \label{fig:rep:bipedal_walker}
    \end{figure}

\begin{figure}[h!]
            \subfloat[PoMS]{\includegraphics[width=0.49\textwidth]
            {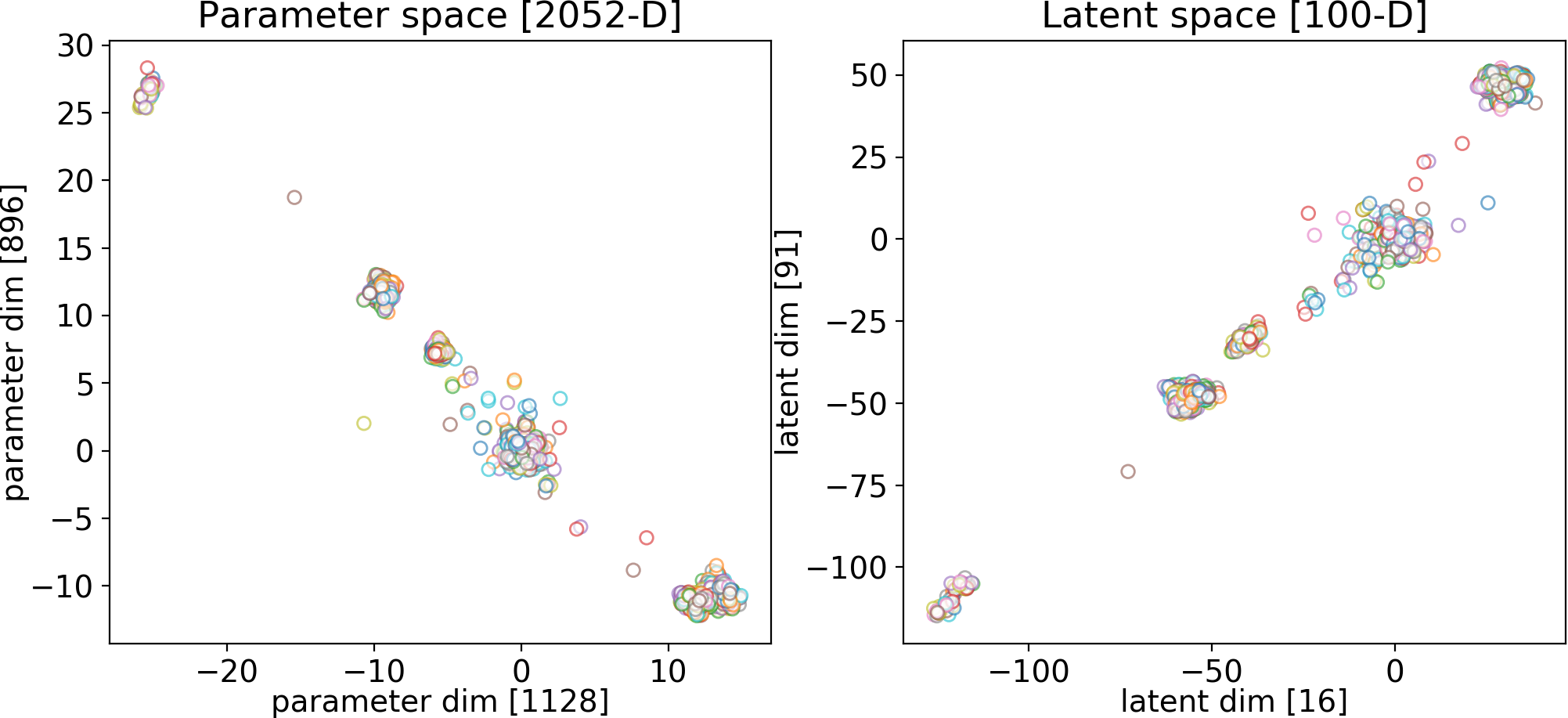}
            \label{fig:rep:bipedal_walker_mix:poms}} 
            \hfill
            \subfloat[PoMS-no-jacobian]{\includegraphics[width=0.495\textwidth]
            {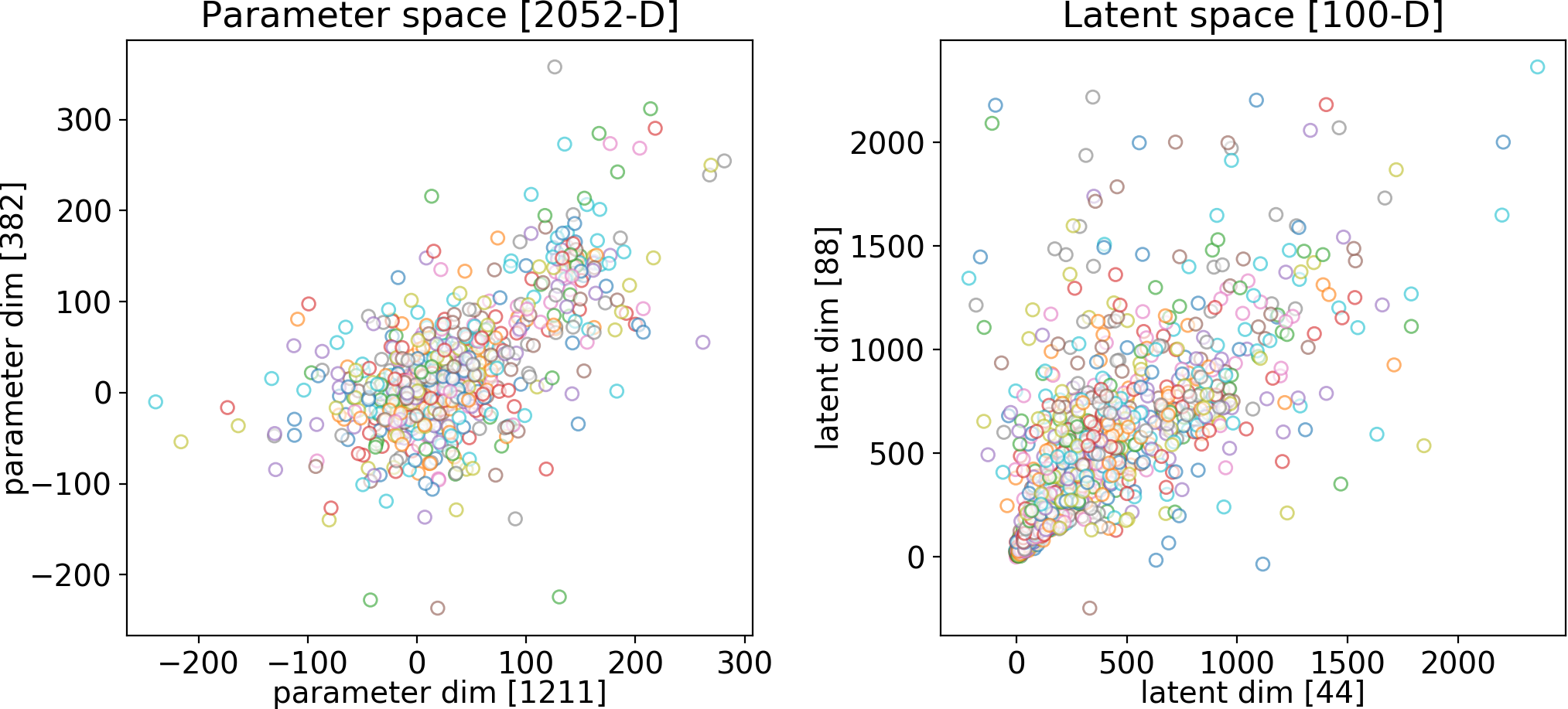}
            \label{fig:rep:bipedal_walker_mix:poms_no}} 
    \caption{Comparison of parameter and latent spaces for Bipedal-Walker mix-scale experiment.}
    \label{fig:rep:bipedal_walker_mix}
    \end{figure}

\begin{figure}[h!]
            \subfloat[PoMS]{\includegraphics[width=0.487\textwidth]
            {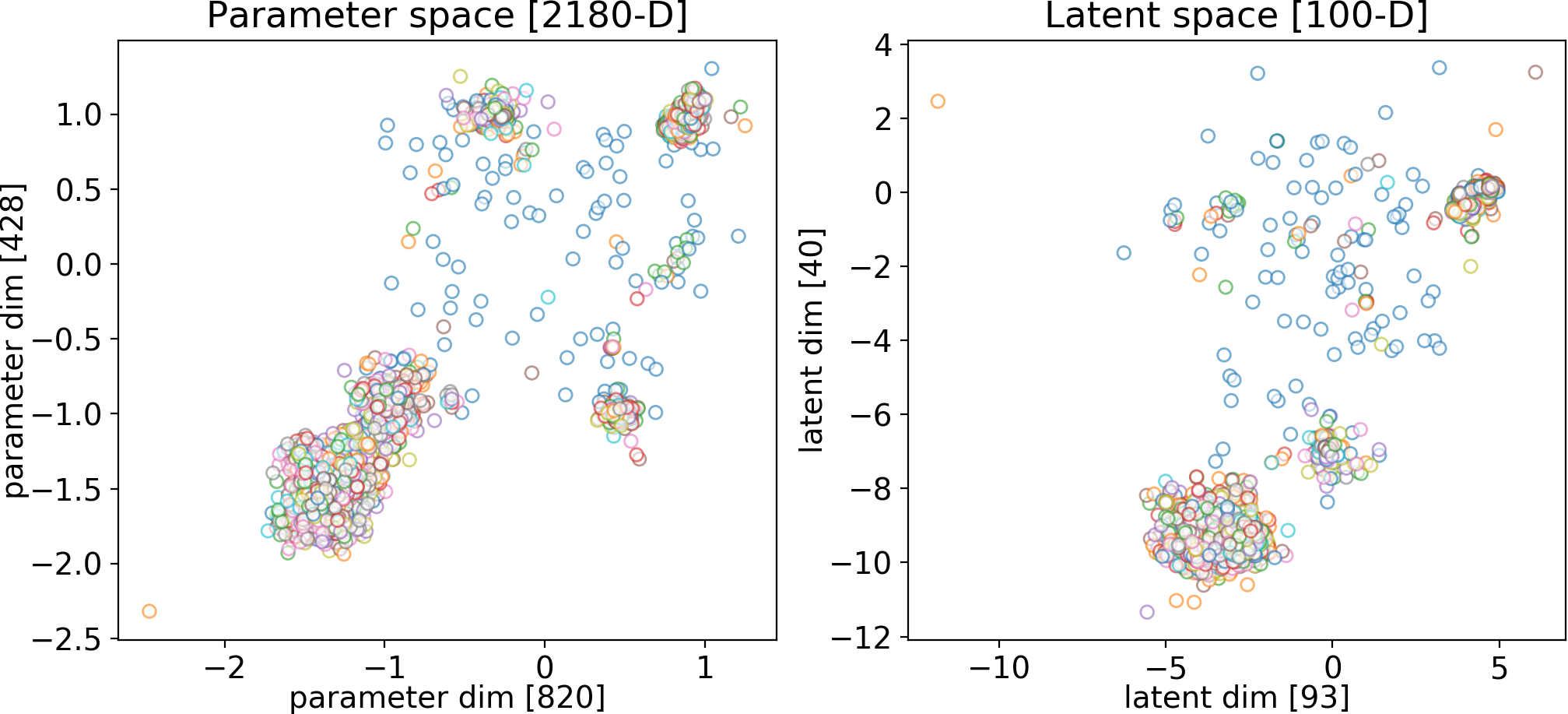}
            \label{fig:rep:bipedal_kicker:poms}} 
            \hfill
            \subfloat[PoMS-no-jacobian]{\includegraphics[width=0.5\textwidth]
            {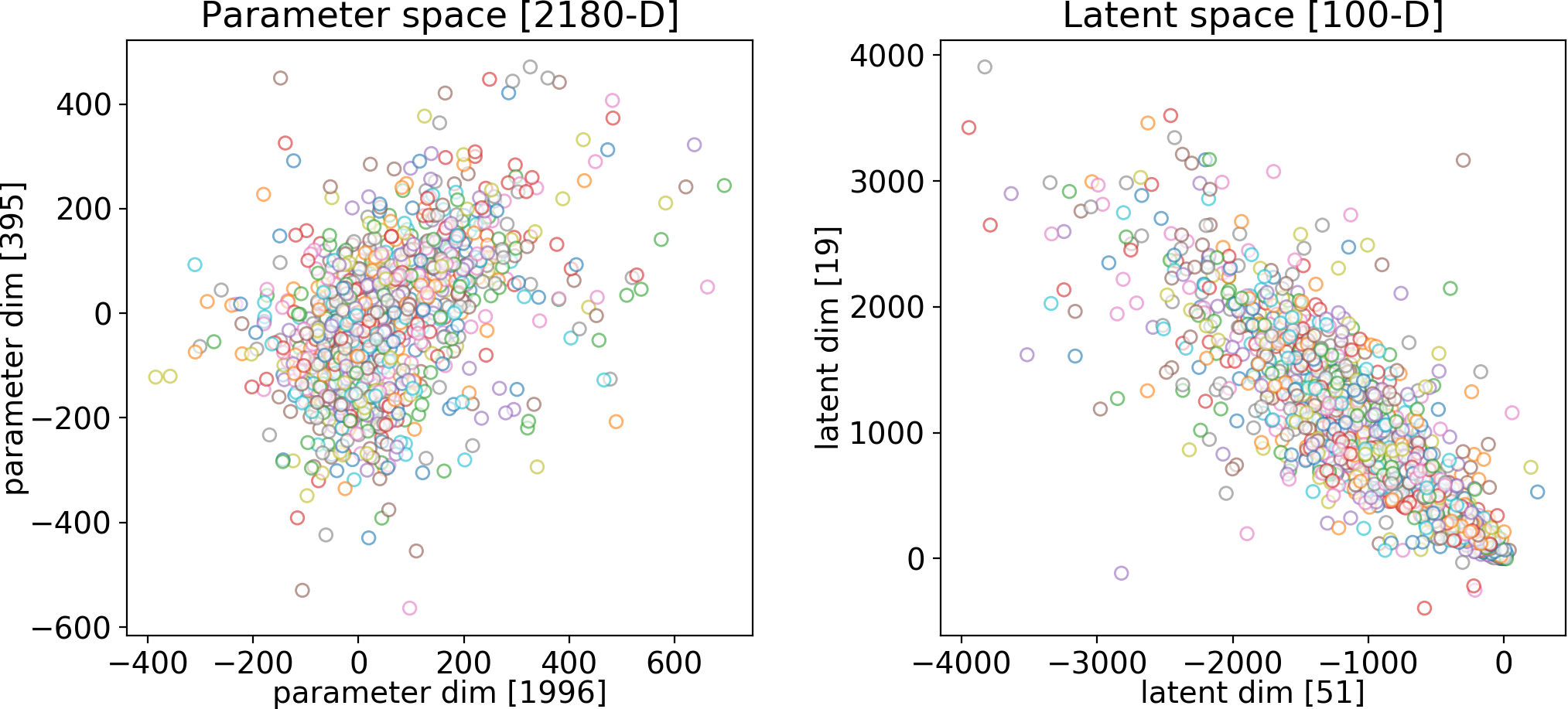}
            \label{fig:rep:bipedal_kicker:poms_no}} 
    \caption{Comparison of parameter and latent spaces for Bipedal-Kicker experiment.}
    \label{fig:rep:bipedal_kicker}
    \end{figure}

\begin{figure}[h!]
            \subfloat[PoMS]{\includegraphics[width=0.49\textwidth]
            {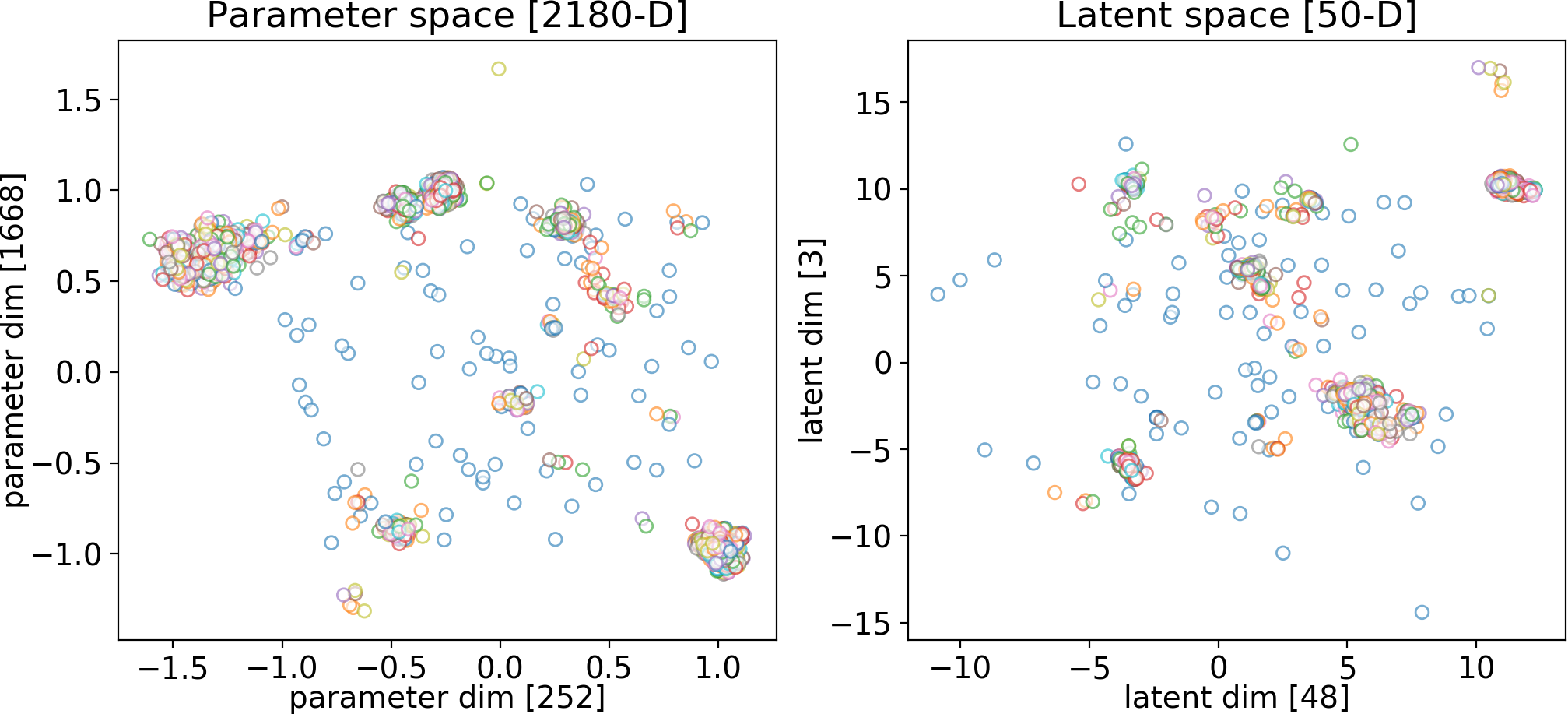}
            \label{fig:rep:bipedal_kicker_mix:poms}} 
            \hfill
            \subfloat[PoMS-no-jacobian]{\includegraphics[width=0.495\textwidth]
            {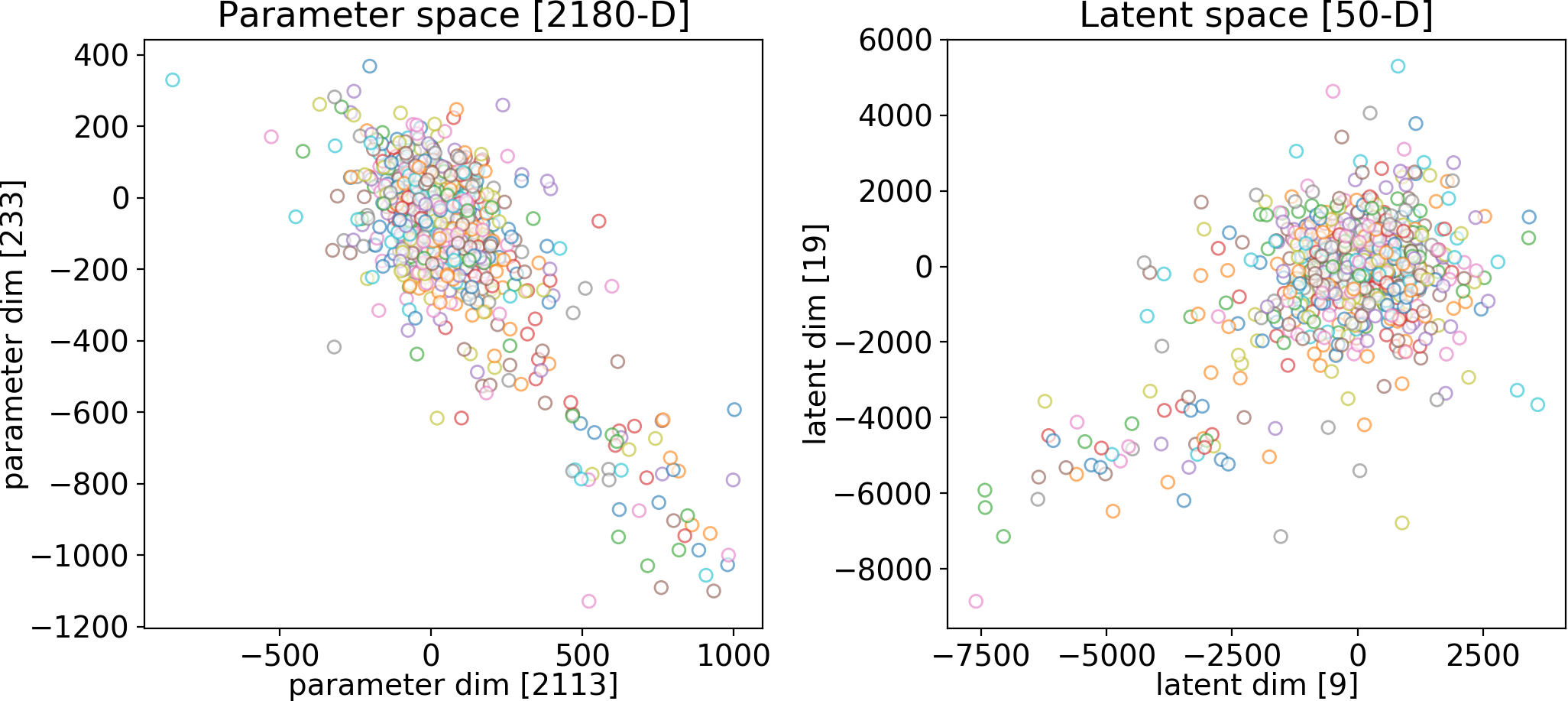}
            \label{fig:rep:bipedal_kicker_mix:poms_no}} 
    \caption{Comparison of parameter and latent spaces for Bipedal-Kicker mix-scale experiment.}
    \label{fig:rep:bipedal_kicker_mix}
    \end{figure}

    We examine the effect that the Jacobian scaling has on the search process, and by extension the final policy collection. 
    By visualising the parameter and latent spaces along dimensions of highest variance, we can see how PoMS-no-jacobian usually diverges to extremely large parameter values (Figures \ref{fig:rep:striker}-\ref{fig:rep:bipedal_kicker_mix}). 
    This is due to the additional decoder transformation which is not accounted for.
    The colouring of the points is the same as in the trajectory plots from the previous section.
    %
    %

    
\end{appendices}


\newpage



\printbibliography

@Article{rakicevic2019informed,
         title="Active learning via informed search in movement parameter space for efficient robot task learning and transfer",
         author="Rakicevic, Nemanja and Kormushev, Petar",
         journal="Autonomous Robots",
         year="2019"
}

@article{schulman2017proximal,
  title={Proximal policy optimization algorithms},
  author={Schulman, John and Wolski, Filip and Dhariwal, Prafulla and Radford, Alec and Klimov, Oleg},
  journal={arXiv preprint arXiv:1707.06347},
  year={2017}
}

@article{salimans2017evolution,
  title={Evolution strategies as a scalable alternative to reinforcement learning},
  author={Salimans, Tim and Ho, Jonathan and Chen, Xi and Sidor, Szymon and Sutskever, Ilya},
  journal={arXiv preprint arXiv:1703.03864},
  year={2017}
}

@article{such2017deep,
  title={Deep neuroevolution: Genetic algorithms are a competitive alternative for training deep neural networks for reinforcement learning},
  author={Such, Felipe Petroski and Madhavan, Vashisht and Conti, Edoardo and Lehman, Joel and Stanley, Kenneth O and Clune, Jeff},
  journal={arXiv preprint arXiv:1712.06567},
  year={2017}
}

@inproceedings{finn2017model,
  title={Model-agnostic meta-learning for fast adaptation of deep networks},
  author={Finn, Chelsea and Abbeel, Pieter and Levine, Sergey},
  booktitle={International Conference on Machine Learning},
  year={2017},
}

@inproceedings{rusu2018meta,
  author    = {Andrei A. Rusu and
               Dushyant Rao and
               Jakub Sygnowski and
               Oriol Vinyals and
               Razvan Pascanu and
               Simon Osindero and
               Raia Hadsell},
  title     = {Meta-Learning with Latent Embedding Optimization},
  booktitle = {International Conference on Learning Representations},
  year      = {2019},
}

@article{chang2018agent,
  title={Agent Embeddings: A Latent Representation for Pole-Balancing Networks},
  author={Chang, Oscar and Kwiatkowski, Robert and Chen, Siyuan and Lipson, Hod},
  journal={arXiv preprint arXiv:1811.04516},
  year={2018}
}

@article{jegorova2018generative,
  title={Generative Adversarial Policy Networks for Behavioural Repertoire},
  author={Jegorova, Marija and Doncieux, St{\'e}phane and Hospedales, Timothy},
  journal={arXiv preprint arXiv:1811.02945},
  year={2018}
}

@article{li2018measuring,
  title={Measuring the intrinsic dimension of objective landscapes},
  author={Li, Chunyuan and Farkhoor, Heerad and Liu, Rosanne and Yosinski, Jason},
  journal={arXiv preprint arXiv:1804.08838},
  year={2018}
}

@inproceedings{chen2018active,
  title={Active learning based on data uncertainty and model sensitivity},
  author={Chen, Nutan and Klushyn, Alexej and Paraschos, Alexandros and Benbouzid, Djalel and Van der Smagt, Patrick},
  booktitle={International Conference on Intelligent Robots and Systems},
  year={2018}
}

@article{richemond2019biologically,
  title={Biologically inspired architectures for sample-efficient deep reinforcement learning},
  author={Richemond, Pierre H and Kolbeinsson, Arinbj{\"o}rn and Guo, Yike},
  journal={Deep Reinforcement Learning Workshop, NeurIPS},
  year={2019}
}

@article{cayton2005algorithms,
  title={Algorithms for manifold learning},
  author={Cayton, Lawrence},
  journal={Univ. of California at San Diego Tech. Rep},
  year={2005}
}

@article{bengio2013representation,
  title={Representation learning: A review and new perspectives},
  author={Bengio, Yoshua and Courville, Aaron and Vincent, Pascal},
  journal={IEEE Transactions on Pattern Analysis and Machine Intelligence},
  year={2013},
  publisher={IEEE}
}

@article{pere2018unsupervised,
  title={Unsupervised Learning of Goal Spaces for Intrinsically Motivated Goal Exploration},
  author={P{\'e}r{\'e}, Alexandre and Forestier, S{\'e}bastien and Sigaud, Olivier and Oudeyer, Pierre-Yves},
  journal={arXiv preprint arXiv:1803.00781},
  year={2018}
}

@inproceedings{ziebart2008maximum,
  title={Maximum entropy inverse reinforcement learning.},
  author={Ziebart, Brian D and Maas, Andrew L and Bagnell, J Andrew and Dey, Anind K},
  booktitle={Conference on Artificial Intelligence},
  year={2008}
}

@inproceedings{gregor2017variational,
  author    = {Karol Gregor and
               Danilo Jimenez Rezende and
               Daan Wierstra},
  title     = {Variational Intrinsic Control},
  booktitle = {International Conference on Learning Representations},
  year      = {2017}
}

@InProceedings{haarnoja2018soft,
  title = 	 {Soft Actor-Critic: Off-Policy Maximum Entropy Deep Reinforcement Learning with a Stochastic Actor},
  author    = {Haarnoja, Tuomas and Zhou, Aurick and Abbeel, Pieter and Levine, Sergey},
  booktitle = {International Conference on Machine Learning},
  year      = {2018}
}

@inproceedings{eysenbach2018diversity,
  author    = {Benjamin Eysenbach and
               Abhishek Gupta and
               Julian Ibarz and
               Sergey Levine},
  title     = {Diversity is All You Need: Learning Skills without a Reward Function},
  booktitle = {International Conference on Learning Representations},
  year      = {2019}
}

@article{sharma2019dynamics,
  title={Dynamics-aware unsupervised discovery of skills},
  author={Sharma, Archit and Gu, Shixiang and Levine, Sergey and Kumar, Vikash and Hausman, Karol},
  journal={arXiv preprint arXiv:1907.01657},
  year={2019}
}

@article{achiam2018variational,
  title={Variational option discovery algorithms},
  author={Achiam, Joshua and Edwards, Harrison and Amodei, Dario and Abbeel, Pieter},
  journal={arXiv preprint arXiv:1807.10299},
  year={2018}
}

@inproceedings{hausman2018transferable,
  author    = {Karol Hausman and
               Jost Tobias Springenberg and
               Ziyu Wang and
               Nicolas Heess and
               Martin A. Riedmiller},
  title     = {Learning an Embedding Space for Transferable Robot Skills},
  booktitle = {International Conference on Learning Representations},
  year      = {2018}
}

@article{cully2017quality,
  title={Quality and diversity optimization: A unifying modular framework},
  author={Cully, Antoine and Demiris, Yiannis},
  journal={IEEE Transactions on Evolutionary Computation},
  year={2017},
  publisher={IEEE}
}

@article{pugh2016quality,
  title={Quality diversity: A new frontier for evolutionary computation},
  author={Pugh, Justin K and Soros, Lisa B and Stanley, Kenneth O},
  journal={Frontiers in Robotics and AI},
  year={2016},
  publisher={Frontiers}
}

@inproceedings{gaier2020discovering,
  title={Discovering Representations for Black-box Optimization},
  author={Gaier, Adam and Asteroth, Alexander and Mouret, Jean-Baptiste},
  booktitle={Genetic and Evolutionary Computation Conference},
  year={2020}
}

@inproceedings{urquhart2019illumination,
    title={An illumination algorithm approach to solving the micro-depot routing problem},
    author={Urquhart, Neil and H{\"o}hl, Silke and Hart, Emma},
    booktitle={Genetic and Evolutionary Computation Conference},
    year={2019}
}

@article{gravina2019procedural,
    title={Procedural content generation through quality diversity},
    author={Gravina, Daniele and Khalifa, Ahmed and Liapis, Antonios and Togelius, Julian and Yannakakis, Georgios N},
    journal={arXiv preprint arXiv:1907.04053},
    year={2019} 
}

@inproceedings{vassiliades2018discovering,
  title={Discovering the elite hypervolume by leveraging interspecies correlation},
  author={Vassiliades, Vassilis and Mouret, Jean-Baptiste},
  booktitle={Genetic and Evolutionary Computation Conference},
  year={2018}
}

@inproceedings{cully2019autonomous,
  title={Autonomous skill discovery with quality-diversity and unsupervised descriptors},
  author={Cully, Antoine},
  booktitle={Genetic and Evolutionary Computation Conference},
  year={2019}
}

@inproceedings{colas2020scaling,
  author    = {C{\'{e}}dric Colas and
               Vashisht Madhavan and
               Joost Huizinga and
               Jeff Clune},
  editor    = {Carlos Artemio Coello Coello},
  title     = {Scaling MAP-Elites to Deep Neuroevolution},
  booktitle = {Genetic and Evolutionary Computation Conference},
  year      = {2020}
}

@article{cully2015robots,
  title={Robots that can adapt like animals},
  author={Cully, Antoine and Clune, Jeff and Tarapore, Danesh and Mouret, Jean-Baptiste},
  journal={Nature},
  year={2015},
  publisher={Nature Publishing Group}
}

@article{mouret2015illuminating,
  title={Illuminating search spaces by mapping elites},
  author={Mouret, Jean-Baptiste and Clune, Jeff},
  journal={arXiv preprint arXiv:1504.04909},
  year={2015}
}

@inproceedings{lehman2011evolving,
  title={Evolving a diversity of virtual creatures through novelty search and local competition},
  author={Lehman, Joel and Stanley, Kenneth O},
  booktitle={Genetic and Evolutionary Computation Conference},
  year={2011}
}

@inproceedings{meyerson2016learning,
  title={Learning behavior characterizations for novelty search},
  author={Meyerson, Elliot and Lehman, Joel and Miikkulainen, Risto},
  booktitle={Genetic and Evolutionary Computation Conference},
  year={2016}
}

@inproceedings{rifai2011contractive,
  title     = {Contractive Auto-Encoders: Explicit Invariance During Feature Extraction},
  author    = {Rifai, Salah and Vincent, Pascal and Muller, Xavier and Glorot, Xavier and Bengio, Yoshua},
  booktitle = {International Conference on Machine Learning},
  year      = {2011}
}

@inproceedings{rifai2012generative,
  title={A generative process for sampling contractive auto-encoders},
  author={Rifai, Salah and Bengio, Yoshua and Dauphin, Yann N and Vincent, Pascal},
  booktitle={International Conference on Machine Learning},
  year={2012}
}

@inproceedings{bellemare2016unifying,
  title={Unifying count-based exploration and intrinsic motivation},
  author={Bellemare, Marc and Srinivasan, Sriram and Ostrovski, Georg and Schaul, Tom and Saxton, David and Munos, Remi},
  booktitle={Advances in neural information processing systems},
  year={2016}
}

@inproceedings{tang2017exploration,
  title={\# exploration: A study of count-based exploration for deep reinforcement learning},
  author={Tang, Haoran and Houthooft, Rein and Foote, Davis and Stooke, Adam and Chen, OpenAI Xi and Duan, Yan and Schulman, John and DeTurck, Filip and Abbeel, Pieter},
  booktitle={Advances in neural information processing systems},
  year={2017}
}

@inproceedings{badia2020nevergiveup,
  author    = {Adri{\`{a}} Puigdom{\`{e}}nech Badia and
               Pablo Sprechmann and
               Alex Vitvitskyi and
               Daniel Guo and
               Bilal Piot and
               Steven Kapturowski and
               Olivier Tieleman and
               Mart{\'{\i}}n Arjovsky and
               Alexander Pritzel and
               Andrew Bolt and
               Charles Blundell},
  title     = {Never Give Up: Learning Directed Exploration Strategies},
  booktitle = {International Conference on Learning Representations},
  year      = {2020}
}

@inproceedings{plappert2017parameter,
  author    = {Matthias Plappert and
               Rein Houthooft and
               Prafulla Dhariwal and
               Szymon Sidor and
               Richard Y. Chen and
               Xi Chen and
               Tamim Asfour and
               Pieter Abbeel and
               Marcin Andrychowicz},
  title     = {Parameter Space Noise for Exploration},
  booktitle = {International Conference on Learning Representations},
  year      = {2018}
}

@inproceedings{fortunato2017noisy,
  author    = {Meire Fortunato and
               Mohammad Gheshlaghi Azar and
               Bilal Piot and
               Jacob Menick and
               Matteo Hessel and
               Ian Osband and
               Alex Graves and
               Volodymyr Mnih and
               R{\'{e}}mi Munos and
               Demis Hassabis and
               Olivier Pietquin and
               Charles Blundell and
               Shane Legg},
  title     = {Noisy Networks For Exploration},
  booktitle = {International Conference on Learning Representations},
  year      = {2018}
}

@inproceedings{glorot2010understanding,
  title={Understanding the difficulty of training deep feedforward neural networks},
  author={Glorot, Xavier and Bengio, Yoshua},
  booktitle={International Conference on Artificial Iintelligence and Statistics},
  year={2010}
}

@article{catto2011box2d,
  title={Box2d: A 2d physics engine for games},
  author={Catto, Erin},
  journal={URL: http://www.box2d.org},
  year={2011}
}

@article{brockman2016openai,
  title={Openai gym},
  author={Brockman, Greg and Cheung, Vicki and Pettersson, Ludwig and Schneider, Jonas and Schulman, John and Tang, Jie and Zaremba, Wojciech},
  journal={arXiv preprint arXiv:1606.01540},
  year={2016}
}

@inproceedings{flageat2020fast,
  title={Fast and stable MAP-Elites in noisy domains using deep grids},
  author={Flageat, Manon and Cully, Antoine},
  booktitle={Artificial Life Conference Proceedings},
  pages={273--282},
  year={2020},
  organization={MIT Press}
}

@inproceedings{justesen2019map,
  title={MAP-Elites for noisy domains by adaptive sampling},
  author={Justesen, Niels and Risi, Sebastian and Mouret, Jean-Baptiste},
  booktitle={Proceedings of the Genetic and Evolutionary Computation Conference Companion},
  pages={121--122},
  year={2019}
}
\end{refsection}



\end{document}